\documentclass[11pt,a4paper,oneside]{book}

\usepackage[utf8]{inputenc}
\usepackage[T1]{fontenc}
\usepackage[english]{babel}

\usepackage{amsmath}
\usepackage{amsfonts}
\usepackage{amssymb}
\usepackage{amsthm}
\usepackage{mathtools}
\usepackage{mathrsfs}
\usepackage{nicefrac}
\usepackage{stmaryrd}
\usepackage{trimclip}

\usepackage[dvipsnames]{xcolor}
\usepackage{soul}
\usepackage[labelfont=bf,textfont=md]{caption}

\usepackage{graphicx}
\usepackage{float}
\usepackage{subcaption}
\usepackage{wrapfig}
\usepackage{tikz}
\usepackage[skins,breakable]{tcolorbox}

\usepackage{booktabs}
\usepackage{longtable}
\usepackage{tabularx}
\usepackage{multirow}

\usepackage{geometry}
\usepackage{setspace}
\usepackage{parskip}
\usepackage{microtype}
\usepackage{titlesec}
\usepackage{fancyhdr}
\usepackage{titling}

\usepackage{natbib}
\usepackage{url}
\usepackage{hyperref}

\usepackage[nameinlink,capitalize]{cleveref}
\Crefformat{equation}{Eq.~(#2#1#3)}
\crefformat{equation}{Eq.~(#2#1#3)}
\Crefformat{figure}{Fig.~(#2#1#3)}
\crefformat{figure}{Fig.~(#2#1#3)}

\Crefformat{section}{Section~(#2#1#3)}
\crefformat{section}{Section~(#2#1#3)}
\Crefformat{chapter}{Chapter~(#2#1#3)}
\crefformat{chapter}{Chapter~(#2#1#3)}

\usepackage{listings}
\usepackage{algorithm}
\usepackage{algorithmic}

\usepackage{enumitem}
\usepackage{lipsum}
\usepackage{appendix}

\fancypagestyle{plain}{%
    \fancyhf{}
    \fancyfoot[C]{\thepage}
    
}

\titleformat{\chapter}[display]
{\normalfont\huge\bfseries}{\chaptertitlename\ \thechapter}{20pt}{\Huge}
\titlespacing*{\chapter}{0pt}{0pt}{40pt}

\hypersetup{
    colorlinks=true,
    linkcolor=blue,
    filecolor=magenta,
    urlcolor=cyan,
    citecolor=red,
    pdftitle={Your Document Title},
    pdfauthor={Your Name},
    pdfsubject={Subject},
    pdfkeywords={keyword1, keyword2, keyword3}
}

\theoremstyle{definition}

\newtheorem{theorem}{Theorem}[chapter]

\definecolor{darkblue}{RGB}{0,0,139}
\definecolor{darkgreen}{RGB}{0,100,0}
\definecolor{darkred}{RGB}{139,0,0}

\newcommand{\norm}[1]{\left\lVert#1\right\rVert}

\newcommand{\E}[1]{\mathbb{E}\left[#1\right]}

\setcitestyle{numbers,square}

{\cleardoublepage\null\vfill\begin{center}%
\bfseries\abstractname\end{center}}%
{\vfill\null}

\usepackage{amsmath,amsfonts,bm}

\def\eqref#1{equation~\ref{#1}}

\def\1{\bm{1}}

\def\rvc{{\mathbf{c}}}

\def\rvf{{\mathbf{f}}}

\def\rvv{{\mathbf{v}}}
\def\rvw{{\mathbf{w}}}
\def\rvx{{\mathbf{x}}}
\def\rvy{{\mathbf{y}}}
\def\rvz{{\mathbf{z}}}

\def\ervx{{\textnormal{x}}}

\def\rmXi{{\boldsymbol{\Xi}}}

\def\vtheta{{\bm{\theta}}}

\def\vm{{\bm{m}}}

\def\mI{{\bm{I}}}

\def\mM{{\bm{M}}}

\def\mT{{\bm{T}}}

\DeclareMathAlphabet{\mathsfit}{\encodingdefault}{\sfdefault}{m}{sl}
\SetMathAlphabet{\mathsfit}{bold}{\encodingdefault}{\sfdefault}{bx}{n}

\def\gE{{\mathcal{E}}}
\def\gF{{\mathcal{F}}}

\def\gH{{\mathcal{H}}}

\def\gM{{\mathcal{M}}}
\def\gN{{\mathcal{N}}}

\def\gU{{\mathcal{U}}}

\def\gX{{\mathcal{X}}}
\def\gY{{\mathcal{Y}}}
\def\gZ{{\mathcal{Z}}}

\def\bxi{{\boldsymbol{\xi}}}
\def\bXi{{\boldsymbol{\Xi}}}
\def\bomega{{\boldsymbol{\omega}}}

\newcommand{\pdata}{p_{\rm{data}}}
\newcommand{\Ls}{\mathcal{L}}
\newcommand{\R}{\mathbb{R}}

\newcommand{\ip}[2]{\left\langle #1, #2 \right\rangle}

\DeclareMathOperator*{\argmin}{arg\,min}

\newcommand{\pp}[2]{\frac{\partial #1}{\partial #2}}

\newcommand{\nbr}[1]{\mathtt{N}(#1)}

\makeatletter
\DeclareRobustCommand{\shortto}{%
  \mathrel{\mathpalette\short@to\relax}%
}

\newcommand{\short@to}[2]{%
  \mkern2mu
  \clipbox{{.5\width} 0 0 0}{$\m@th#1\vphantom{+}{\shortrightarrow}$}%
  }
\makeatother 
\newcommand{\relu}[1]{\operatorname{ReLU}\left(#1\right)}

\renewcommand{\llbracket}{\left[ \! \left[}
\renewcommand{\rrbracket}{\right] \! \right]}
\newcommand{\iset}[1]{\llbracket #1 \rrbracket}

\newcommand{\hamux}{\textsc{HAMUX}}
\definecolor{hamux-orange}{HTML}{F6511D}
\definecolor{hamux-blue}{HTML}{3291F9}
\definecolor{hamux-violet}{HTML}{E18EFF}

\newtcolorbox{calloutInfo}[1][]{
    enhanced,
    colback=green!5,
    colframe=green!50,
    coltitle=black,
    breakable = true,
    fonttitle=\bfseries,
    title={#1},
    before upper={\setlength{\parskip}{0.75em}},
}

\newtcolorbox{calloutDef}[1][]{
    enhanced,
    colback=gray!5,
    colframe=gray!50,
    coltitle=black,
    title=\textbf{Definition:} {#1},
    before upper={\setlength{\parskip}{0.75em}},
}

\newtcolorbox{calloutProof}[1][]{
    enhanced,
    after skip=8mm,%
    title={\textbf{Proof:} #1},
    breakable = true,
    fonttitle=\sffamily\bfseries,
    coltitle=black,
    before upper={\setlength{\parskip}{0.75em}},
    colback=orange!5,
    colframe=orange!50,
    }

\newcounter{notebookcounter}[chapter]
\newcounter{exercisecounter}[chapter]

\renewcommand{\thenotebookcounter}{\arabic{notebookcounter}}
\renewcommand{\theexercisecounter}{\arabic{exercisecounter}}

\newtcolorbox{calloutNotebook}[1][]{
    enhanced,
    after skip=8mm,%
    title={\refstepcounter{notebookcounter}\textbf{Notebook \thechapter.\thenotebookcounter:} #1},
    fonttitle=\sffamily\bfseries,
    coltitle=black,
    before upper={\setlength{\parskip}{0.75em}},
    colback=hamux-blue!3,
    colframe=hamux-blue!40,
}

\newcommand{\nblinks}[3]{
    \footnotesize\sffamily\bfseries
    Checkout the notebook as a \href{#1}{blog post}, a \href{https://colab.research.google.com/github/bhoov/amtutorial/blob/main/tutorial_ipynbs/#3.ipynb}{colab notebook} or as a \href{#2}{raw .ipynb} file.
}

\newtcolorbox{calloutExercise}[1][]{
    enhanced,
    after skip=8mm,%
    title={\refstepcounter{exercisecounter}\textbf{Exercise \thechapter.\theexercisecounter:} #1},
    breakable = true,
    fonttitle=\sffamily\bfseries,
    coltitle=black,
    before upper={\setlength{\parskip}{0.75em}},
    colback=hamux-violet!5,
    colframe=hamux-violet!40,
} 
\newcommand{\tutorialtitle}{Modern Methods in Associative Memory}

\newcommand{\tutorialdate}{July 14, 2025}

\newcommand{\tutorialwebsite}{https://tutorial.amemory.net}

\newcommand{\tutorialauthors}{
    \textbf{Dmitry Krotov}\textsuperscript{1,2}, 
    \textbf{Benjamin Hoover}\textsuperscript{1,3}, 
    \textbf{Parikshit Ram}\textsuperscript{1}, 
    \textbf{Bao Pham}\textsuperscript{1,4}
}

\newcommand{\tutorialaffiliations}{
    \textsuperscript{1}IBM Research,
    \textsuperscript{2}MIT,
    \textsuperscript{3}Georgia Tech,
    \textsuperscript{4}RPI\\
    Contact Email: krotov@ibm.com
}

\newcommand{\tutorialabstract}{
    Associative Memories like the famous Hopfield Networks are elegant models for describing fully recurrent neural networks whose fundamental job is to store and retrieve information. In the past few years they experienced a surge of interest due to novel theoretical results pertaining to their information storage capabilities, and their relationship with SOTA AI architectures, such as Transformers and Diffusion Models. These connections open up possibilities for interpreting the computation of traditional AI networks through the theoretical lens of Associative Memories. Additionally, novel Lagrangian formulations of these networks make it possible to design powerful distributed models that learn useful representations and inform the design of novel architectures. This tutorial provides an approachable introduction to Associative Memories, emphasizing the modern language and methods used in this area of research, with practical hands-on mathematical derivations and coding notebooks.
}

\newcommand{\maketutorialtitle}{
    \vspace*{1cm}

    {\fontsize{22pt}{26pt}\selectfont\bfseries
    \tutorialtitle
    }
    
    \vspace{0.5cm}

    {\fontsize{11pt}{15pt}\selectfont\sffamily
    \tutorialauthors
    }

    {\small
    \tutorialaffiliations
    }
    \vspace{0.5cm}

    {\small
        \textbf{Date:} \tutorialdate\\
        \textbf{Tutorial:} ICML 2025, \textit{Vancouver, BC, Canada}\\
        \textbf{Website:} \href{\tutorialwebsite}{\tutorialwebsite}
    }
    \thispagestyle{empty}
    \vspace{1cm}
    
    \textbf{Abstract}\\
    \tutorialabstract
}

\begin{document}

\maketutorialtitle

\tableofcontents
\thispagestyle{empty}
\newpage

\mainmatter

\clearpage{}%
\chapter{Introduction}
\label{chap:introduction}

\begin{figure}[htbp]
    \centering
    \includegraphics[width=0.81\textwidth]{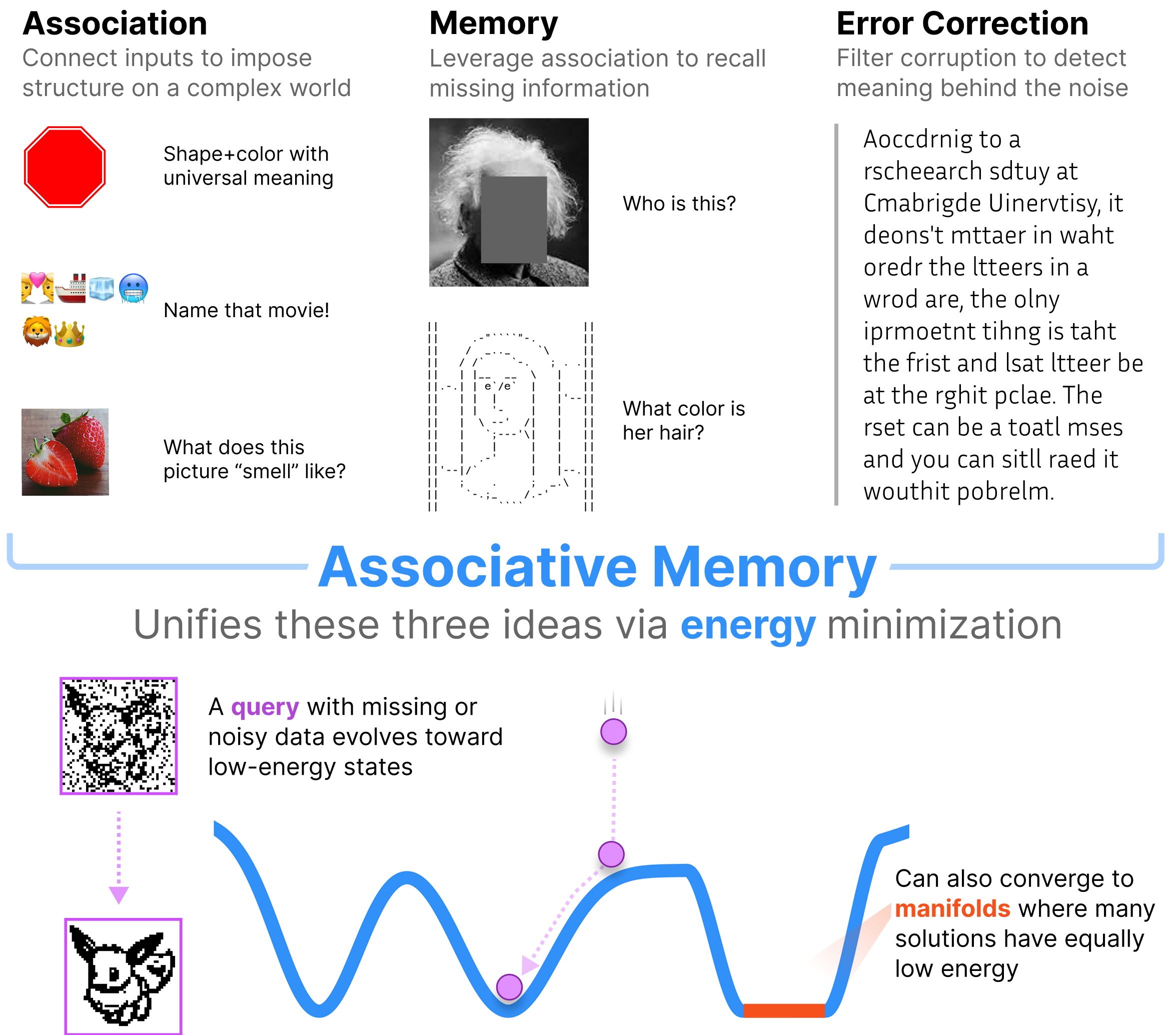}
    \caption{The form of Associative Memory discussed in this tutorial uses an energy function to unify three important aspects of human cognition: association, memory, and error correction. We are capable of associating images, sights, sounds, smells, and symbols with each other. These associations allow us to retrieve memories using partial or corrupted information, making it a content-addressable memory with error-correction capabilities. The functionality of Associative Memory is modeled by an energy function, where low values of the energy correspond to stored memories and constitute the most likely states of the system.}
    \label{fig:intro:fig1}
\end{figure}

Associative Memory (AM) is a core concept in psychology responsible for linking related items \cite{TulvingMemory}. For instance, if one is shown an image of a strawberry, it is likely that they can recall the smell and taste of this fruit; or, in the case of an image of a person, an acquaintance of them would be able to name them, see \Cref{fig:intro:fig1} for the demonstration of AM. These are examples of input-output pairs that are associated in our memory, where prompting for an element of the pair results in content-addressable retrieval of the other element.

Another important aspect of Associative Memory is the notion of error correction. You can easily read the text in \Cref{fig:intro:fig1} without much difficulty, despite the fact that almost no words in that paragraph are proper English words. The reason why you are able to comprehend this text is because there are powerful error correction mechanisms that are constantly working in your brain that associate imperfect inputs with the proper semantic meaning of individual words. The same applies to the example above: the image of the strawberry can be presented with all kinds of distortions and imperfections. Despite all that variability in the input, Associative Memory manages to link those inputs with the proper smell and taste. 

Thus, {\bf Associative Memory is a content addressable information storage system that is capable of error correction}. 

Associative Memory played a major role in the history of AI \cite{mcclelland2025profile, stone2024theartificial}. Following the 1943 model of artificial neuron by McCulloch and Pitts \cite{mcculloch1943logical}, and a body of work \cite{rosenblatt1962principles} on artificial neural networks (ANNs) -- Perceptrons -- by Frank Rosenblatt in the 1950-1960s, the community at large has been extremely enthusiastic about the future of AI. Popular media outlets from that period promised that the Perceptron ``will be able to walk, talk, see, write, reproduce itself and be conscious of its existence''~\cite{NYTPerceptrons}, similar to what we read in popular press about AI today. However, in 1969, Minsky and Papert demonstrated that simple Perceptrons (without hidden layers) could not compute even the simplest logical gates, e.g., XOR \cite{marvin1969perceptrons}. The public perception of this result led to the drop of enthusiasm in ANNs. Most of the computer science community at that time left the field of ANNs --- triggering what historians of science later called the ``AI Winter'' \cite{nobelprizephysics2024}.

John Hopfield's seminal paper of 1982 \cite{hopfield1982neural} on what is now called the Hopfield network of Associative Memory was the major driving force that ended that period. In his paper, Hopfield connected computational aspects of Associative Memory with collective properties of Ising \cite{Lenz-Ising-history1967} magnets in condensed matter physics, which were a ``hot topic'' at the time. Specifically, Hopfield posed a simple quantifiable problem: given a network of $D$ neurons, how much information (or memories) can such a network store and retrieve? Content-addressable Associative Memory retrieval was a sufficiently non-trivial problem to illustrate the potential of ANNs' computational abilities. At the same time, it was simple enough to be analytically solvable using powerful methods developed in physics. The convergence of these two properties created a ``harmonic oscillator'' level abstraction for ANN computation, and set the grounds for many extensions and generalizations that followed.

Associative memory has been a prominent theme of ANN research in the 1960s-1980s. A highly incomplete and subjective list of seminal papers from that period includes: Anderson \cite{anderson1968memory}, Willshow, Buneman, Longuet-Higgins \cite{willshaw1969non}, Amari \cite{amari1972learning,amari1977neural}, Cohen and Grossberg \cite{cohen1983absolute}, Hopfield \cite{hopfield1984Neurons}, Amit, Gutfreund, Sompolinsky \cite{amit1985storing}, and many others. 

The main focus of this tutorial is on the {\bf Energy-based Associative Memories}. This is the class of ANNs which are recurrent neural networks that can be described by a state vector, which evolves in time according to some non-linear rule. This state vector can be either continuous or discrete. The update rule can be written either in terms of continuous time (differential equation), or discrete set of update steps (which we will usually treat as a discretization of that differential equation). Finally, there exist multiple options for how one updates the state vector. The most common choices are: \textit{synchronous} --- all compute elements of the state vector are updated simultaneously, or \textit{asynchronous} --- at any given time a subset of all elements of the state vector are updated, i.e., a random element of the state vector is updated while the remaining elements are kept intact. For the purposes of this tutorial, we will mostly work with continuous states, continuous time, and synchronous updates. Thus, the state vector of the network $\mathbf{x}\in \mathbb{R}^D$, which has individual elements $x_i$ (index $i$ runs from $1$ to $D$), evolves according to the following differential equation:
\begin{equation}
    \frac{dx_i}{dt} = f_i(\mathbf{x}, t) \label{eq:intro:general dynamical equation} 
\end{equation}
where the functions $f_i(\mathbf{x}, t)$ represent the vector field that defines the dynamics. We will refer to individual elements $x_i$ of the state vector as ``neurons'', although in many situations these variables may describe a different biological structure, e.g., an astrocyte or their processes (long tentacles originating from the astrocyte's cell body).  

A general system of coupled non-linear differential equations may have many complex behaviors: fixed points, limit cycles, strange attractors, or chaotic behavior. 
Energy-based AMs are a special subclass of general systems (\ref{eq:intro:general dynamical equation}) that have the notion of an energy function (sometimes also referred to as a Lyapunov function). You can think of the temporal evolution of the state vector as a ball rolling downhill in a sophisticated energy landscape as seen in \Cref{fig:intro:fig1}. The energy is bounded from below, and the ball is only allowed to move in a way that decreases its energy. Because of these restrictions, eventually, the ball must either stop at one of the local minima or reach a manifold that corresponds to flat energy. In the latter case, the ball may continue to move along that manifold as long as the energy does not increase. 

The local minima of the energy (which can be point-like attractors --- zero-dimensional manifolds --- or alternatively, manifolds of higher dimension) are called {\bf memories}. The process of shaping the energy landscape corresponds to writing information into the AM network, or learning. The dynamical trajectory of energy descent, illustrated by \Cref{eq:intro:general dynamical equation}, corresponds to memory recall, or inference. Association happens between the initial state of the network $\mathbf{x}(t=0)$ and the final asymptotic state of the network $\mathbf{x}(t\rightarrow\infty)$. Finally, the asymptotic states of the dynamics are typically stable (unless they lie on the flat portions of the energy landscape). Intuitively, this means that small perturbations that do not push the state vector outside the basin of the fixed point's attraction gets auto-corrected by the network itself. For these reasons, this network is an AM system. 

In some settings, memories in this system may correspond to individual instances of the training data. Alternatively, they may correspond to emergent attracting manifolds that are shaped by the learning algorithm (e.g., backpropagation \cite{rumelhart1986learning, WERBOS1988339}, Hebbian learning \cite{Hebb1949}, contrastive training \cite{1983Boltzmann, Hopfield1983UnlearningHA}, etc.). In the latter case, the memories do not typically correspond to individual instances of the training data, but rather describe consolidated memories --- ``knowledge'' --- that the network acquired through the synergy of AM architecture, a specific learning algorithm, and the choice of training data.   

Intuitively, you can think about the initial state $\mathbf{x}(t=0)$ as a ``question'' that you ask the neural network. This question positions the state vector at some high-energy location on the energy landscape. The network will perform computation by moving that state to a local minimum (or a metastable state) --- the process of ``thinking.'' Once the local minimum is reached, the computation terminates and the network's state stops evolving in time. You can read out that final state $\mathbf{x}(t\rightarrow\infty)$ and convert it to the answer to the posed question. Importantly, this computation is very different from conventional feed-forward architectures, e.g., convolutional neural networks, or transformers. These conventional architectures are described by a computational graph with a fixed number of steps. This means that if the network has 10 layers, it must produce some kind of an answer to the question after exactly 10 steps (without chain of thought). This happens regardless of the complexity of the posed question. AM architectures are very different. They can dynamically adapt the computational graph based on the complexity of the posed question. For simple questions the network may produce an answer in 5 steps, but for more complicated questions the network may need to ``think'' longer.

Finally, because of the network's energy-based architecture, the final answer is {\bf asymptotically stable}. This means that once the computation or ``thinking'' stopped and the network converged to an answer, the precise timing of the output's readout  doesn't matter. Assuming that the readout time $T$ is large enough, we can use $\mathbf{x}(t=T)$ or $\mathbf{x}(t=T+0.5\ \text{seconds})$ as the final answer, and the two must be identical. This property of asymptotic stability makes AM framework extremely appealing for neuromorphic devices, where hardware imperfections may prevent the ability to read the network's state at a precise timing. 

In the past few years there has been significant advances in the field of AMs. These advances pertain to the development of {\bf Dense Associative Memories} or DenseAMs \cite{krotov2016dense}. They are flexible energy-based AM architectures that are capable of storing large amounts of information, enable incorporation of many useful inductive biases (e.g., convolutions \cite{Fukushima1980NeocognitronAS}, attention \cite{vaswani2017attention, smart2025incontext}, etc.) in their architecture, and have mathematically controllable properties of emergent local minima. DenseAM ideas have triggered a large amount of innovative ideas about the potential use cases of AMs and we believe they will enable a new frontier for AM research \cite{krotov2023new}. In this tutorial, we will cover many of these new developments from both the theoretical perspective and practical implementations. The tutorial is supplemented with a collection of notebooks and suggested problems that the readers can explore on their own to better understand the core ideas and methods, and to get hands-on experience coding an AM network suitable for their own use case. We intentionally designed these problems so that they are simple but, at the same time, illustrate a useful mathematical concept or a core idea in AM. We hope that you enjoy this learning experience.

\clearpage{}%
\clearpage{}%
\chapter{Dense Associative Memory: Discrete State Vector}
\label{chap:DenseAM}

This chapter introduces a popular class of AMs --- Dense Associative Memory (DenseAM). This family of models is a generalization of celebrated Hopfield networks. While Hopfield networks are very elegant mathematical models that satisfy all of the AM requirements, they are known to have a very small information storage capacity, which is insufficient for practical AI applications. DenseAMs are specifically designed to retain all of the benefits of Hopfield networks, but rectify their small information storage issue \cite{krotov2016dense,krotov2018dense}. 

As discussed earlier, AMs can be formulated both in discrete and continuous variables, and in discrete or continuous time. In this chapter, we focus on DenseAMs with discrete state vector, and discrete asynchronous updates. Specifically, we will be working with a set of discrete variables $\sigma_i = \{\pm 1\}$, index $i=1,...,D$, which compose a state vector $\boldsymbol{\sigma}$. In addition to that, the network will have $K$ memory vectors $\boldsymbol{\xi}^\mu$ with index $\mu=1,...,K$. Each memory is a $D$-dimensional vector with individual elements denoted by $\xi^\mu_i$.  

The energy function is defined as: 
\begin{equation}
    E = - \sum\limits_{\mu=1}^K F\Big(\sum\limits_{i=1}^D \xi^\mu_i \sigma_i\Big)
    \label{eq: DenseAM energy}
\end{equation}
The goal of the network is to start at some initial state $\sigma_i^{(t=0)}$, which typically corresponds to a high-energy state, and lower the energy by flipping the elements of the state vector. The dynamics of flipping stops when no further single element flip can reduce the energy. At that point, the network has reached a local minimum of the energy. As usual, we will refer to the individual elements of the state vector are neurons or spins. 

In order to formalize this intuitive dynamical equation, pick a single neuron and define its state at the next iteration as:
\begin{equation}
\begin{split}
    \sigma_i^{(t+1)} &= \text{Sign}\bigg[E\Big(\sigma_i=-1, \sigma_{j\neq i} = \sigma_j^{(t)}\Big) - E\Big(\sigma_i=+1, \sigma_{j\neq i} = \sigma_j^{(t)}\Big)\bigg]  \\
    &=\text{Sign}\bigg[\sum\limits_{\mu=1}^K F\Big(\xi^\mu_i+ \sum\limits_{j\neq i}^D \xi^\mu_j \sigma_j^{(t)}\Big) - \sum\limits_{\mu=1}^K F\Big(-\xi^\mu_i+ \sum\limits_{j\neq i}^D \xi^\mu_j \sigma_j^{(t)}\Big)\bigg] \\ 
    & \approx \text{Sign}\bigg[\sum\limits_{\mu=1}^K 2\ \xi^\mu_i\ F^\prime \Big(\sum\limits_{j\neq i}^D \xi^\mu_j \sigma_j^{(t)}\Big) + \text{higher order subleading terms}\bigg] .
    \label{eq: DenseAM update}
\end{split}
\end{equation}
This update rule compares the energies of two states: $\sigma_i = -1$ with states of all other neurons clamped to their current values, and $\sigma_i = +1$ with all the other neurons clamped. The $\text{Sign}[\cdot]$ function assignes the state of the $i^\text{th}$ neuron to the one corresponding to the lowest energy among these two possibilities. Finally, in the last line of \Cref{eq: DenseAM update} we have used the Taylor series to expand the function $F(x+\varepsilon)\approx F(x) + \varepsilon F^\prime(x) + \text{higher order terms}$. It is legitimate to terminate the expansion after the first term since $\varepsilon = \pm 1$ is much smaller than the overlap between the clamped part of the state vector and the memories. 

This update rule is typically written in the following form: 
\begin{equation}
    \sigma_i^{(t+1)} =\text{Sign}\bigg[\sum\limits_{\mu=1}^K \xi^\mu_i\ f \Big(\sum\limits_{j\neq i}^D \xi^\mu_j \sigma_j^{(t)}\Big) \bigg] ,
    \label{eq: DenseAM update simplified}
\end{equation}
where we dropped the factor of $2$ in the argument of the sign function (it doesn't play any role there) and introduced an activation function $f(\cdot) = F^\prime(\cdot)$, which is a derivative of the function $F(\cdot)$ defining the energy.  

The energy function (\ref{eq: DenseAM energy}) is a finite sum of smooth functions (we assume that the function $F(\cdot)$ does not have singularities - infinite values for finite arguments) that depend on the finite number of discrete variables. Thus, the energy is finite and bounded from below. Additionally, the dynamical equations (\ref{eq: DenseAM update}) and (\ref{eq: DenseAM update simplified}) decrease the value of energy at each iteration. Thus, if we keep applying these update equations to the state vector for a long time, eventually the system will reach a steady state -- no single neuron flip can further reduce the energy. 

\section{Information Storage Capacity}
How many memories or local minima can such a system store and successfully retrieve? The network, specified by \Cref{eq: DenseAM energy}, can be defined for any number $K$ of memories. But, it turns out, if you pack too many of such vectors inside the $D$-dimensional discrete space, the local minima of the energy will no longer correspond to the stored patterns. In what follows we will compute the largest value of $K$ that permits successful remembering of the stored patterns. 

In general, this maximal value $K^\text{max}$ will depend on the specific choices for the stored memories. We will derive a statistical scaling law for this memory capacity assuming that the patterns are drawn at random from the following distribution:
\begin{equation}
    \xi^\mu_i = \begin{cases}
            +1, \ \ \text{with probability}\  \frac{1}{2} \\
            -1, \ \ \text{with probability}\  \frac{1}{2}
    \end{cases}
\end{equation}
With this distribution, it is easy to compute the correlation functions for these variables. The one-point and two-point correlation functions are given by: 
\begin{equation}
    \langle\ \xi^\mu_i\ \rangle = 0, \ \ \ \ \  \langle\ \xi^\mu_i\ \xi^\nu_j\ \rangle = \delta^{\mu \nu} \delta_{i j} \label{eq: 1 and 2 point correlatiors}
\end{equation}

In order to quantify the information storage capacity of this network we will use the following trick. We will initialize the network in the state corresponding to one of the memories, say $\xi^1_i$, and let it evolve in time according to the update rule. If the pattern $\xi^1_i$ corresponds to a local minimum, that state must be stable. In other words, the dynamics should not change that initial state. Mathematically, this means that 
\begin{equation}
\begin{split}
    \sigma_i^{(t+1)} & =\text{Sign}\bigg[ \xi^1_i\ f \Big(\sum\limits_{j\neq i}^D \xi^1_j\ \xi^1_j\Big) + \sum\limits_{\mu=2}^K \xi^\mu_i\ f \Big(\sum\limits_{j\neq i}^D \xi^\mu_j\ \xi^1_j\Big) \bigg]  \\
    & = \text{Sign}\bigg[ \underbrace{\xi^1_i\ f \Big(D-1\Big)}_{\text{\large signal}} \ +\ \underbrace{\sum\limits_{\mu=2}^K \xi^\mu_i\ f \Big(\sum\limits_{j\neq i}^D \xi^\mu_j\ \xi^1_j\Big)}_{\text{\large noise}} \bigg] \ \stackrel{\mathclap{\normalfont\mbox{?}}}{=}\ \xi^1_i 
    \label{eq:stability analysis}
\end{split}
\end{equation}

\begin{calloutInfo}[Derivation of the generating function]
It is helpful to introduce a new variable 
\begin{equation}
    \Xi = \sum\limits_{j=2}^D \xi^1_j
\end{equation}
and compute the generating function defined as a statistical average of the exponent of that variable
\begin{equation}
    M(\tau) = \langle\ e^{\tau\Xi}\ \rangle 
\end{equation}
Since $\xi^1_j$ are independent for different indices $j$, the statistical average can be factorized and computed explicitly 
\begin{equation}
    M(\tau) = \frac{1}{2^{D-1}} \sum\limits_{\xi_2=\pm 1}\sum\limits_{\xi_3=\pm 1} ... \sum\limits_{\xi_D=\pm 1} e^{\tau \xi_2} e^{\tau \xi_3} ... e^{\tau \xi_D} = \cosh(\tau)^{D-1}
\end{equation}
All correlation function can be computed by taking derivatives of the generating function. For instance
\begin{equation}
    \langle\ \Xi^{2p}\ \rangle = \frac{\partial^{2p}M}{\partial \tau^{2p}} \Bigg|_{\tau=0} = (2p-1)!! D^p
\end{equation}
\end{calloutInfo}

Assuming that the function $f(\cdot)$ is non-negative, the signal term pushes the argument of the Sign function towards aligning it with the desired pattern $\xi^1_i$. The noise term generally pushes that argument away from the desired pattern and in some situations may outweigh the signal term. Below, we will compute the characteristic magnitude of the noise term and determine when it becomes dominant and destroys the stability of the target memory. Specifically, we can compute the mean and variance of the noise term. The mean
\begin{equation}
    \langle\ \text{noise}\ \rangle = \bigg\langle\ \sum\limits_{\mu=2}^K \xi^\mu_i\ f \Big(\sum\limits_{j\neq i}^D \xi^\mu_j\ \xi^1_j\Big)\ \bigg\rangle = 0 
\end{equation}
is equal to zero since index $i$ appears only once in the correlator, see \Cref{eq: 1 and 2 point correlatiors}. The variance is given by
\begin{equation}
    \begin{split} 
    \langle\ \text{noise}^2\ \rangle & = \bigg\langle\ \sum\limits_{\mu=2}^K \xi^\mu_i\ f \Big(\sum\limits_{j\neq i}^D \xi^\mu_j\ \xi^1_j\Big)\ \sum\limits_{\lambda=2}^K \xi^\lambda_i\ f \Big(\sum\limits_{k\neq i}^D \xi^\lambda_k\ \xi^1_k\Big)\ \bigg\rangle \\
    & = \sum\limits_{\mu=2}^K \bigg\langle\ f \Big(\sum\limits_{j\neq i}^D \xi^\mu_j\ \xi^1_j\Big)\  f \Big(\sum\limits_{k\neq i}^D \xi^\mu_k\ \xi^1_k\Big)\ \bigg\rangle \ \ \stackrel{\mathclap{\normalfont\mbox{i.d.}}}{=} \ \ 
    \sum\limits_{\mu=2}^K \bigg\langle\ f \Big(\sum\limits_{j\neq i}^D \xi^\mu_j\Big)\  f \Big(\sum\limits_{k\neq i}^D \xi^\mu_k\ \Big)\ \bigg\rangle \\ 
    &=  (K-1) \Big\langle\ f \Big(\sum\limits_{j\neq i}^D \xi^\mu_j\Big)^2 \Big\rangle,
    \end{split}
\end{equation}
where we used that $\langle\ \xi^\mu_i\ \xi^\lambda_i\ \rangle = \delta^{\mu\lambda}$ and the property that in distribution $\xi^\mu_j \xi^1_j\ \stackrel{\mathclap{\normalfont\mbox{i.d.}}}{=}\ \xi^\mu_j$. 

Now, it is instructive to restrict our calculation to the class of power energy functions so that
\begin{equation}
    F(\cdot) = \frac{1}{n}(\cdot)^n, \ \ \ \ f(\cdot) = (\cdot)^{n-1},  \ \ \ \ \text{where} \ n \ \text{is an integer} .
\end{equation}
In this case, the variance of the noise can be computed exactly (through the generating function) and is equal to\footnote{We assume that $K$ is large so that $K-1\approx K$.} \cite{chaudhry2023long}: 
\begin{equation}
    \Sigma^2 = \langle\ \text{noise}^2\ \rangle = (2n-3)!! K D^{n-1} \, .
\end{equation}

\begin{figure}[ht]
    \centering
    \includegraphics[width=0.855\textwidth]{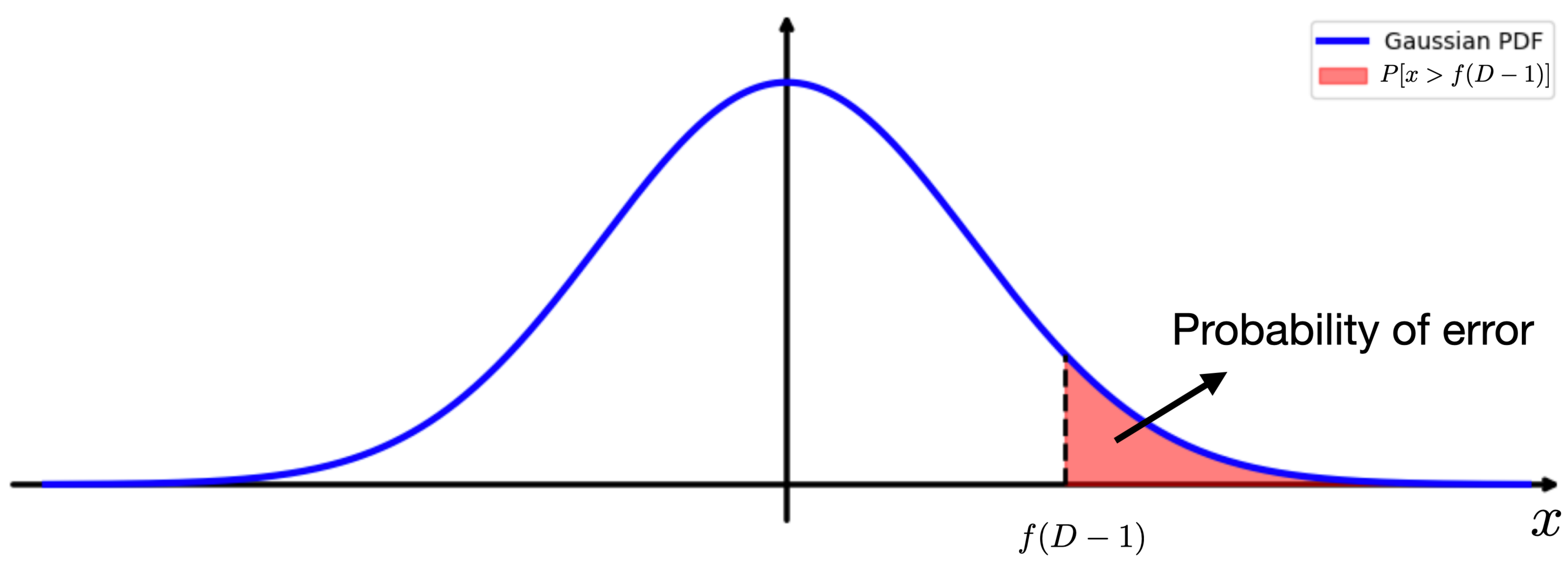}
    \caption{Gaussian probability distribution function. Shaded area indicates the probability of an error or spin flip.}
    \label{fig:info_capacity:Gaussian}
\end{figure}

Now we are ready to compute the probability of an error. The noise term in \Cref{eq:stability analysis} is a sum over many independent random variables. When $K$ and $D$ are large, this noise term behaves approximately as a Gaussian random variable. When the sign of the noise term is the same as the sign of the signal, the noise term pushes the update in the right direction and does not cause issues. The problem arises when the noise is large and its sign is opposite to that of the signal. In this situation, it is possible that the noise can outweigh the signal and flip the spin of interest. The probability of this event is given by the area under a Gaussian distribution, as shown in \Cref{fig:info_capacity:Gaussian}:
\begin{equation}
    P(\text{error}) = \int\limits_{f(D-1)}^\infty \frac{dx}{\sqrt{2\pi\Sigma^2}}e^{-\frac{x^2}{2\Sigma^2}} = \int\limits_{\frac{f(D-1)}{\Sigma}}^\infty \frac{dy}{\sqrt{2\pi}}e^{-\frac{y^2}{2}} = g\Big(\frac{f(D-1)}{\Sigma}\Big) < 1\% \, .
\end{equation}
Thus, if we want the probability of error be smaller than a certain value the following inequality must be satisfied :
\begin{equation}
    f(D-1)> \alpha \Sigma , 
\end{equation}
where $\alpha$ is a numerical constant independent of $K$, $D$, and $n$ (for $1\%$ error $\alpha\approx 2.576$). This translates into the following bound for the number of memories :
\begin{equation}
    K<K^{\max} = \frac{1}{\alpha^2 (2n-3)!!} D^{n-1} \, .
    \label{eq: capacity bound}
\end{equation}
Thus, as long as the number of memories is smaller than $K^{\max}$, the network initialized in one of the memories remains there and the dynamics does not flow away from it. It turns out, this is precisely the point when associative memory recall breaks. If the number of memories is smaller than $K^{\max}$ our network works as intended. Once $K$ exceeds $K^{\max}$, reliable recall breaks. This does not mean that the network becomes useless in that regime. In fact, it instead becomes a generative model. We will discuss this aspect later. 

\begin{calloutInfo}[What have we learned so far?] 
\begin{itemize}
    \item The number of memories $K$ is upper bounded.
    \item The Memory storage capacity heavily depends on the shape of the energy function $F(\cdot)$ and the shape of the activation function $f(\cdot)$.
    \item The sharper the energy peaks around memories -- the larger the memory storage capacity. 
\end{itemize}
\end{calloutInfo}

\section{Limiting Cases}
It is instructive to study a few limiting cases of the general family \Cref{eq: DenseAM energy}. Each of these models are frequently studied in the literature and have distinct properties.

{\bf The Hopfield Model $n=2$.} The simplest, and the most popular, example of the Dense Associative Memory is the Hopfield model. One can obtain it from the general form \Cref{eq: DenseAM energy} choosing the function as $F(\cdot) = \frac{1}{2}(\cdot)^2$. The energy function can be written as 
\begin{equation}
    E = - \frac{1}{2} \sum\limits_{\mu=1}^K \Big(\sum\limits_{i=1}^D \xi^\mu_i \sigma_i\Big)^2 = - \frac{1}{2} \sum\limits_{i,j=1}^D \sigma_i T_{ij} \sigma_j, \ \ \ \ \text{where} \ \ \ \ T_{ij} = \sum\limits_{\mu=1}^K \xi^\mu_i \xi^\mu_j  \, .
\end{equation}
In this case, according to the general result \Cref{eq: capacity bound}, the memory storage capacity scales linearly with the size of the network:
\begin{equation}
    K^{\max} \sim D \,.
\end{equation}
This is the famous $K^{\max} \approx 0.14 D$ scaling law from the Hopfield's 1982 paper \cite{hopfield1982neural}. Later, it was also derived by \cite{amit1985storing} using tools from statistical mechanics. While this model is appealing from the perspective of  mathematical  elegance and simplicity, this scaling law presents a major practical limitation. In the end, the hallmark of modern AI applications is the ability to store and process large amounts of information, a property severely limited by this scaling law. 

{\bf DenseAM with $n=3$.} Fortunately, this problem disappears for a more rapidly peaking energy function (obtained via an alternative activation function). For $F(\cdot) = \frac{1}{3}(\cdot)^3$, for example, the energy is given by 
\begin{equation}
    E = - \frac{1}{3} \sum\limits_{\mu=1}^K \Big(\sum\limits_{i=1}^D \xi^\mu_i \sigma_i\Big)^3 = - \frac{1}{3} \sum\limits_{i,j,k=1}^D  T_{ijk} \sigma_i \sigma_j \sigma_k, \ \ \ \ \text{where} \ \ \ \ T_{ijk} = \sum\limits_{\mu=1}^K \xi^\mu_i \xi^\mu_j \xi^\mu_k ,
\end{equation}
and the memory storage capacity scales as:
\begin{equation}
    K^{\max} \sim D^2 ,
\end{equation}
which is significantly faster than linearly. 

{\bf DenseAM with $F(\cdot) = \exp(\cdot)$.} It turns out that one can even achieve the exponentially large memory storage capacity. For exponential function $F(\cdot)$ \cite{demircigil2017model,lucibello2024exponential}, the number of memories that this DenseAM can store and retrieve scale as:  
\begin{equation}
    K^{\max} \sim 2^{\frac{D}{2}}, 
\end{equation}
which is more than sufficient for storing any practically relevant amount of information. Note, this number is the square root of the total number of binary states of the network (all possible binary vectors of length $D$). Despite its huge memory storage capacity, this model retains strong error correcting capabilities and has large size basins of attraction around each stored memories.

\section{General Dense Associative Memory with Binary State Variables}
Although simple models represented by \Cref{eq: DenseAM energy} illustrate the computational capabilities of Dense Associative Memories,  more general energy functions are also frequently studied. For binary DenseAM models, the general form of the energy function is given by
\begin{equation}
   E = - Q\Big[ \sum\limits_{\mu=1}^K F\Big( S\big[ \boldsymbol{\xi}^\mu, \boldsymbol{\sigma}\big]\Big) \Big],\label{DenseAM energy}
\end{equation}
where the function $F(\cdot)$ is a rapidly growing separation function (e.g., power $F(\cdot) = (\cdot)^n$ or exponent), $S[\boldsymbol{x}, \boldsymbol{x'}]$ is a similarity function (e.g., a dot product or a Euclidean distance), and $Q$ is a scalar monotone function (e.g., linear or logarithm). There are many possible combinations of various functions $F(\cdot), S(\cdot,\cdot)$, and $Q(\cdot)$ that lead to different models from the DenseAM family \cite{krotov2016dense,demircigil2017model,ramsauer2021hopfield,krotov2021large, millidge2022universal,burns2022simplicial}. We will discuss the relationship between these binary models and DenseAMs with continuous states in the next Chapter.

\begin{calloutNotebook}[Storage and recovery of memories in DenseAM] \label{nb:storage-recovery}
In this notebook, we offer the reader the possibility to experience storage and retrieval of patterns in DenseAM models. A set of simple Pokemon images can be embedded in the memory pool of the model. The model can then be queried by a corrupted version of a memory. The dynamical trajectory of the recall process retrieves the desired memory. By varying parameter $n$ the reader can experience both successful recovery of the memories and memory failures, when the recovered image does not correspond to the desired memory. All these numerical results are to illustrate the general theory discussed in this chapter.   

\nblinks{https://tutorial.amemory.net/tutorial/dense_storage.html}{https://github.com/bhoov/amtutorial/blob/main/tutorial_ipynbs/00_dense_storage.ipynb}{00_dense_storage}

\begin{center}
    \vspace{2em}
    \includegraphics[width=0.7\textwidth]{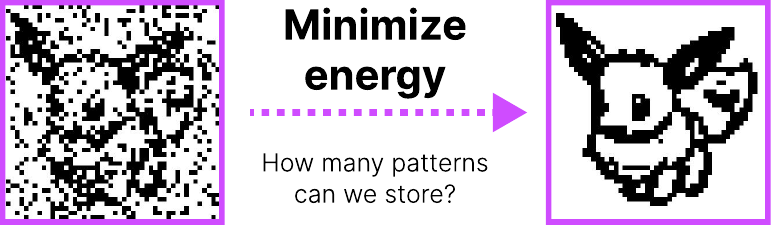}\\
    \vspace{1em}
\end{center}

\end{calloutNotebook}
\clearpage{}%
\clearpage{}%
\chapter{General Dense Associative Memory} \label{chap:am-blocks}

In the previous Chapter, we introduced Dense Associative Memories with discrete state vectors. While mathematically aesthetic and simple to analyze, such models do not allow backpropagation training --- due to their discrete nature. It turns out, most of the desired properties of  
Dense Associative Memories with discrete states are inherited by models with continuous variables. Moreover, the discrete state models can be derived as limiting cases of the continuous models.  

There are two other limitations of the models represented by \Cref{eq: DenseAM energy}. First, they do not have the hierarchical structure of representations, a crucial aspect which limits their ability to handle complex patterns from real-world datasets. Second, they have a rigid energy function --- although the energy depends on the learnable parameters $\xi^\mu_i$, its specific form may constrain the types of patterns and relationships the network can model.

In this Chapter, we introduce ``building blocks'' of Dense Associative Memories. Specifically, we develop a \textit{modular energy} perspective where the energy of any model from this family can be decomposed into standardized components: \textbf{neuron layers} that encode dynamic variables and \textbf{hypersynapses} that encode their interactions. \textit{The total energy of the system is the sum of each individual component energy.} This framework of energy-based building blocks for memory not only clarifies how existing methods relate to each other, but also provides a systematic language for designing new architectures. This abstraction is incredibly flexible and can be used to formulate all of the known models from this family, including Hierarchical Associative Memories \cite{krotov2021hierarchical},  Energy Transformers \cite{hoover2024energy}, neuron-astrocyte networks \cite{kozachkov2025neuron}, and many others. 

We refer to this generalized abstraction of Energy-based AMs as \textbf{\hamux}~\cite{hoover2022universal} after the software library that introduced it (here, {\hamux} stands for ``\textbf{H}ierarchical \textbf{A}ssociative \textbf{M}emory \textbf{U}ser e\textbf{X}perience''). We emphasize, however, that the abstraction is more fundamental than its specific software implementation. %

\section{Building Blocks of AMs with Modular Energies}

\begin{figure}[t]
    \centering
    \includegraphics[width=0.95\textwidth]{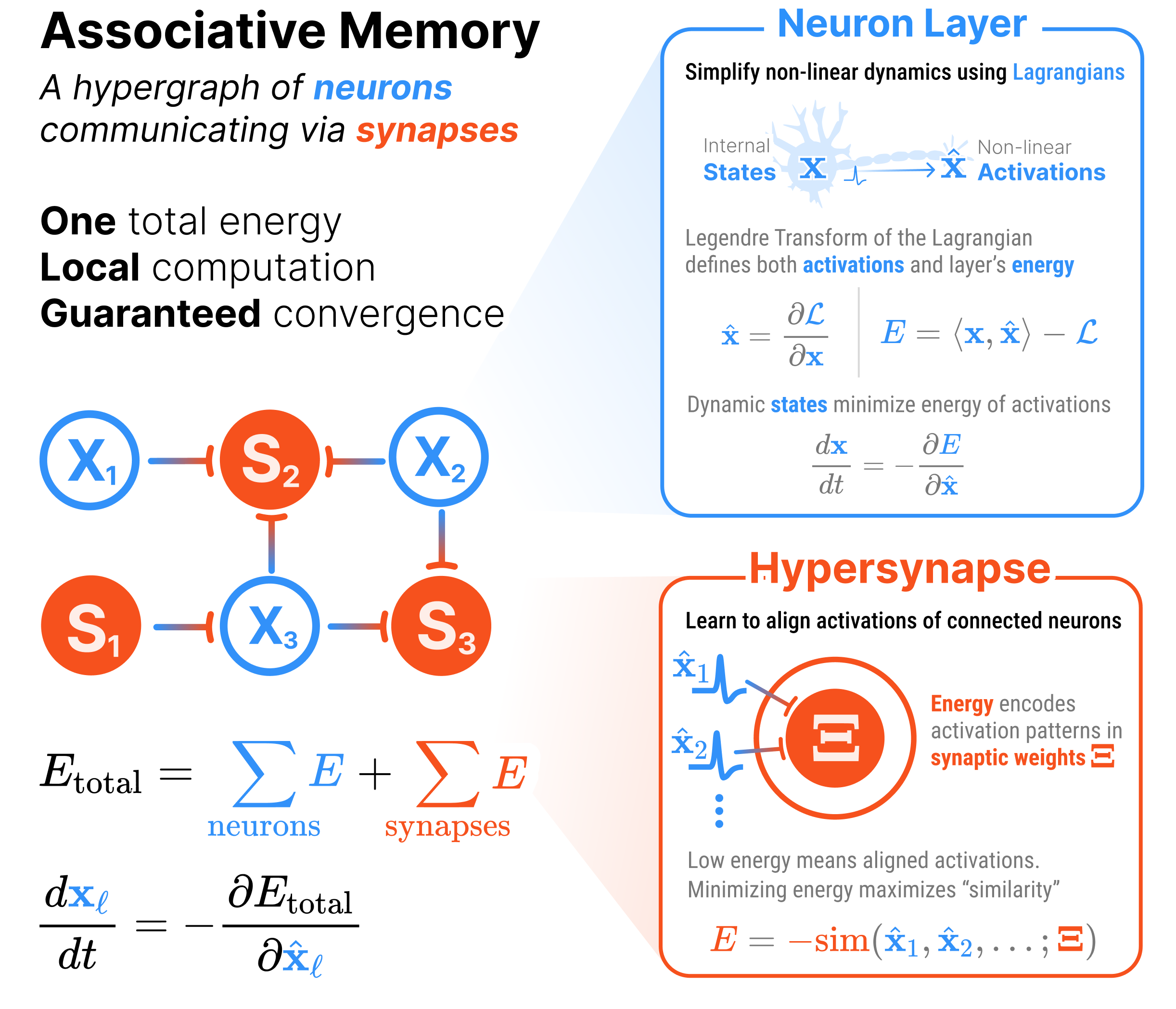}
    \caption{\textbf{HAMUX hypergraph diagrams are a graphical depiction of an AM} whose total energy is the sum of the {\color{hamux-blue}\textbf{neuron layer}} (node) and {\color{hamux-orange}\textbf{hypersynapse}} (hyperedge) energies. Inference is done recurrently, modeled by a system of differential equations where each neuron layer's hidden state updates to minimize the total energy. 
    When all non-linearities are captured in the dynamic neurons, inference becomes a local computation that avoids differentiating through non-linearities.}
    \label{fig:hamux-diagram}
\end{figure}

{\hamux} builds deep AMs by summing the modular energies of \textit{neuron layers} and \textit{hypersynapses}.

A \textbf{neuron layer} captures a non-linearity in the network (e.g., \texttt{ReLU}, \texttt{sigmoid}, \texttt{tanh}, \texttt{softmax}, \texttt{layernorm}, etc.). We call these non-linearities \textit{activations}, and their inputs or \textit{pre-activations} serve as the dynamic variables of the system. For example, a neuron layer can capture the computation $\hat{\rvx} = \texttt{ReLU}(\rvx)$, which has activations $\hat{\rvx}$ and pre-activations $\rvx$ which serve as the dynamic \textit{internal state} for this neuron layer. Structurally, neuron layers are the \textit{nodes} of our energy-based computation graph.

A \textbf{hypersynapse} is a parameterized energy function that captures how similar or \textit{aligned} the activations of its connected neuron layers are. For example, a simple hypersynapse may take the form $E_\text{Dense}(\hat{\rvx}, \hat{\rvy}; \rmXi) = -\hat{\rvx}^\intercal \rmXi \hat{\rvy}$, where $\rmXi$ is a synaptic weight matrix --- if we assume L2 normalized $\hat{\rvx}$ and $\hat{\rvy}$, minimizing $E_S$ w.r.t. the activations maximizes their cosine similarity or \textit{alignment} as modulated by the synaptic matrix $\rmXi$. The negative gradient of $E_\text{Dense}$ w.r.t. $\hat{\rvx}$ or $\hat{\rvy}$ looks like a \texttt{Dense} linear transformation, though more complex synaptic energies can be chosen to look like \texttt{Conv}, \texttt{Pooling}, or even \texttt{Attention} layers. In general, hypersynapses define the interactions between neurons and are the \textit{hyperedges} of our energy-based computation graph.

For a system of $L$ neuron layers and $S$ hypersynapses, the total energy of the system is

\begin{align}
    \label{eq:hamux-energy-dynamics}
    E_\text{total} = \sum_{\ell=1}^L E^\text{neuron}_\ell + \sum_{s=1}^S E^\text{synapse}_s.
\end{align}
The total energy is structured such that the activations of a neuron layer affect only connected hypersynapses and itself. Let $\hat{\rvx}_\ell$ and $\rvx_\ell$ represent the activations and internal states of neuron layer $\ell$, and let $\nbr{\ell}$ represent the set of hypersynapses that connect to neuron layer $\ell$. The following update rule describes how neuron internal states $\rvx_\ell$ minimize the total energy using only local signals
\begin{align}
    \label{eq:hamux-local-update}
    \tau_\ell\frac{d \rvx_\ell}{dt} = - \pp{E_\text{total}}{\hat{\rvx}_\ell} = - \left(\sum_{s \in \nbr{\ell}} \pp{E^\text{synapse}_s}{\hat{\rvx}_\ell}\right) - \pp{E^\text{neuron}_\ell}{\hat{\rvx}_\ell} = \mathcal{I}_{x_\ell} - \rvx_\ell, 
\end{align}
where $\mathcal{I}_{x_\ell} := - \sum_{s \in \nbr{\ell}} \nabla_{\hat{\rvx}_\ell} E^\text{synapse}_s$ is the \textit{total synaptic input current} to neuron layer $\ell$, which is fundamentally local and serves to minimize the energy of connected hypersynapses. See sections (\ref{sec:hamux-neurons}) and (\ref{sec:hamux-hypersynapses}) to understand the above equation in more detail. The time constant for neurons in layer $\ell$ is denoted by $\tau_\ell$.  
When the activations $\hat{\rvx}_\ell$ are bounded, the above system is guaranteed to converge for any choice of hypersynapse energies.

\section{Dynamical Neurons and their Lagrangians}
\label{sec:hamux-neurons}

A \textit{neuron layer} is a fancy term to describe the dynamic variables in AM and represent \textit{nodes} of the computational hypergraph. Each neuron layer has an \textit{internal state} $\rvx$ which evolves over time and an \textit{activation} $\hat{\rvx}$ that forwards a signal to the rest of the network. Think of neurons like the activation functions of standard neural networks, where $\rvx$ are the ``pre-activations'' and $\hat{\rvx}$ are the outputs e.g., $\hat{\rvx} = \texttt{ReLU}(\rvx)$.

In order to define neuron's layer energy, AMs employ two mathematical tools from physics: \textit{convex Lagrangian functions} and the \textit{Legendre transform}. For each neuron layer, we define a convex, scalar-valued Lagrangian $\mathcal{L}_x(\rvx)$. The Legendre transform $\mathcal{T}$ of this Lagrangian produces the dual variable $\hat{\rvx}$ (our activations) and the dual function $E_x(\hat{\rvx})$ (the neuron's energy) as in:
\begin{equation}
    \label{eq:neuron-legendre-transform}
    \begin{aligned}
    \hat{\rvx} &= \nabla \mathcal{L}_x(\rvx) \quad \text{(activation function)} \\
    E_x(\hat{\rvx}) = \mathcal{T}[\mathcal{L}_x] &= \langle \rvx, \hat{\rvx} \rangle - \mathcal{L}_x(\rvx) \quad \text{(dual energy)}
    \end{aligned}
\end{equation}
where $\langle \cdot, \cdot \rangle$ is the element-wise inner product. Because $\mathcal{L}_x$ is convex, the Jacobian of the activations $\frac{\partial \hat{\rvx}}{\partial \rvx} = \nabla^2 \mathcal{L}_x(\rvx)$ (i.e., the Hessian of the Lagrangian) is positive definite. This important point is summarized in \Cref{fig:hamux-diagram}.

The energy $E_x(\hat{\rvx})$ has another nice property --- \textit{its gradient equals the hidden states}. Thus, when we minimize the energy of our neurons (in the absence of any other signal), we observe exponential decay. This is nice to keep the dynamic behavior of our system bounded and well-behaved, especially for very large values of $\rvx$:

\begin{equation}
    \label{eq:exponential-decay}
    \frac{d \rvx}{dt} = - \nabla_{\hat{\rvx}} E_x(\hat{\rvx}) = - \rvx.
\end{equation}

\paragraph{Summary}
For non-physicists, the terminology used in this section can be daunting. The key insight is simple: a neuron layer is just a convex function $\mathcal{L}_x$ (the Lagrangian) applied to an internal state  $\rvx$. The Legendre transform of this Lagrangian then automatically provides two things: (1) the activation function $\hat{\rvx} = \nabla \mathcal{L}_x(\rvx)$, and (2) the dual energy representation $E_x(\hat{\rvx})$. This mathematical machinery abstracts away some of the complexity of non-linearities and gives us a simpler system to work with.

\begin{calloutProof}[Energy gradient equals hidden states]
    \textit{Show that $\pp{E_x(\hat{\rvx})}{\hat{\rvx}} = \rvx$.}
    \begin{align*}
        \pp{E_x(\hat{\rvx})}{\hat{\rvx}} &= \pp{}{\hat{\rvx}} \left(\langle \rvx, \hat{\rvx} \rangle - \mathcal{L}_x(\rvx)\right) \\
        &= \rvx + \hat{\rvx} \pp{\rvx}{\hat{\rvx}} - \pp{\mathcal{L}_x(\rvx)}{\rvx} \pp{\rvx}{\hat{\rvx}} \\
        &= \rvx + \hat{\rvx} \pp{\rvx}{\hat{\rvx}} - \hat{\rvx} \pp{\rvx}{\hat{\rvx}} \\
        &= \rvx
    \end{align*}
\end{calloutProof} %
\section{Hypersynapses}
\label{sec:hamux-hypersynapses}

The activations of one neuron layer are sent to other neurons via communication channels called \textit{hypersynapses}. At its most general, a hypersynapse is a scalar valued energy function defined on top of the activations of connected neuron layers. For example, a hypersynapse connecting neuron layers $\mathsf{X}$ and $\mathsf{Y}$ has an \textit{interaction energy} $E_{xy}(\hat{\rvx}, \hat{\rvy}; \rmXi)$, where $\rmXi$ represents the \textit{synaptic weights} or learnable parameters.
$E_{xy}(\hat{\rvx}, \hat{\rvy}; \rmXi)$ encodes the desired relationship between activations $\hat{\rvx}$ and $\hat{\rvy}$: when this energy is low, the activations satisfy the relationship encoded by the synaptic weights $\rmXi$. 
During energy minimization, the system adjusts the activations to reduce all energy terms, which means synapses effectively ``pull'' the connected neuron layers toward configurations encoded in the parameters that minimize their interaction energy. 

\begin{figure}[htbp]
    \centering
    \includegraphics[width=0.5\textwidth]{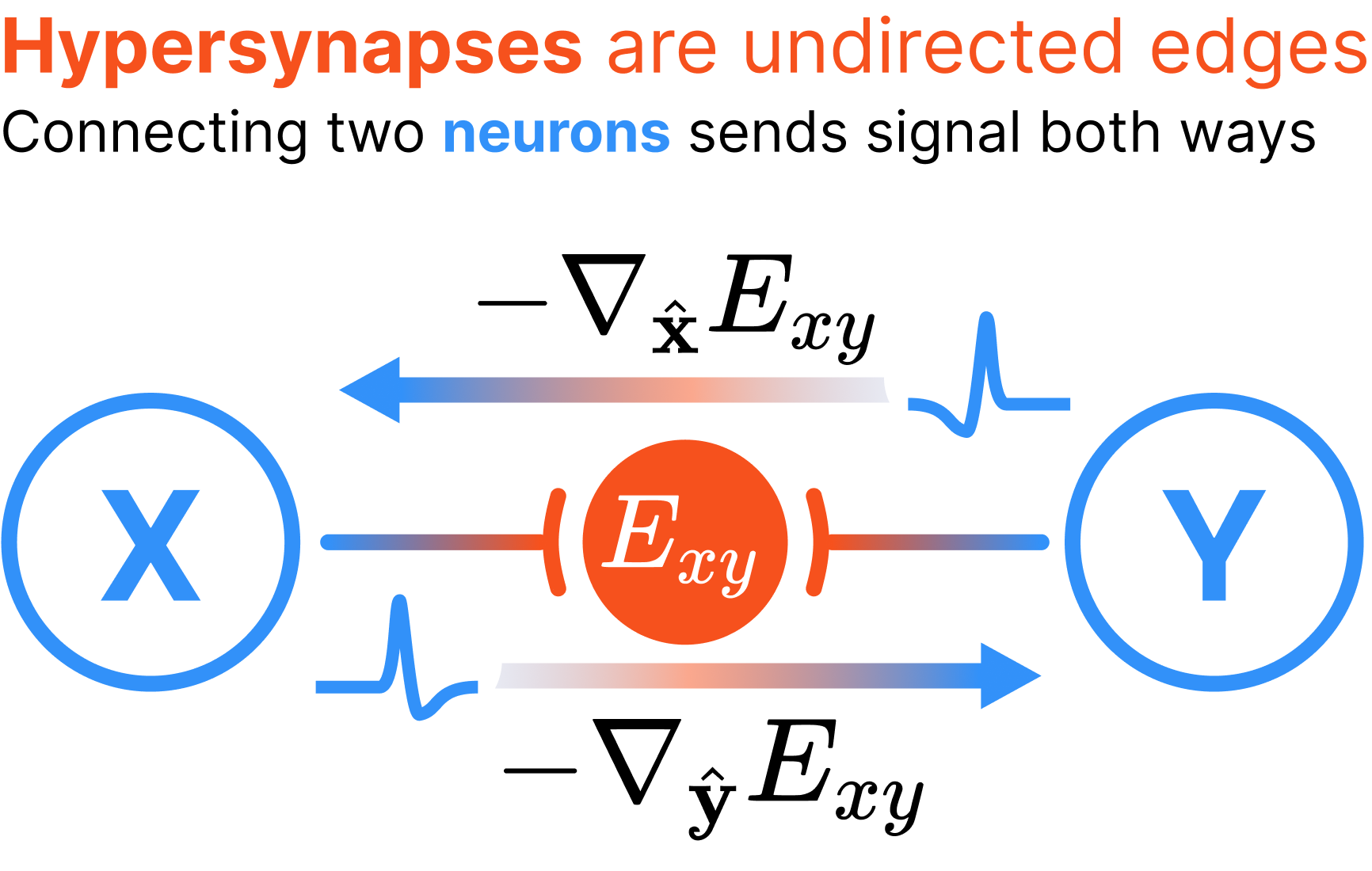}
    \caption{\textbf{Hypersynapses are represented as undirected (hyper)edges in a hypergraph.} Shown is an example pairwise synapse, which is a single energy function $E_{xy}(\hat{\rvx}, \hat{\rvy}; \rmXi)$ defined on the activations $\hat{\rvx}$ and $\hat{\rvy}$ from connected nodes, which necessarily propagates signal to both connected nodes. Here, \textit{signal} is defined as the negative gradient of the interaction energy {w.r.t.} the connected layer's activations (e.g., layer $\mathsf{X}$ receives signal $\mathcal{I}_x = -\nabla_{\hat{\rvx}} E_{xy}(\hat{\rvx}, \hat{\rvy}; \rmXi)$ while layer $\mathsf{Y}$ receives signal $\mathcal{I}_y = -\nabla_{\hat{\rvy}} E_{xy}(\hat{\rvx}, \hat{\rvy}; \rmXi)$). This is in contrast to biological synapses which are directional and only propagate signal in one direction from layer $\mathsf{X}$ to $\mathsf{Y}$, needing a separate synapse to bring information back from $\mathsf{Y}$ to $\mathsf{X}$.}
    \label{fig:undirected-synapse}
\end{figure}

Hypersynapses in the {\hamux} framework differ from biological synapses in two fundamental ways:
\begin{enumerate}
    \item \textbf{Hypersynapses can connect any number of layers simultaneously}, while biological synapses connect only two neurons. This officially makes each hypersynapse a \textit{hyperedge} in graph theory terms. 
    \item \textbf{Hypersynapses are undirected}, meaning that all connected layers influence each other bidirectionally during energy minimization. Meanwhile, biological synapses are unidirectional, meaning signal flows from a presynaptic to postsynaptic neuron. 
\end{enumerate}
Because of these differences, we choose the distinct term ``hypersynapses'' to distinguish them from biological synapses.

\begin{calloutInfo}[Hypersynapse notation conventions]
    For synapses connecting multiple layers, we subscript with the identifiers of all connected layers. For example:
    \begin{itemize}
        \item $E_{xy}$ --- synapse connecting layers $\mathsf{X}$ and $\mathsf{Y}$
        \item $E_{xyz}$ --- synapse connecting layers $\mathsf{X}$, $\mathsf{Y}$, and $\mathsf{Z}$.     
        \item $E_{xyz\ldots}$ --- synapses connecting more than three layers are possible, but rare.
    \end{itemize}

    However, synapses can also connect a layer to itself (self-connections). To avoid confusion with neuron layer energy $E_x$, we use curly brackets for synaptic self-connections. For example, $E_{\{x\}}$ represents the interaction energy of a synapse that connects layer $\mathsf{X}$ to itself.

    Because almost every interaction energy is parameterized in some way, we generally omit $\rmXi$ from the notation in subsequent sections when it's not central to the discussion
\end{calloutInfo}

The undirected nature of hypersynapses fundamentally distinguishes AM from traditional neural networks. Whereas feed-forward networks follow a directed computational graph with clear input-to-output flow, AMs have no inherent concept of ``forward'' or ``backward'' directions. All connected layers influence each other bidirectionally during energy minimization, with information propagating from deeper layers to shallower layers as readily as the other way around. See \Cref{fig:hamux-diagram} for a visual illustration.

Unlike the neuron layer's energies, the interaction energies of the hypersynapses are completely unconstrained: \textit{any function} that takes activations as input and returns a scalar is admissable and will have well-behaved dynamics\footnote{Some energies could be more meaningful than the others.}.
The interaction energy of a synapse may choose to introduce its own non-linearities beyond those handled by the neuron layers. When this occurs, the energy minimization dynamics must compute gradients through these ``synaptic non-linearities'', unlike the case where all non-linearities are abstracted into the neuron layer Lagrangians.

\section{Energy Descent Dynamics}
The central result is that dynamical equations \Cref{eq:hamux-local-update} decrease the global energy of the network \Cref{eq:hamux-energy-dynamics}. In order to demostrate this, consider the total time derivative of the energy 
\begin{equation}
    \frac{dE_\text{total}}{dt} = \sum\limits_{\ell=1}^L \frac{\partial E_\text{total}}{\partial \hat{\rvx}_\ell} \frac{\partial \hat{\rvx}_\ell}{\partial \rvx_\ell} \frac{d\rvx_\ell}{dt} = -\sum\limits_{\ell=1}^L \tau_\ell \frac{d \rvx_\ell }{dt} \frac{\partial^2 \mathcal{L}_x}{\partial \rvx_\ell \partial \rvx_\ell} \frac{d\rvx_\ell}{dt} \leq 0,
\end{equation}
where we expressed the partial of the energy w.r.t. the activations through the velocity of the neuron's internal states \Cref{eq:hamux-local-update}. The Hessian matrix $\frac{\partial^2 \mathcal{L}_x}{\partial \rvx_\ell \partial \rvx_\ell}$ has the size number of neurons in layer $\ell$ multiplies by the number of neurons in layer $\ell$. As long as this matrix is positive semi-definite, a property resulting from the convexity of the Lagrangian, the total energy of the network is guaranteed to either decrease or stay constant --- increase of the energy is not allowed. 

Additionally, if the energy of the network is bounded from below, the dynamics in \Cref{eq:hamux-local-update} are guaranteed to lead the trajectories to fixed manifolds corresponding to local minima of the energy. If the fixed manifolds have zero-dimension, they are fixed \textit{point attractors} and the velocity field will vanish once the network arrives at the local minimum. This correspondes to Hessians being strictly positive definite. Alternatively, if the Lagrangians have zero modes, resulting in existence of zero eigenvalues of the Hessian matrices, the network may converge to fixed manifolds of one or more dimensions, but the velocity fields may stay non-zero while the network's state moves along that manifold.

\section{Implementing AMs}
\label{sec:hamux-implementing}
We have established how the computational graph is built and the rules for how neuron layers and hypersynapses are constructed. We now discuss how the above mathematical framework can be used to recreate some of the commonly used AM models.

\begin{calloutExercise}[Designing the energy for a custom DenseAM]

\paragraph{Problem} Consider a DenseAM model consisting of $D$ neurons with the following activation function $\hat{x}_i = \tanh(\beta x_i)$. Design the synaptic energy and the global energy to recreate DenseAM with discrete variables discussed in Eqs. (\ref{eq: DenseAM energy}) and (\ref{eq: DenseAM update simplified}), in the limit $\beta\rightarrow\infty$.

\vspace{0.5cm}

\textbf{Solution}

First, define the Lagrangian for this network so that its partial gives the desired activation
\begin{equation}
    \mathcal{L} = \frac{1}{\beta} \sum\limits_{i=1}^D \log\Big(\cosh(\beta x_i)\Big), \ \ \ \ \ \text{resulting in }\ \ \ \ \hat{x}_i = \tanh(\beta x_i)
\end{equation}

The synapse connects neuron layer to itself and its synaptic energy is given by 
\begin{equation}
    E^\text{synapse} = -\sum\limits_{\mu=1}^K F\Big(\sum\limits_{j=1}^D \xi^\mu_j \hat{x}_j\Big)\label{eq: continuous DenseAM Energy}
\end{equation}
The total energy of the network is 
\begin{equation}
\begin{split}
    E^\text{total}& =  E^\text{neuron} + E^\text{synapse} =\Big[\sum\limits_{i=1}^D \hat{x}_i x_i - \mathcal{L}\Big]  -\sum\limits_{\mu=1}^K F\Big(\sum\limits_{j=1}^D \xi^\mu_j \hat{x}_j\Big) =\\ &
    = \sum\limits_{i=1}^D\Big[\tanh(\beta x_i)x_i - \frac{1}{\beta} \log\Big(\cosh(\beta x_i)\Big)\Big] - \sum\limits_{\mu=1}^K F\Big(\sum\limits_{j=1}^D \xi^\mu_j \hat{x}_j\Big)
\end{split}
\end{equation}

The dynamical update equation \Cref{eq:hamux-local-update} is given by

\begin{equation}
    \tau \frac{d x_i}{dt} = \sum\limits_{\mu=1}^K \xi^\mu_i \ f\Big(\sum\limits_{j=1}^D\xi^\mu_j \hat{x}_j\Big) - x_i ,
\end{equation}

where $f := F'$ is the derivative of the DenseAM's \textit{separation function} $F$. Now, let's discretize time. Set $\tau = 1$ and write the above equation in finite differences ($dt=1$). The result is 
\begin{equation}
    \frac{x^{t+1}_i - x^{t}_i}{dt} = \sum\limits_{\mu=1}^K \xi^\mu_i \ f\Big(\sum\limits_{j=1}^D \xi^\mu_j \hat{x}^{t}_j\Big) - x^t_i 
\end{equation}
which leads to 
\begin{equation}
    x^{t+1}_i = \sum\limits_{\mu=1}^K \xi^\mu_i \ f\Big(\sum\limits_{j=1}^D \xi^\mu_j \hat{x}^{t}_j\Big)\label{eq: finite difference limit}
\end{equation}
Finally, express everything through the activations $\hat{x}_i$ and take the limit $\beta\rightarrow\infty$. In this limit $\hat{x}_i = \text{Sign}(x_i) = \sigma_i$ and the energy of the layer vanishes, resulting in the total energy 
\begin{equation}
    E^\text{total} =  -\sum\limits_{\mu=1}^K F\Big(\sum\limits_{j=1}^D \xi^\mu_j \sigma_j^t\Big) 
\end{equation}
The discrete update equation can be obtained by acting with the $\text{Sign}(\cdot)$ function on both sides of \Cref{eq: finite difference limit}, resulting in \Cref{eq: DenseAM update simplified}. 
\end{calloutExercise}

\subsection{Energy Transformer Block}\label{sec:ET-block}
We now explain how the techniques developed above can be used for building the Energy Transformer (ET) architecture \cite{hoover2024energy}. For clarity of presentation, we use language associated with the image domain, although this architecture can also be used for language or graphs with minimal modifications. 

\begin{figure}[t]
 \begin{center}
\includegraphics[width=1.0\linewidth]{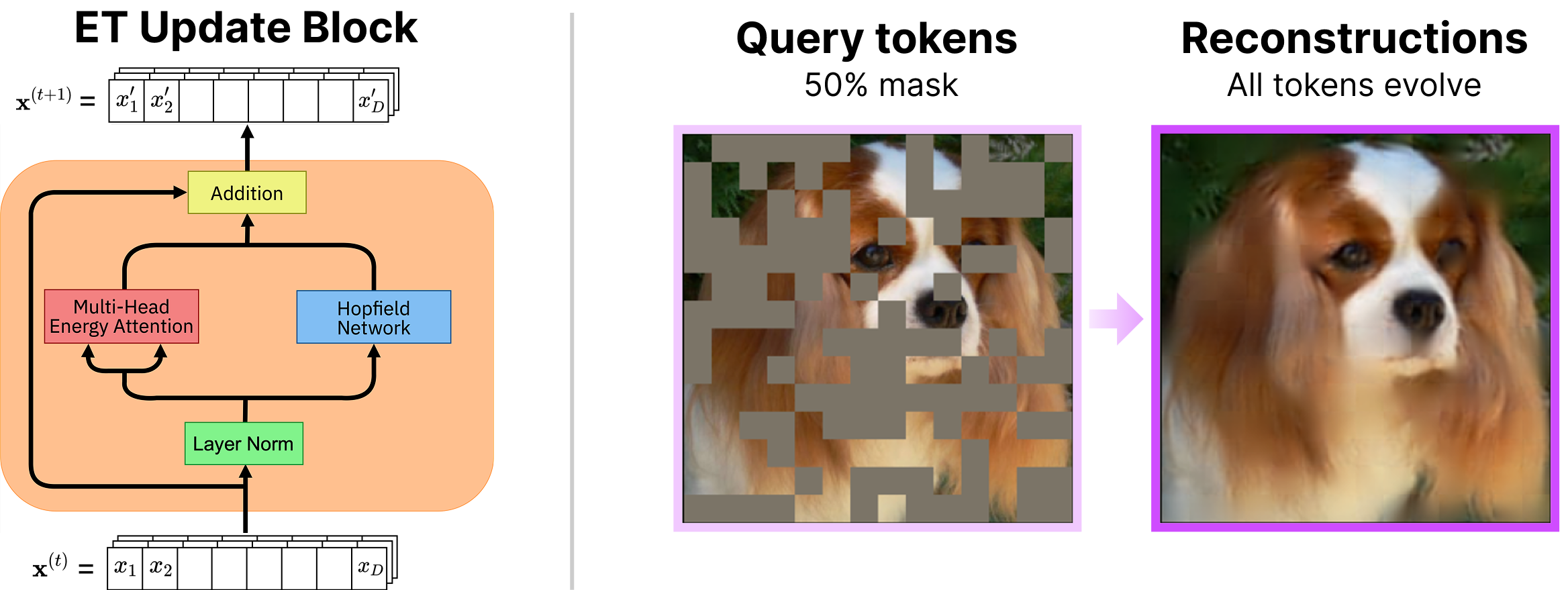}
\end{center}
\caption{Energy Transformer (ET) describes an energy-based Associative Memory whose gradient looks like a transformer block. \textbf{Left}: Inside the ``ET block'', the gradient of ET's energy. The input tokens $\mathbf{x}^{(t)}$ passes through a sequence of operations and gets updated to produce the output tokens $\mathbf{x}^{(t+1)}$. The operations inside the ET block are carefully engineered so that the entire network has a global energy function, which decreases with time and is bounded from below. In contrast to conventional transformers, the ET-based analogs of the attention module and the feed-forward MLP module are applied in parallel as opposed to consecutively. \textbf{Right}: The ET block recurrently minimizes the energy of a corrupted image represented by a collection of tokens, where 50\% of the tokens are occluded. Shown is an image of a dog not seen when training ET.} 
\label{fig:ET_block}
\end{figure}

The overall pipeline is similar to the Vision Transformer networks (ViTs) \cite{Dosovitskiy2021}. An input image is split into non-overlapping patches. After passing these patches through the encoder and adding the positional information, the semantic content of each patch and its position is encoded in the token $x_{iA}$. In the following the indices $i,j,k = 1,...,D$ are used to denote the token vector's elements, indices $A,B,C = 1,...,N$ are used to enumerate the patches and their corresponding tokens. It is helpful to think about each image patch as a physical particle, which has a complicated internal state described by a $D$-dimensional vector $\mathbf{x}_A$. This internal state describes the semantic and positional ``identity'' of the particle, simultaneously representing the patch $A$'s pixels, its contextual meaning as derived from other patches, and its positional embedding. 

The ET block is described by a continuous time differential equation, which describes interactions between these particles. Initially, at $t = 1$ the network is given a set containing two groups of particles corresponding to open and masked patches. The ``open'' particles know their identity and location in the image. The ``masked'' particles only know where in the image they are located, but are not provided the information about what image patch they represent. The goal of ET's non-linear dynamics is to allow the masked particles to find an identity consistent with their locations and the identities of open particles. This dynamical evolution is designed so that it minimizes a global energy function. The identities of the masked particles are considered to be revealed when the dynamical trajectory reaches the fixed point. Thus, the central question is: how can we design the energy function that accurately captures the task that the Energy Transformer needs to solve? 

The masked particles' search for identity is guided by two pieces of information: identities of the open particles, and the general knowledge about what patches are in principle possible in the space of all possible images. These two pieces of information are described by two contributions to the ET's energy function: the energy-based attention and the Hopfield Network. Below we define each element of the ET block in the order they appear in \Cref{fig:ET_block}.

\subsection*{Layer-Norm}\label{sec:layernorm}
Each token (particle) is represented by a vector $\rvx \in \mathbb{R}^D$. Most of the operations inside the ET block are defined on top of a layer-normalized version of token representation $\rvx$:
\begin{equation}
    \hat{x}_i = \gamma \frac{x_i - \Bar{x}}{\sqrt{\frac{1}{D}\sum\limits_j \big(x_j - \Bar{x}\big)^2 +\varepsilon}} + \delta_i, \ \ \ \ \ \text{where} \ \ \ \ \ \Bar{x} = \frac{1}{D}\sum\limits_{k=1}^D x_k
\end{equation}
The scalar $\gamma$ and the vector elements $\delta_i$ are learnable parameters, $\varepsilon$ is a small regularization constant. Following the general recipe of {\hamux}, this operation can be viewed as an activation function for the neural layer and can be derived as a partial derivative of the Lagrangian function: 
\begin{equation}
    \mathcal{L}(\rvx) = D\gamma\sqrt{\frac{1}{D}\sum\limits_j \big(x_j - \Bar{x}\big)^2 +\varepsilon} \ +\  \sum\limits_j \delta_j x_j, \ \ \ \ \ \text{so that} \ \ \ \ \ \hat{x}_i=\frac{\partial \mathcal{L}(\rvx)}{\partial x_i} \label{eq:lnorm-lagrangian}
\end{equation}
See \cite{krotov2021large,tang2021remark,krotov2021hierarchical} for the discussion of this property.

\subsection*{Multi-Head Energy Attention}
The first contribution to the ET's energy function is responsible for exchanging information between the tokens (particles). Similarly to the conventional attention mechanism, each token generates a pair of queries and keys (ET does not have a separate value matrix; instead the value matrix is a function of keys and queries). The goal of the energy-based attention is to evolve the tokens in such a way that the keys of the open patches are aligned with the queries of the masked patches in the internal space of the attention operation. Below we use index $\alpha = 1,...,Y$ to denote elements of this internal space, and index $h = 1,...,H$ to denote different heads of this operation. With these notations the energy-based attention operation is described by the following energy function: 
\begin{equation} \label{energy attention}
    E{^\text{ATT}} = -\frac{1}{\beta}\sum\limits_{h=1}^H\sum\limits_{C=1}^N \textrm{log} \left(\sum\limits_{B \neq C} \textrm{exp}\left(\beta A_{hBC} \right) \right)
\end{equation}
where the attention matrix $A_{hBC}$ is computed from query and key tensors as follows:
\begin{equation}
    \begin{split}
        A_{h B C} &= \sum\limits_{\alpha} K_{ \alpha h B} \; Q_{\alpha h C}, \ \ \ \ \ \ \mathbf{A} \in \R^{H \times N \times N} \\
        K_{\alpha h B} &= \sum\limits_j W^K_{\alpha h j}\; \hat{x}_{jB}, \ \ \ \ \ \ \ \ \ \ \ \mathbf{K} \in \R^{Y \times H \times N} \\
        Q_{\alpha h C} &= \sum\limits_j W^Q_{\alpha h j}\; \hat{x}_{jC}, \ \ \ \ \ \ \ \ \ \ \ \mathbf{Q} \in \R^{Y \times H \times N}
    \end{split}\label{notation explain}
\end{equation}
and the tensors $\mathbf{W}^K \in \R^{Y \times H \times D}$ and $\mathbf{W}^Q \in \R^{Y \times H \times D}$ are learnable parameters. From the perspective of {\hamux}, \Cref{energy attention} is the energy of the synapse, which mixes layers of neurons or tokens.

From the computational perspective each token generates two representations: a \textit{query} (given the position of the patch and its current content, where in the image should it look for information on how to evolve in time?), and a \textit{key} (given the current content of the patch and its position, what should be the contents of the patches that attend to it?). The log-sum energy function (\ref{energy attention}) is minimal when for every patch in the image its queries are aligned with the keys of a small number of other patches connected by the attention map (where higher values for tunable hyperparameter $\beta$ decreases the number of patches a query needs to align with). Different heads (index $h$) contribute to the energy additively.

\subsection*{Hopfield Network Module}
The next step of the ET block, which we call the Hopfield Network (HN), is responsible for ensuring that the token representations are consistent with what one expects to see in realistic images. The energy of this sub-block is defined as:
\begin{equation}\label{eq:energy-chn}
    E^{\text{HN}} = -\sum\limits_{B=1}^N\sum\limits_{\mu=1}^K G\Big(\sum\limits_{j=1}^D \xi_{\mu j} \; \hat{x}_{jB}\Big), \ \ \ \ \ \ \ \ \ \ \boldsymbol{\xi} \in \mathbb{R}^{K \times D} 
\end{equation}
where $\xi_{\mu j}$ is a set of learnable weights (memories in the Hopfield Network), and $G(\cdot)$ is an integral of the activation function $r(\cdot)$, so that $G(\cdot)^\prime = r(\cdot)$. This formula is identical to the energy \Cref{eq: continuous DenseAM Energy}. Depending on the choice of the activation function this step can be viewed either as a classical continuous Hopfield Network \cite{hopfield1984Neurons} if the activation function grows slowly (e.g., $r(\cdot)=\,$ \texttt{ReLU}),  or as a Dense Associative Memory \cite{krotov2016dense, krotov2021large} if the activation function is sharply peaked around the memories (e.g., $r(\cdot)=$ \texttt{power} or \texttt{softmax}). The HN sub-block is analogous to the feed-forward MLP step in the conventional transformer block but requires that the weights of the projection from the token space to the hidden neurons' space to be the same (transposed matrix) as the weights of the subsequent projection from the hidden space to the token space.
Thus, the HN module here is an MLP with shared weights that is {\em applied recurrently}. The energy contribution of this block is low when the token representations are aligned with some rows of the matrix $\mathbf{\xi}$, which represent memories, and high otherwise.\looseness=-1 

\subsection*{Dynamics of Token Updates}
The inference pass of the ET network is described by the continuous time differential equation, which minimizes the sum of the two energies described above. The whole ET network consists of layers of tokens coupled through two types of synapses, an attention synapse and a Hopfield Network synapse, so that 
\begin{equation}
\begin{split}
    E_\text{total} & = E^\text{neuron} + \sum_{s \in \{\text{ATT, HN} \}} E^\text{synapse}_s \\
    &= \Big[\sum\limits_{A=1}^N\sum\limits_{i=1}^D x_{iA} \hat{x}_{iA} - \sum\limits_{A=1}^N \mathcal{L}(\rvx_A)\Big] + E^\text{ATT} + E^\text{HN} \\ &
    \approx E^\text{ATT} + E^\text{HN} + O(\varepsilon)
\end{split}
\end{equation}
We work in the regime when the parameter $\varepsilon$ in the definition of the layer-norm Lagrangian is small --- it only serves as a regularization to prevent the division by zero. In this limit, neuron layer energy vanishes, and the total {\hamux} energy is the sum of $E^{\text{ATT}}$ and $E^{\text{HN}}$
\begin{equation}
\tau \frac{dx_{iA}}{dt} = -\frac{\partial E_\text{total}}{\partial \hat{x}_{iA}}, \ \ \ \ \ \text{where} \ \ \ \ \ E_\text{total} = E^{\text{ATT}} + E^{\text{HN}}  \label{eq:dynamical-equations}
\end{equation}
Here $x_{iA}$ is the token representation (input and output from the ET block), and $\hat{x}_{iA}$ is its layer-normalized version. The first energy is low when each patch's queries are aligned with the keys of its neighbors. The second energy is low when each patch has content consistent with the general expectations about what an image patch should look like (memory slots of the matrix $\mathbf{\xi}$). The dynamical system, represented by \Cref{eq:dynamical-equations}, finds a trade-off between these two desirable properties of each token's representation. For numerical evaluations, \Cref{eq:dynamical-equations} is discretized in time. 

To demonstrate that the dynamical system (\ref{eq:dynamical-equations}) minimizes the energy, consider the temporal derivative 
\begin{equation}
    \frac{dE_\text{total}}{dt} = \sum\limits_{i,j,A} \frac{\partial E_\text{total}}{\partial \hat{x}_{iA}}\ \frac{\partial \hat{x}_{iA}}{\partial x_{jA}}\ \frac{d x_{jA}}{dt} = - \frac{1}{\tau} \sum\limits_{i,j,A} \frac{\partial E_\text{total}}{\partial \hat{x}_{iA}}\ M^A_{ij}\ \frac{\partial E_\text{total}}{\partial \hat{x}_{jA}} \leq 0 
\end{equation}
The last inequality sign holds if the symmetric part of the matrix 
\begin{equation}
M^A_{ij} = \frac{\partial \hat{x}_{iA}}{\partial x_{jA}} = \frac{\partial^2 \mathcal{L}}{\partial x_{iA}\partial x_{jA}}
\end{equation}
is positive semi-definite (for each value of index $A$). The Lagrangian~(\ref{eq:lnorm-lagrangian}) satisfies this condition.

\begin{calloutNotebook}[Energy Transformer] 
In this notebook, we offer the reader the possibility to build the ET block in code following the general rules of {\hamux}. We have pre-trained this network on ImageNet and loaded the weights of the model, so that the reader can quickly play with parameters and visualize energy decent dynamics and learned representations at inference time. All these numerical results are to illustrate the general theory discussed in this chapter.   

\nblinks{https://tutorial.amemory.net/tutorial/energy_transformer.html}{https://github.com/bhoov/amtutorial/blob/main/tutorial_ipynbs/01_energy_transformer.ipynb}{01_energy_transformer}

\begin{center}
    \includegraphics[width=0.7\textwidth]{
    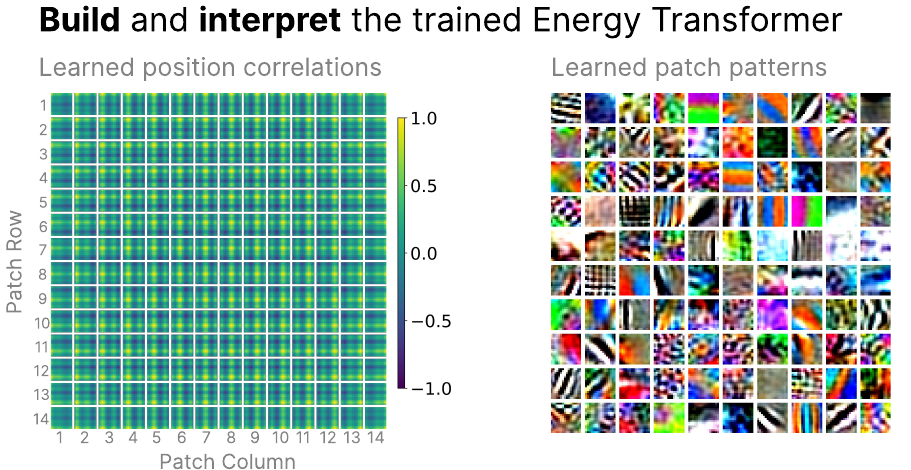
    }\\
    \vspace{1em}
\end{center}

\end{calloutNotebook}

\section{Bridging Energy Minimization and Feedforward Prediction} 
\label{sec:enmin-ff}
Although Associative Memories fundamentally differ from feedforward networks, they can both be used to solve the same tasks. Let $\vtheta$ describe the network parameters. Traditional feedforward networks transform input tensors $\rvx \in \mathcal{X}$ to output tensors $\rvy \in \mathcal{Y}$ via $f_\vtheta : \mathcal{X} \mapsto \mathcal{Y}$, where $\rvy^* = f_\vtheta(\rvx)$ represents the model's prediction. In contrast, an AM builds an energy-based computational graph that maps input tensors $\rvz \in \mathcal{Z}$ to a scalar energy value via $E_\vtheta : \mathcal{Z} \mapsto \mathbb{R}$. 
\begin{figure}[h]
    \centering
    \includegraphics[width=0.95\textwidth]{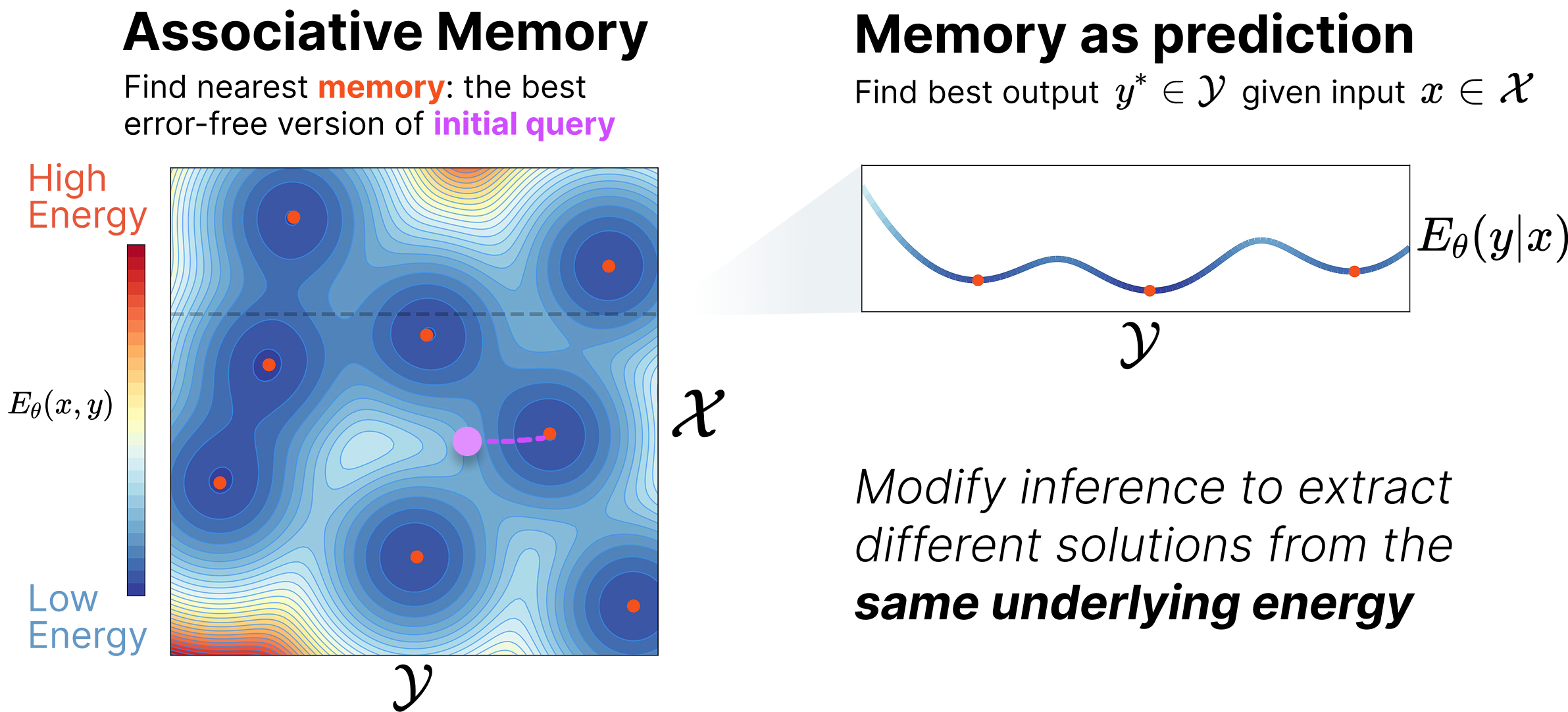}
    \caption{\textbf{Associative Memory is fully compatible with traditional prediction tasks}. By fixing a subset of variables (input) and minimizing the energy with respect to the remaining variables, we can predict optimal values for the output variables. Left: a 2D energy landscape used for the general Associative Memory task of cleaning an input. Right: a slice through the total energy landscape represents the energy objective for a prediction task.}
    \label{fig:ebm-landscape-prediction}
\end{figure}

How can we use an AM for prediction? When input space $\mathcal{X}$ and output space $\mathcal{Y}$ are distinct (e.g., in classification or segmentation), the energy function takes both spaces as input: $E_\vtheta : \mathcal{X} \times \mathcal{Y} \mapsto \mathbb{R}$. Given an input $\rvx \in \mathcal{X}$ for which we want to know the best output $\rvy^* \in \mathcal{Y}$ (as described in \Cref{eq:am-general-recall}), prediction or \textit{inference} becomes a coordinate-wise constrained energy minimization problem where we fix one of the variables and minimize the energy with respect to the other.

\begin{equation}
    \label{eq:am-general-recall}
    \rvy^* = \arg\min_{\rvy \in \mathcal{Y}} E_\vtheta(\rvx, \rvy)
\end{equation}

Sometimes the output space $\mathcal{Y}$ of a feedforward network represents a noiseless, inpainted version of $\mathcal{X}$, as in masked-token prediction where $\rvx^* = f_\vtheta(\rvx^{(0)})$. In this case, $f_\vtheta : \mathcal{X} \mapsto \mathcal{X}$ is doing a proxy of ``error minimization'' inside its computation graph (which likely has many residual connections), and the energy function is $E_\vtheta : \mathcal{X} \mapsto \mathbb{R}$.

The inference process of \Cref{eq:am-general-recall} is flexible and can be adapted for different prediction tasks. Let's view the \textit{optimization objective} of \Cref{eq:am-general-recall} as a higher order function $\mathcal{F}$ applied to energy function $E_\vtheta$. In this way, we can describe different input-output mappings from the energy. For instance, \Cref{eq:am-general-recall} is a mapping $\mathcal{F}_{x \shortto y}[E_\vtheta] : \mathcal{X} \mapsto \mathcal{Y}$ that performs a global search of $E_\vtheta$ over possible $\rvy$ given a clamped $\rvx$. We can define other useful inference objectives. For example, we can invert the search to instead search over $\rvx$ in the region of some initial guess $\rvx^{(0)}$ given a clamped $\rvy$. This would be represented via $\mathcal{F}_{(x,y) \shortto x}[E_\vtheta] : \mathcal{X} \times \mathcal{Y} \mapsto \mathcal{X}$. Or we could jointly optimize both variables given only an initial guess for both, as in $\mathcal{F}_{(x,y)\shortto (x,y)}[E_\vtheta] : \mathcal{X} \times \mathcal{Y} \mapsto \mathcal{X} \times \mathcal{Y}$. The Hopfield Network generally considers this last scenario, but each inference process represents a different way to extract solutions from the same underlying energy landscape.

\begin{calloutInfo}[Energy function vs. Loss function]
    The key idea in AMs is that \textbf{every computation serves to optimize some objective}. However, we choose to distinguish between two types of objectives: the \textit{energy} function and the \textit{loss} function. We say the energy function governs the dynamics of neuron states during \textit{inference}, while the loss function governs the dynamics of model parameters during \textit{training}. The primary difference is in whether the gradient is taken with respect to the states of the network, or the parameters.
\end{calloutInfo}

\clearpage{}%
\clearpage{}%
\chapter{Failure of Memory and Generative AI} 
\label{chap:am-ebm}

In 1977, psychologists Roger Brown and James Kulik described a famous experiment, in which respondents were asked to self-report the circumstances in which they found out about the highly surprising and consequential news of the President John F. Kennedy assassination \cite{brown1977flashbulb}. Among many insightful findings from this study, peculiar responses have been recorded containing detailed, emotional, highly realistic, and convincing descriptions of learning about this news for the first time that were factually inaccurate. For instance, one of the respondents (person A) vividly describes how person B came down the stairs to the first floor of the house, while person A was focusing on work and told person A that she heard about the assassination on the news. This recollection is detailed enough to include specific phrases and portions of the conversation between persons A and B during this recalled event. Although both persons are well familiar with each other and their recollections are plausible, documented evidence suggests that person A and person B could not be present in the same location after the JFK's assassination \cite{linton1979remember,loftus1988memory}. 

This is an example of misremembering, a phenomenon that is general and can be observed in many other situations. For instance, during a crime investigation two eyewitnesses can give mutually contradictory accounts of what they saw. Sometimes, both accounts can be different from what has actually happened. These examples of misremembering highlight a failure mode of human memory in which multiple observed events (training data) blend together and form novel memories, which are different from any of the observed events (training data points). Misremembering leads to creation of novel hypothetical memories, which in certain aspects share a degree of similarity (correlated) to the training data, but are distinct from individual training instances. Thus, misremembering can lead to creativity. 

In generative AI, \textit{creativity} is a key objective. For instance, diffusion models, being trained on a sufficiently large set of training images, can generate genuinely novel photorealistic images of previously unseen events \cite{sohl-dickstein15, ho2020denoising, song2020denoising, song2021scorebased}. A typical diffusion model training pipeline contains of two phases: the forward process when the noise is injected into the training samples, and the reverse process when a neural network is used to predict how much noise should be removed from the noisy sample with the goal to reconstruct the original uncorrupted training data point. At runtime, random noise is passed through the reverse process and converted into generated samples. The training pipeline of diffusion models can be conceptualized as the process of writing the training data into a memory network \cite{hoover2023memory,ambrogioni2023search}. By doing so, the information about training samples is written into the synaptic weights of the neural network that is used for denoising. The reverse process can be conceptualized as an attempt of memory recall --- that memorized information should be retrieved from the synaptic weights and turned into a generated sample. It is well established that the memory recall from the diffusion models can be successful; in that case the generated sample matches exactly at least one of the training samples. It can also be unsuccessful; in that case the generated sample will be novel and will not match any of the training samples. This is what is called the memorization-to-generalization transition in diffusion models \cite{carlini2023extracting, somepalli2023diffusion,pham2025memorization}, which occurs when the size of the training set is increased. Successful memory retrieval is typically viewed as a negative outcome in diffusion models and can often lead to privacy and copyright violations. Similarly, in LLMs training samples can often be extracted verbatim from the synaptic weights of the neural network, a property that has been a subject of intense discussions in the research community and public discourse \cite{carlini2023extracting}. At the same time, LLMs can also generate novel previously unseen responses. Importantly for us, in all these examples \textit{creativity arises as a result of a failure of memory recall}.

In Energy-based Associative Memory networks, memories are identified as local minima of an energy landscape, and the process of memory recall is conceptualized as a dynamical trajectory starting at a high energy state (corrupted memory) and leading to the best matching local minimum (recovered memory). Misremembering arises when the recovered memory (local minimum) of the energy function is different from any of the training data. These misremembered local minima are called \textit{spurious states}, see the below \cref{fig:general-energy-transition} for their illustration. In AM literature, they are typically viewed as an obstacle to the faithful memory recall. For this reason, researchers in this field typically aim to either remove them entirely from the energy landscape, or mitigate their contribution to computation (e.g., by raising their energy) \cite{Hopfield1983UnlearningHA, hertz1991introduction}. 

In this Chapter, we discuss the emergence of \textit{spurious states} in Dense Associative Memories, and the general relationship between these memory models and diffusion models, popular in generative AI. In previous chapters, AM models were studied in two situations. First, when the models have small memory capacity and are trained on a small amount of data, e.g., classical Hopfield Networks. Second, in situations when the models have large memory storage capacity, but still are trained on a small amount of data. The focus of this Chapter are settings in which the models are big (large information storage capabilities), and the amount of training data is even bigger (exceeds the critical memory storage capacity of the model). In this regime, DenseAMs turn into generative AI models.

\section{Diffusion Models}

Diffusion models have recently gained popularity, due to their flexibility and accuracy in modeling high-dimensional distributions for a variety of domains, including image generation \citep{ho2020denoising,  song2021scorebased, song2020denoising}, audio \citep{chen2020wavegrad, kong2020diffwave, liu2023audioldm}, video synthesis \citep{ho2022video, singer2022make, blattmann2023videoldm, videoworldsimulators2024}, and other scientific applications.  However, these powerful and flexible models pose great challenges related to privacy and security, as concerns grow about their tendency to generate their training data \cite{somepalli2023diffusion, somepalli2023understanding, carlini2023extracting}. Such matters consequently emphasize the need for further understanding of memorization and generalization behaviors in diffusion models.

\begin{figure}[t]
    \centering
    \includegraphics[keepaspectratio, width=1.0\textwidth, height=1\textheight]{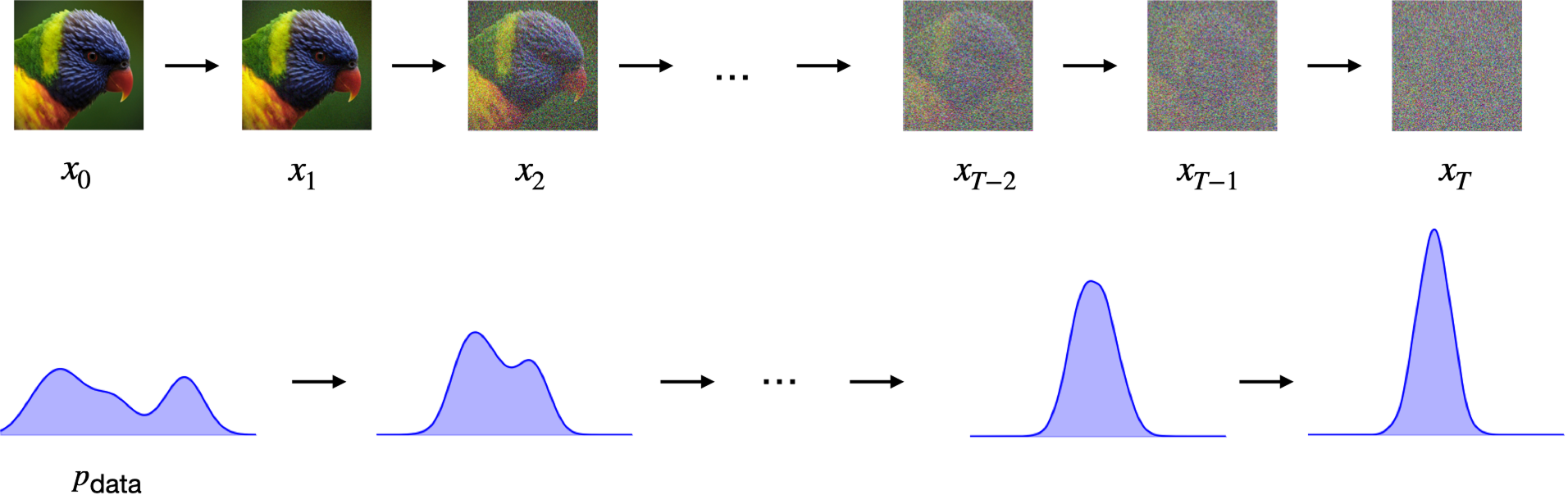}
    \caption{A general illustration of diffusion models. Addition of noise transforms the complex data distribution into a simple distribution --- an isotropic Gaussian. The reverse process removes the noise and transforms a noise sample into a sample from the data distribution.}
    \label{fig:diffusion-simple-illustration}
\end{figure}

There are two fundamental processes which govern the aspects of diffusion models. Firstly, the \textit{forward process} typically described by the following Itô Stochastic Differential Equation (SDE) \citep{song2021scorebased}: 
\begin{equation}
    \mathrm{d}\rvx_t = \mathbf{f}(\rvx_t, t) \mathrm{d}t + {g}(t) \mathrm{d} \rvw_t,
     \label{eq:forward_sde} 
\end{equation}
transforms the given data distribution ($\rvx_{0} = \rvy$) into a simpler distribution\footnote{Assume that $\rvx_0 = \rvy \in \mathbb{R}^D$ are i.i.d samples coming from a data distribution $p(\rvy)$.}, e.g., an isotropic Gaussian distribution. Here, $\rvw_t$ is the standard Wiener process (or Brownian motion) and $\mathbf{f}(\rvx_t, t)$ denotes the drift term that guides the diffusion process, which we will assume to be zero for the most part of this section. Meanwhile, $g(t)$ represents the diffusion coefficient that controls the noise at each time step $t \rightarrow T$. Secondly, the \textit{reverse process} removes the injected noise at each step $t$ and it is described as 
\begin{align}
    \mathrm{d} \rvx_t = [\mathbf{f}(\rvx_t, t) - g(t)^2 \nabla_{\rvx_t} \log p_t (\rvx_t)] \mathrm{d}t + g(t) \mathrm{d} \bar{\rvw}_t,
    \label{eq:backward_sde}
\end{align}
where $\bar{\rvw}_t$ is the standard Wiener process. To effectively solve \Cref{eq:backward_sde}, one must reliably estimate the score $\nabla_{\rvx} \log p_t(\rvx)$ via training a neural network $s_\theta(\rvx, t)$. The learned weights $\theta^*$ are obtained using methods for denoising score matching across multiple times steps \cite{sohl-dickstein15, ho2020denoising, song2021scorebased}. The general description of this optimization problem, given by \cite{song2021scorebased}, is formulated as
\begin{align}
     \theta^* = \argmin_\theta \underset{t, \rvy, \rvx_t}{\mathbb{E}} \big{[} \lambda(t) \, \lVert s_\theta(\rvx_t, t) - \nabla_{\rvx_t} \log p_{t}(\rvx_t | \rvy) \rVert^2 \big{]},
    \label{eq: score function definition}
\end{align}
where $t \sim \mathcal{U}(t_0, T)$  is sampled from the uniform distribution $\mathcal{U}$ over the set of times ranging from a small time $t_0 \approx 0$ to a larger time $T$, while $\rvy \sim p(\rvy)$ and $\rvx_t \sim p(\rvx_t | \rvy)$. Here, $p(\rvx_t | \rvy)$ is the forward process and $\lambda(t)$ is a positive weighting function.

\section{Diffusion Models from Associative Memories}

A fascinating aspect about diffusion models is their process of shaping their energy landscape. Instead of directly learning their energy function $E_\theta (\rvx_t, t)$, diffusion models learn the negative gradient of their energy function or the score function:
\begin{equation}
    \nabla_{\rvx_t} \log p_t(\rvx_t) = - \nabla_{\rvx_t} E_\theta (\rvx_t, t),
\end{equation}
using the process denoted in \Cref{eq: score function definition}. However, this particular process does not explain how generalization happens in such models. Instead, it tells a story of diffusion models behaving like AM systems. Specifically, during training, diffusion models are learning how to remove noise from a perturbed memory cue (or query) to obtain a clean memory accordingly to \Cref{eq: score function definition}. At some point, these models must behave like AM systems where they can effectively recover memories (or stored training data points) from noises. But once a certain threshold (memorization capacity) is exceeded, diffusion models can no longer act like effective denoisers or AM systems, the successive failure in memory recall of these models must facilitate and signal their transition to generative modeling.

Consequently, following the derivation done in \cite{pham2025memorization}, we can establish a fundamental connection between diffusion and AM models. Consider the training data distribution in the variance-exploding (VE) setting of $f(\rvx_t, t) = 0$ and $g(t) = \sigma$. In this case, the marginal probability distribution of new samples can be computed exactly as
\begin{equation}
    p(\rvx_t, t) = \underset{\rvy \sim \text{data}}{\mathbb{E}} \Bigg[\frac{1}{(2\pi\sigma^2 t)^{\frac{D}{2}}}\exp\Big(- \frac{\lVert \rvx_t - \mathbf{y} \rVert^2_2}{2 \sigma^2 t}\Big) \Bigg] .
\end{equation}
Assuming the empirical distribution of the data $p(\mathbf{y}) = \frac{1}{K}\sum\limits_{\mu=1}^K \delta^{(D)}(\mathbf{y} - \boldsymbol{\xi}^\mu)$, where $\boldsymbol{\xi}^\mu$ represents an individual data point (with data size $K$), this marginal distribution can be written as 
\begin{equation}
    p(\rvx_t, t) = \frac{1}{K} \sum \limits_{\mu=1}^K \frac{1}{(2\pi\sigma^2 t)^{\frac{D}{2}}}  \exp\Big({ - \frac{\lVert \rvx_t - \boldsymbol{\xi}^\mu \rVert^2_2}{2 \sigma^2 t}}\Big)
    \overset{\text{def}}{\equiv} \exp\Bigg (- {\frac{E^\text{DM}(\rvx_t, t)}{2 \sigma^2 t}}\Bigg ),
    \label{eq:marginal}
\end{equation}
where we also defined the energy $E^\mathrm{DM}$ of diffusion model, which up to state-- or $\rvx$-- independent terms is equal to 
\begin{equation}
    E^\text{DM}(\rvx_t, t) = -2 \sigma^2 t \log\bigg[\sum\limits_{\mu=1}^K \exp \Big(- \frac{\lVert \rvx_t - \boldsymbol{\xi}^\mu \rVert^2_2}{2 \sigma^2 t}\Big) \bigg] .
    \label{eq:energy-diffusion}
\end{equation}
As already observed in \cite{ambrogioni2023search}, the above energy function (\ref{eq:energy-diffusion}) is closely related to that of DenseAMs, which are large memory storage variants of classical Hopfield networks, see \Cref{chap:DenseAM}.

The core idea behind DenseAMs is to design an energy function that peaks very sharply around the intended memory patterns to prevent the overlapping (or cross talk) between them. Hence, such networks can store and retrieve a much larger number of patterns, compared to the classical Hopfield networks, and scales super-linearly (and possibly exponentially) to the size of the network, allowing the decoupling of  information storage capacity from the dimensionality of the data \cite{krotov2016dense, demircigil2017model}. Of particular interest here is the DenseAM model studied in \cite{saha23end} (see also \cite{millidge2022universal}), which bears strong resemblance to \Cref{eq:energy-diffusion}:
\begin{align}
    E^\text{AM}(\rvx) = -\beta^{-1} \log\bigg[\sum\limits_{\mu=1}^K \exp\Big(- \beta \lVert \rvx - \boldsymbol{\xi}^\mu \rVert^2_2\Big) \bigg], 
    \label{eq:DenseAM}
\end{align}
where $\beta$ is the inverse ``temperature'', which controls the steepness of the energy landscape around the memories $\boldsymbol{\xi}^\mu$. 

\begin{figure}[t]
    \centering
    \includegraphics[keepaspectratio, width=1.0\textwidth, height=1\textheight]{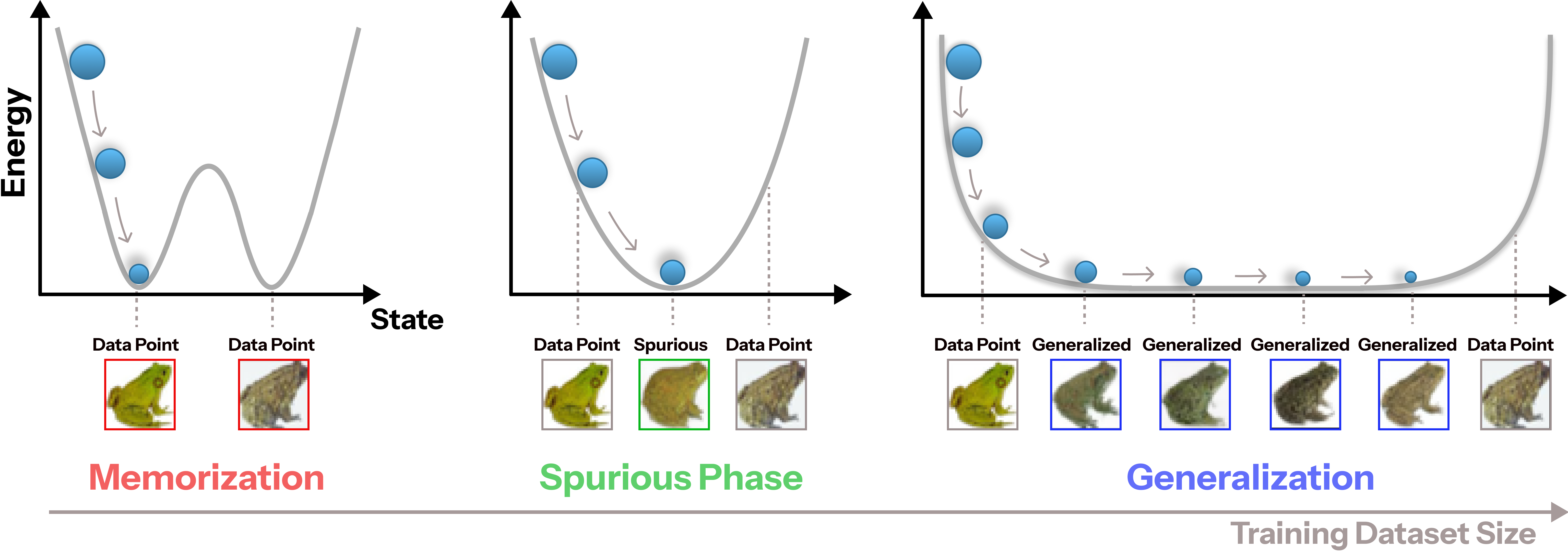}
    \caption{A simple illustration depicting the change in the energy landscape as the size of the training dataset is increased. In the small data regime, the diffusion model memorizes the training data points as local minima of the energy. When the amount of training data exceeds the memory capacity of the model, spurious patterns are formed and training data points are no longer energy minima. Subsequent increase of the training set size leads to the generalization phase, which is defined by the formation of continuous manifold of the low energy states. Figure is obtained from \cite{pham2025memorization}.}
    \label{fig:general-energy-transition}
\end{figure}

Notice that Eqs. (\ref{eq:energy-diffusion}) and (\ref{eq:DenseAM}) are identical if $\beta = 1/(2\sigma^2t)$. The two systems described above have important differences and similarities. In typical AM tasks, the inverse temperature $\beta$ is kept constant (and is typically large). At the same time, the diffusion energy $E^\text{DM}(\rvx_t, t)$ describes an intrinsically non-equilibrium system, since the effective temperature explicitly depends on time. However, notice that since the reverse process (\ref{eq:backward_sde}) is guaranteed to invert the forward step (\ref{eq:forward_sde}), the fixed points of the denoising process are guaranteed to coincide with the original data points. Specifically, both of the above energy functions, Eqs. (\ref{eq:energy-diffusion}) and (\ref{eq:DenseAM}), express a competition among the stored data points $\rvy \sim p(\rvy)$ to see which one is closer to the query $\rvx_t$ at time $t$. Hence, although there might be differences in dynamical trajectories for $E^\text{DM}$ and $E^\text{AM}$, their fixed points must be the same\footnote{We remind the reader that the fixed points retrieved from the reverse process correspond to $t = 0$.}.

Specifically, the manifolds of the data in diffusion models must emerge from the point-like memory storing systems, like AMs, in the limit when they are overloaded with amount of data above the critical memory capacity. In this regime, distinct basins of attraction corresponding to separate memories merge, forming the manifolds of the data. At the boundary of this transition, a separate ``phase'' corresponding to spurious states, which is ubiquitous in AMs around the \textit{critical memory load}, appears and signals the onset of generalization, see \Cref{fig:general-energy-transition} for the simple illustration of this memorization-generalization transition. It is worth noting that DenseAMs typically have an exponentially large memory storage capacity (in the number of neurons $D$) for uncorrelated patterns. However, in the cases of real data, due to the high correlation of samples, the \textit{critical memory load} is much lower than the exponentially large capacity of uncorrelated data --- a well-known fact in associative memories  \cite{cortes1987hierarchical, gutfreund1988neural, krogh1988mean, kanter1987associative, van1997hebbian,cowsikdense}.

\section{Memorization - Spurious - Generalization Transition}
\begin{figure}[!t]
    \centering  
    \setlength{\abovecaptionskip}{10pt}
    \setlength{\belowcaptionskip}{0pt} 
    \centering
    \includegraphics[width=1.025\linewidth]{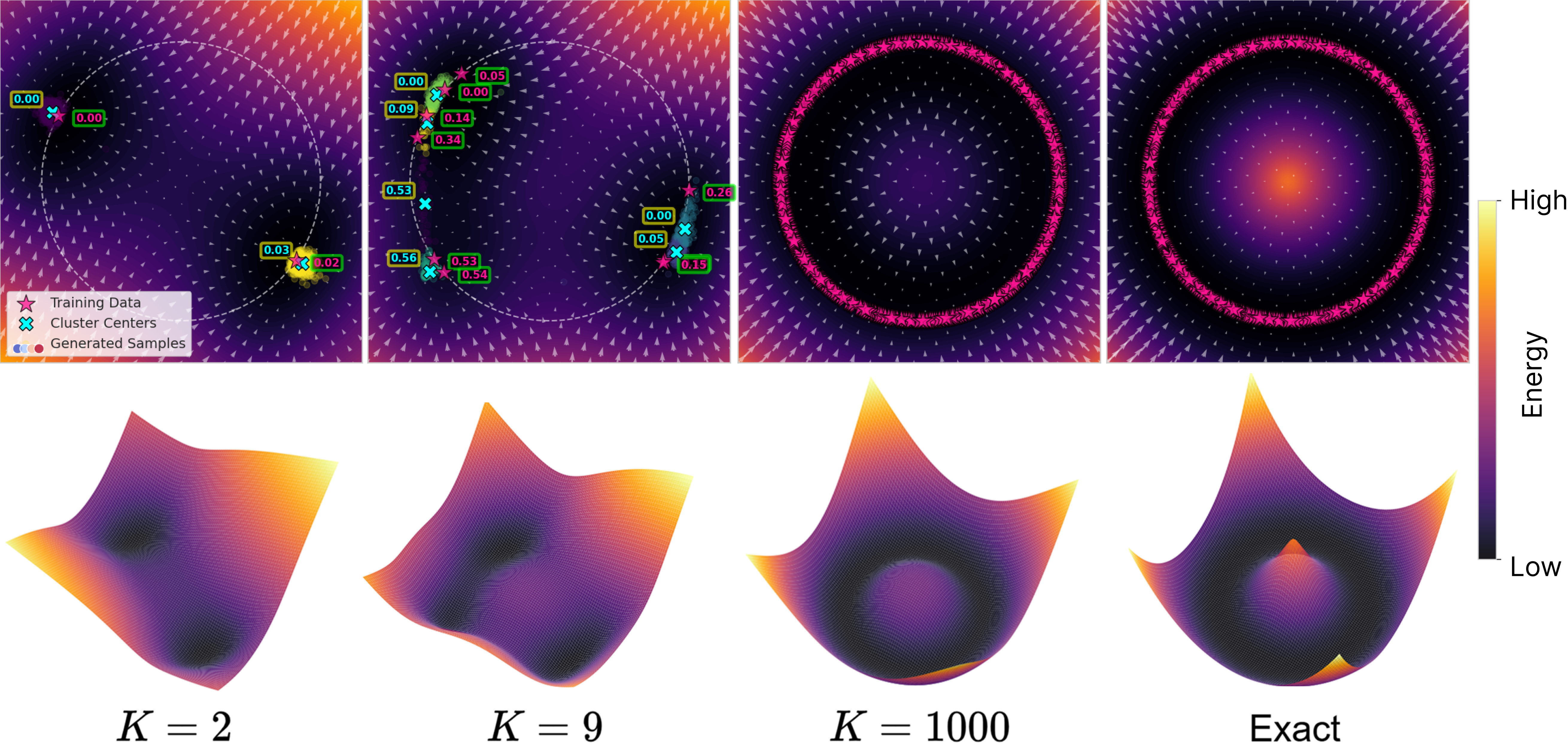} 
    \caption{
    Energy landscape evolution for the 2D toy model as training data size $K$ increases. Models trained at $K \in \{2, 9, 1000\}$ use standard VE-SDE based diffusion pipeline with training data sampled from the unit circle, shown in white for $K \in \{2, 9\}$. Generated samples are shown alongside the learned score field $s_{\theta}(\rvx_t, t)$ done via a neural network, aligned with the negative gradient of the energy \Cref{eq:energy-diffusion}. Hierarchical clustering identifies structure within the generations, with cluster centroid energies visualized by \textcolor{cyan}{$\times$} and numerical value. The rightmost panel shows the exact solution as $K \rightarrow \infty$ derived in \Cref{eqn:toy-model-energy}. As $K$ grows, the model initially memorizes individual data points, forming isolated basins. Around $K = 9$, 
    \textit{spurious patterns} emerge --- distinct low-energy attractors not present in the data --- which mark the onset of generalization. At large $K$, the model enters a fully generalized regime, where low-energy states lie on a flat, continuous manifold shown in \Cref{fig:general-energy-transition}. Top-row figure is replicated from \cite{pham2025memorization}. 
    }
    \label{fig:toy-example}
\end{figure}

To better illustrate the connection between diffusion models and DenseAMs, we can investigate a simple 2-dimensional toy model, see \Cref{fig:toy-example}, that exhibits many aspects of the memorization-generalization transition in these two types of models. Specifically, imagine that the training data lies on a unit circle. We are interested in exploring how the shape of the energy function (\ref{eq:DenseAM}) changes as the number of training data points, used in training a diffusion model, increases. 

Specifically, in the trivial case of a single training point $(K = 1)$, there exists only a single memory $\boldsymbol{\xi}^1$ on the energy landscape of \Cref{eq:DenseAM}, making it independent from the inverse temperature or sharpness value $\beta$. In contrast, when there exists two training points $(K = 2)$, there exists two corresponding minima (or memories) $\boldsymbol{\xi}^1$ and $\boldsymbol{\xi}^2$:
\begin{equation}
\begin{split}
    E^\text{AM}(\rvx) = -\beta^{-1} & \log \Big[\exp \Big(- \beta \lVert \rvx - \boldsymbol{\xi}^1 \rVert^2_2\Big) + \exp\Big(- \beta \lVert \rvx - \boldsymbol{\xi}^2 \rVert^2_2\Big) \Big] ,
\end{split}
\end{equation}
when $\beta \rightarrow \infty$. However, for finite values of $\beta$, there exists a configuration which yields a minimum:
\begin{equation}
    \boldsymbol{\eta} = \argmin_{\rvx} E^\text{AM}(\rvx),
\end{equation}
that does not correspond with any of the two training data points or stored patterns, i.e., $\boldsymbol{\eta}\neq\boldsymbol{\xi}^1$ and $\boldsymbol{\eta}\neq\boldsymbol{\xi}^2$. This ``novel'' local minimum of the energy is the {\em spurious state} illustrated in the cartoonish \Cref{fig:general-energy-transition}.

When the training data size $K \rightarrow \infty$, the empirical data distribution of the toy model can be described as a continuous density of states: 
\begin{equation}
    p(\rvy) = \frac{1}{\pi} \delta\big(y_1^2+y_2^2-1\big) .
    \label{2D density}
\end{equation}
The probability of the generated data is proportional (up to terms independent of the state $\rvx$) to 
\begin{equation}
    p(\rvx) \sim \int\limits_{-\infty}^{+\infty} dy_1 dy_2\ p(\rvy)\ e^{- \beta \lVert \rvx - \rvy \rVert^2_2} = e^{-\beta (R^2+1)} I_0(2\beta R) ,
    \label{eq: continous circle}
\end{equation}
where $I_0(\cdot)$ is a modified Bessel function of the first kind\footnote{In order to obtain \Cref{eq: continous circle}, it is easiest to introduce polar coordinates for both the state vector $\rvx$ and the training data $\rvy$: $$\begin{cases}
    x_1 = R \cos(\phi)\\
    x_2 = R \sin(\phi)
\end{cases} \ \ \ \ \ \ \ \ \ \ \begin{cases}
    y_1 = r \cos(\varphi)\\
    y_2 = r \sin(\varphi)
\end{cases}$$ 
The integral (\ref{eq: continous circle}) can then be written as $$p(\rvx) \sim \int\limits_{0}^{2\pi} d\varphi\int\limits_0^\infty r dr \frac{1}{\pi} \delta(r^2-1) e^{-\beta[R^2+r^2-2Rr\cos(\varphi-\phi)]}
= e^{-\beta (R^2+1)} I_0(2\beta R)
$$ and explicitly computed using the definition of the modified Bessel functions \cite{gradshteyn2014table}.}. Thus, the energy of the 2D circle model is given by 
\begin{equation}
\begin{split}
     E^\text{AM}(R, \phi) & =  R^2+1 - \frac{1}{\beta} \log\big[I_0(2\beta R)\big] \underset{\beta\rightarrow\infty}{\approx} (R - 1)^2 ,
\end{split}
    \label{eqn:toy-model-energy}
\end{equation}
where $R$ is the radius of the unit circle and $\phi$ is the polar angle. Keep in mind, the dependence on $\phi$ in \Cref{eqn:toy-model-energy} disappears for the final result. 

For our diffusion model, we consider the following forward process which describes the VE-setting: 
\begin{equation}
    \mathrm{d} \rvx_t = \sigma \mathrm{d} \rvw_t,
\end{equation}
where the drift term $\rvf(\rvx_t, t) = 0$ and $\rvw_t$ is Brownian motion. The corresponding reverse process is described as 
\begin{equation}
    \mathrm{d} \rvx_t = \big [ -\sigma^2 \nabla_{\rvx_t} \log p_t (\rvx_t) \big ] \mathrm{d}t + \sigma^2 \mathrm{d} \mathbf{w}_t, 
\end{equation}
where the diffusion coefficient $g(t) = \sigma$ is fixed as $1$, matching the radius of the unit circle. Using these SDEs, we trained a set of SDE-based diffusion models, for $K \in \{ 2, 9, 1000\}$, over the time domain of $t \in [\epsilon, 1]$ where $\epsilon = 10^{-5}$, using the objective (\ref{eq: score function definition}). Then, we visualize the energy landscape of each model and record our results in \Cref{fig:toy-example}.

As expected, the local minima of the resulting $E^\text{AM}$ in \Cref{eqn:toy-model-energy} form a {\em continuous manifold}, corresponding to $R=1$. The data samples from the model in \Cref{fig:toy-example} occupy the vicinity of that manifold. This behavior describes the fully generalized phase. In the limit of large $\beta$, the energy landscape is described by a parabola centered around $R=1$. For our trained diffusion model at $K = 1000$, we see that the exact energy and approximated energy, obtained from the diffusion model, are very much aligned to one another in \Cref{fig:toy-example}.

Meanwhile, for small number of data points $(K = 2)$, the diffusion model exhibits memorization. The local minima of the energy correspond to the training data points. Importantly, at $K=9$, we are able to observe the first signs of \textit{spurious states}. At this stage, the model begins to learn emergent (different from the training data) local minima of the energy. Subsequent increase of the size of the training set leads to fully generalized behavior, which is illustrated for $K=1000$. At that stage, all of the samples from the model live in close proximity of the exact data manifold. The right panel shows the analytical expression for the energy landscape, defined by \Cref{eqn:toy-model-energy}. Thus, the conventional diffusion modeling pipeline, following \Cref{eq: score function definition}, agrees very well with the theoretical prediction of the empirical energy (\ref{eqn:toy-model-energy}) and the cartoonish illustration in \Cref{fig:general-energy-transition}.  

From the perspective of DenseAMs, we can see a novel phase also exists in diffusion models --- {\em spurious states} --- previously overlooked in the memorization-generalization literature of these models \cite{somepalli2023diffusion, somepalli2023understanding, meehan2020non, burg2021on, yoon2023diffusion, gu2023memorization, Carlini2023, achilli2024losing, biroli2024dynamical}. As demonstrated in \cite{pham2025memorization}, diffusion models trained on real and high dimensional datasets also follow the same trend illustrated in \Cref{fig:toy-example}: transitioning from memorization to spurious phase to generalization as the training data size $K$ increases. Hence, by viewing the problem of generalization as a failure of storing all of the data points as memories, we can provide a novel understanding of the memorization and generalization in generative diffusion models and interestingly, demonstrate the existence of spurious states in such models. This illustrates that diffusion models behave as AM systems in the small data regime, and as generative models in the large data regime. 

\begin{calloutNotebook}[Comparison of diffusion energy and DenseAM energy] 
In this notebook, we offer the reader the possibility to train a simple diffusion model using data from the 2-D circle as an example. The reader can reconstruct the energy profile of the diffusion model by integrating the score function and compare this energy profile with the energy of DenseAM model. 

\nblinks{https://tutorial.amemory.net/tutorial/diffusion_as_memory.html}{https://github.com/bhoov/amtutorial/blob/main/tutorial_ipynbs/02_diffusion_as_memory.ipynb}{02_diffusion_as_memory}

\begin{center}
    \includegraphics[width=0.65\textwidth]{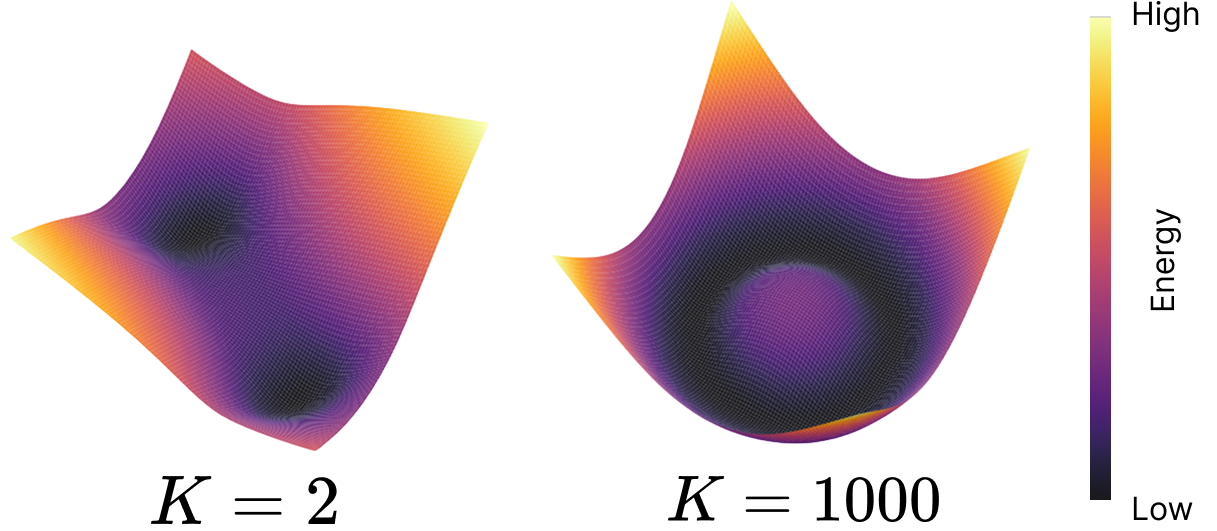}\\
\end{center}

\end{calloutNotebook}\clearpage{}%
\clearpage{}%
\chapter[Associative Memory: A Machine Learning Model]{Associative Memory:\\ A Machine Learning Model} \label{chap:am-stats}

In this Chapter, we will view Associative Memory networks through the lens of machine learning modeling. After presenting a brief discussion on machine learning modeling, and the sources of error in (machine) learning \cref{sec:mlmodel}, we present Associative Memory network as a machine learning model that can be used much like other models in learning, highlighting its inference process, expressivity, application to supervised learning, and its parametric and nonparametric forms, see \cref{sec:am-model}. Then we discuss how this model can be used for the unsupervised learning task of clustering in \cref{sec:cluster}. Finally, we elaborate on the connection between Associative Memory and Kernel Machines, and discuss novel Associative Memory models that emerge from this connection in \cref{sec:kernels}.

\section{Machine Learning Modeling} \label{sec:mlmodel}

The purpose of {\em learning} is to obtain a version of the ground-truth distribution (or equivalently the data-generating function) given (potentially noisy) samples from the ground-truth distribution. The first step is data acquisition. Given data, we choose a model or function class $\gF$ which corresponds to not just a {\em method} (such as Support Vector Machines \cite{VapnikSVM}, Generalized Linear Models \cite{GenLinModel}, Decision trees \cite{quinlan:induction}, etc.), but their specific configuration governed by their respective {\em hyperparameters} such as regularization forms and penalties, trees depth, network architecture and activations, optimization configurations. 
Given our choice of the function class $\gF$, the learning process searches (optimizes) for the function $\hat{f} \in \gF$ that (approximately) minimizes the {\em empirical risk} --- the sample loss computed over the training data --- or some surrogate of it which better represents the {\em true risk} --- the population loss --- or is easier to optimize, such as some continuous version of a discrete loss and/or some form of regularization that mitigates overfitting such as weight penalty or decay, dropout and such.

We currently have an understanding of the factors~\citep{vapnik2006estimation, devroye2013probabilistic, bottou2008tradeoffs, ram2023toward} affecting the {\em excess risk} of this chosen/learned function --- the difference between the true risk of this learned function $\hat{f}$ and the best possible function $f^\star$. At a high level, these factor depend on (i)~the choice of the function class and its capacity to model the data generating process, (ii)~the use of an empirical risk {\em estimate} instead of the true risk to learn this function, and (iii)~the approximation in the empirical risk minimization (ERM) over the class of functions $g \in \gF$.

For a particular method (decision trees, linear models, neural networks), let $\gF$ denote the function class for some {\em fixed} hyperparameter $\lambda \in \Lambda$ (tree depth, number of trees for tree ensembles; regularization parameter for linear and nonlinear models, activation functions, batch size in stochastic gradient descent or SGD, etc.) in the space of valid hyperparameters $\Lambda$. 

Focusing on supervised learning, for any model or function $f : \gX \to \gY$ with $(\rvx, \rvy), \rvx \in \mathcal X, \rvy \in \mathcal Y$ generated from a distribution $\pdata$ over $\gX \times \gY$, and a loss function $\Ls: \gY \times \gY$, the true risk $R(f)$ and the empirical risk $R_m(f)$ of $f$ with $m$ samples $\{(\rvx^i, \rvy^i)\}_{i=1}^m \sim \pdata$ is given by
\begin{equation}\label{eq:risk}
\begin{split}
  & R(f)
  =  \mathbb{E}_{(\rvx, \rvy) \sim \pdata} \left[ \Ls(\rvy, f(\rvx)) \right]
  = \int \Ls(\rvy, f(\rvx)) d \pdata,
\\
  & R_m(f)
  =  \mathbb{E}_m \left[ \Ls(\rvy, f(\rvx) ) \right]
  = \frac{1}{m} \sum_{i=1}^m \Ls(\rvy^i, f(\rvx^i)).
\end{split}
\end{equation}
We denote the {\em Bayes optimal model} as $f^\star$ where, for any $(\rvx, \rvy) \sim \pdata$,
\begin{equation}\label{eq:fstar}
f^\star(\rvx) = \argmin_{\hat{\rvy} \in \gY} \mathbb{E} \left [ \Ls(\rvy,\hat{\rvy}) | \rvx \right].
\end{equation}
We denote with the following:
\begin{equation}\label{eq:fbar-fhat}
  \bar{f} = \argmin_{f \in \gF} R(f),
  \qquad \qquad
  \hat{f}_m = \argmin_{f \in \gF} R_m(f),
\end{equation}
as the {\em true risk minimizer} $\bar{f} \in \gF$ and the {\em empirical risk minimizer} $\hat{f}_m \in \gF$ (with $m$ samples) in model class $\gF$ respectively. 

When performing empirical risk minimization or ERM over $\gF$, the excess risk is given by
\begin{align}
\gE & = R(\hat{f}_m) - R(f^\star) 
      = \underbrace{R(\hat{f}_m) - R(\bar{f})}_{\gE_{\texttt{est}}} 
      + \underbrace{R(\bar{f}) - R(f^\star)}_{\gE_{\texttt{app}}},
\end{align}
which decomposes into two terms: (i)~the {\em approximation risk} $\gE_{\texttt{app}} = R(\bar{f}) - R(f^\star)$, and (ii)~the {\em estimation risk} $\gE_{\texttt{est}}(m) = R(\hat{f}_m) - R(\bar{f})$. For limited number of samples $m$, there is a tradeoff between $\gE_{\texttt{app}}$ and $\gE_{\texttt{est}}$, where a larger function class $\gF$ usually reduces $\gE_{\texttt{app}}$ but increases $\gE_{\texttt{est}}(m)$~\citep{vapnik2006estimation, devroye2013probabilistic}. Roughly speaking, methods are termed {\em universal approximators} if there is some hyperparameter which ensures that the approximation error $\gE_{\texttt{app}}$ can be made arbitrarily small. Of course, the flip side is that this can make the corresponding class of functions $\gF$ very large, often increasing the estimation error $\gE_{\texttt{est}}(m)$ for a fixed $m$.

\begin{figure}[htb]
\centering
\begin{tikzpicture}[scale=0.9]
\draw[line width=1pt] (2, 0) .. controls (0, 1) and (0, 3) .. (3, 4);
\draw[line width=1pt] (3, 4) .. controls (5, 3) and (4, 1) .. (2, 0) node[pos=0.95, right] {\ $\gF$};
\fill[black] (0, 4) circle (4pt) node[right] {\ \  $f^\star$};
  \fill[blue] (1.3, 2.0) circle (4pt) node[left] {$\bar{f}$ \ };
  \fill[brown] (2.7, 2.6) circle (4pt) node[above] {$\hat{f}_m$};
  \fill[magenta] (3.0, 1.2) circle (4pt) node[left] {$\tilde{f}_m$\ };
  \draw[<->, gray, thick] (0.1, 3.9) -- (2.9, 1.3) node[pos=0.3, above] {$\gE$};
  \draw[<->, orange, dashed, thick] (0.1, 3.9) -- (1.2, 2.1) node[pos=0.5, left] {$\gE_{\texttt{app}}$};
  \draw[<->, red!50, dashed, thick] (2.7, 2.45) -- (3, 1.3) node[pos=0.3, right] {$\gE_{\texttt{opt}}$};
  \draw[<->, violet, dashed, thick] (1.4,2.1) -- (2.6, 2.5) node[pos=0.6, above] {$\gE_{\texttt{est}}$};
\end{tikzpicture}
\quad
\begin{tikzpicture}[scale=0.9]
\draw[line width=1pt] (2, 0) .. controls (-3, 2) and (0, 4) .. (3, 4);
\draw[line width=1pt] (3, 4) .. controls (6, 3) and (4, 1) .. (2, 0) node[pos=0.95, right] {\ $\gF$};
\fill[black] (1.2, 2.0) circle (4pt) node[left] {$f^\star \approx \textcolor{blue}{\bar{f}}$\ \ };
\fill[orange] (1.2, 2.55) circle (0pt) node[left] {$\gE_{\texttt{app}} \approx 0$};
  \fill[blue] (1.3, 2.0) circle (4pt);
  \fill[brown] (2.7, 2.6) circle (4pt) node[above] {$\hat{f}_m$};
  \fill[magenta] (3.0, 1.2) circle (4pt) node[below] {$\tilde{f}_m$\ \ \ \ \ };
  \draw[<->, gray, thick] (1.3, 1.9) -- (2.9, 1.3) node[pos=0.4, below] {$\gE$};
  \draw[<->, red!50, dashed, thick] (2.7, 2.45) -- (3, 1.3) node[pos=0.3, right] {$\gE_{\texttt{opt}}$};
  \draw[<->, violet, dashed, thick] (1.4,2.1) -- (2.6, 2.5) node[pos=0.6, above] {$\gE_{\texttt{est}}$};
\end{tikzpicture}
\caption{{\bf Decompositions of excess risk $\gE$.} We depict the decomposition of $\gE$ incurred by the approximate empirical risk minimizer $\tilde{f}_m \in \gF$ found (usually) with a scalable optimization algorithm in the model class $\gF$ with respect to the Bayes optimal model $f^\star$. The true risk minimizer $\bar{f} \in \gF$ is the best approximator of the optimal $f^\star$, while the exact empirical risk minimizer $\hat{f}_m \in \gF$ can be distinct from the true risk minimizer $\bar{f}$ since we are using the empirical risk (obtained with a finite training set) for our learning instead of the true risk. This figure is partially replicated from Ram et al., 2023~\citep{ram2023toward}.
{\bf Left}:~The model class $\gF$ can be such that the Bayes optimal $f^\star \not\in \gF$, hence we would have a nonzero approximation risk $\gE_{\texttt{app}}$ --- the difference in the true risk (defined in \Cref{eq:risk}) between the Bayes optimal $f^\star$ and the true risk minimizer $\bar{f}$ in our model class $\gF$ --- with $\gE_{\texttt{app}} > 0$.
{\bf Right}:~We can also select a large model class $\gF$ such that the Bayes optimal $f^\star$ is in the model class $\gF$ or can be approximated to arbitrary precision with a model in the function class. In this case, the true risk minimizer $\bar{f} \approx f^\star$ will have (almost) zero approximation risk with $\gE_{\texttt{app}} \approx 0$.
However, it is important to note that the estimation risk $\gE_{\texttt{est}}$ --- the difference in the risk between the true risk minimizer and the empirical risk minimizer --- is often related to the size of the model class $\gF$ (for some notion of size), with larger model classes incurring larger estimation risk. The optimization risk $\gE_{\texttt{opt}}$ --- the difference between the risk of the exact and approximate empirical risk minimizers --- can also be affected implicitly by the size of the model class $\gF$, where larger classes require more computational resources for the learning optimization to achieve any specific level of empirical risk approximation; conversely, for fixed computational resources, larger function classes can incur larger optimization risk.}
\label{fig:ed-orig}
\end{figure}
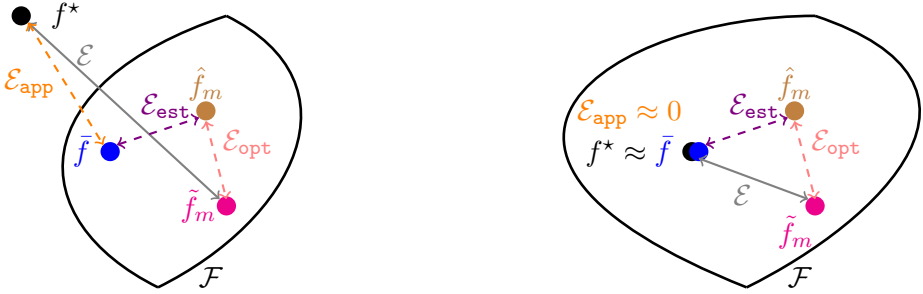

Bottou and Bosquet~\citep{bottou2008tradeoffs} study the tradeoffs in a ``large-scale'' setting where the learning is compute bound (in addition to the limited number of samples $m$). Given any computational budget $T$, they consider the learning setting ``small-scale'' when the number of samples $m$ is small enough to allow for the ERM to be performed to arbitrary precision. In this case, the tradeoff is between the $\gE_{\texttt{app}}$ and $\gE_{\texttt{est}}$ terms (as above). They consider the large scale setting where the ERM needs to be approximated given the computational budget and discuss the tradeoffs in the excess risk of an approximate empirical risk minimizer $\tilde{f}_m \in \gF$. In addition to $\gE_{\texttt{app}}$ and $\gE_{\texttt{est}}$, they introduce the {\em optimization risk} term $\gE_{\texttt{opt}} = R(\tilde{f}_m) - R(\hat{f}_m)$ --- the excess risk incurred due to approximate ERM --- and argue that, in compute-bound large-scale learning, approximate ERM on all the samples $m$ can achieve lower excess risk than high precision ERM on a subsample of size $m' \leq m$. \Cref{fig:ed-orig} provides a visual representation of this excess risk decomposition.

For parametric models, the functions $f \in \gF$ are parameterized with $\theta \in \Theta$ where $\gF \triangleq \{ f_\theta, \theta \in \Theta \}$, where we explicitly denote the dependence of $\theta$ in $f_\theta$. Thus, the true risk minimizer, and the empirical risk minimizer can be respectively written as:
\begin{align}
\bar{f} & \triangleq f_{\bar{\theta}}, \quad \bar{\theta} = \argmin_{\theta \in \Theta} R(f_\theta) =  \argmin_{\theta \in \Theta} \mathbb{E}_{ (\rvx, \rvy) \sim \pdata } \left[ \Ls( \rvy, f_\theta(\rvx) )  \right] \\
\hat{f}_m & \triangleq f_{\hat{\theta}_m}, \quad \hat{\theta}_m = \argmin_{\theta \in \Theta} R_m(f_\theta) = \argmin_{\theta \in \Theta} \frac{1}{m} \sum_{ i = 1 }^m \Ls( \rvy^i, f_\theta(\rvx^i) ).
\end{align}
The approximate empirical risk minimizer would be denoted with $f_{\tilde{\theta}_m}$ with corresponding model parameters $\tilde{\theta}_m$.
The parameters $\theta \in \Theta$ corresponds to model/function specific parameters --- weights and biases for linear models and neural networks; split dimensions and thresholds, and leaf node values for univariate decision trees. Once these parameters $\hat{\theta}_m$ are learned from the data, one can make predictions $f_{\hat{\theta}_m}(\rvx)$ on new inputs $\rvx \in \gX$ without having to keep the training data $\{ (\rvx^i, \rvy^i) \}_{ i = 1 }^m$ around anymore.
For nonparametric models such as nearest neighbor models and (kernel) density estimation based models (such as the Nadaraya-Watson estimator), the training data is also required for making predictions, and thus are considered as part of the ``model parameters''.

\section{Associative Memory Network as a Model} \label{sec:am-model}

The previously discussed Associative Memory networks can be viewed as a parameterized model $f_\bXi: \gX \to \gX$ with parameters $\bXi$. For the sake of simplicity, let us for now assume that $\gX \subseteq \R^D$, the $D$-dimensional Euclidean space. The interpretation is detailed in the following:

\begin{calloutInfo}[\bf Model parameters are stored patterns]
The model parameters $\bXi$ can be reshaped as a $(D \times K)$ matrix, and is usually termed as the $K$ {\em stored patterns} $\{\bxi^\mu \in \R^D, \mu \in \iset{ K } \}$ --- each stored pattern $\bxi^\mu$ is a $D$ dimensional vector. Note that these model parameters can be learned.
\end{calloutInfo}
\begin{calloutInfo}[\bf Energy function]
Given these model parameters $\bXi$, we have an energy function of a state $\rvv \in \gX \subseteq \R^D$, usually of the following general form:
\begin{equation} \label{eq:en-gen}
E_\beta(\rvv; \bXi) = -Q \left( \sum_{\mu = 1}^K F \left( \beta \, S \left[  \sigma(\rvv), \bxi^\mu \right]  \right) \right),
\end{equation}
\end{calloutInfo}
\begin{itemize}
\item One can view the $\rvv$ as the internal state and $\sigma(\rvv)$ as its activation. See \cref{sec:hamux-neurons} of \cref{chap:am-blocks} for a discussion on states and activations.
\item The $S: \gX \times \gX \to \R$ denotes a notion of similarity such as a dot-product or negative squared Euclidean distance.
\item $\beta > 0$ is the inverse temperature and controls how much high similarities are magnified and low similarities are diminished. This inverse temperature $\beta$ controls the sharpness of the energy around the memories, with larger values of $\beta$ inducing sharper energy landscapes while smaller values generating smoother ones. See Notebook 2.1 in \cref{nb:storage-recovery}, \cref{chap:DenseAM} for a demonstration.
\item The separation function $F: \R \to \R$ is a fast-growing function such as $F(z) = z^p$ or $F(z) = \exp(z)$ for some $z \in \R$.
\item The scaling function $Q: \R \to \R$ is a monotonic non-decreasing function such as identity $Q(z) = z$ or logarithm $Q(z) = \log z$ for some $z \in \R$.
\end{itemize}
\begin{calloutInfo}[\bf Inference via energy descent]
Given a learning rate $\eta > 0$, number of steps $T$ and a clamping mask $\vm \in \{0, 1\}^D$, $f_\bXi(\rvx)$ for some $\rvx \in \gX$ is computed as follows via (clamped) coordinate gradient descent over the energy function, with the descent initialized as the input $\rvx$:
\begin{align}
\label{eq:ed-init} \rvv^{(0)} & \gets \rvx, \\
\label{eq:ed-step} \rvv^{(t)} 
  & \gets \rvv^{(t-1)} - \eta \, \vm \odot \left. \nabla_\rvv E_\beta(\rvv; \bXi) \right|_{\rvv = \rvv^{(t-1)}}, \quad t \in \iset{ T }, \\
\label{eq:ed-out} f_\bXi^\vm(\rvx) & \triangleq \rvv^{(T)},
\end{align}
where $\odot$ denotes the element-wise multiplication of vectors. In the absence of the clamping mask, we drop the $\vm$ subscript and use $f_\bXi$.
\end{calloutInfo}
\begin{itemize}
\item As the learning rate $\eta \to 0$, this energy gradient descent is defined by the following dynamics:
  \begin{equation}
  \frac{d \rvv}{dt} = - \vm \odot \nabla_\rvv E_\beta(\rvv; \bXi).
  \end{equation}
\item The learning rate $\eta$ does not need to be fixed, and can also vary with time. For example, the learning rate can decay with time.
\item When the number of steps $T$ (also referred to as the number of layers in the Associative Memory network) goes to infinity (that is, $T \to \infty$), for an appropriately set learning rate $\eta$ (sufficiently small or appropriately scheduled),  $f_\bXi(\rvx)$ will be one of the local minima of the energy function --- that is, $\nabla_\rvv E_\beta(\rvv; \bXi) = \bm{0}, \nabla_\rvv^2 E_\beta(\rvv; \bXi) \succ 0$ at $\rvv = f_\bXi(\rvx)$. These fixed points (local minima) are often termed as the {\em retrieved memories}. Note that we are seeking local minima and not saddle points which correspond to {\em meta-stable states} where it is hard to decrease the energy but it is not a local minima. See Demircigil et al., 2017~\citep{demircigil2017model} for a discussion of the energy landscape and the {\em basins of attraction} for the different local minima.
\item The gradient clamping mask $\vm$ enables {\em clamping} of a subset of the state variables. When $\vm$ is the $D$-dimensional all-one vector $\bm{1}_N$, the complete state vector $\rvv$ is modified in \Cref{eq:ed-step}. If $\vm = \left[ \bm{1}_{N'}^\top, \bm{0}_{(N-N')}^\top \right]^\top$, then only the first $N' < N$ entries of the state vector $\rvv$ are allowed to be modified, while the remaining $(N - N')$ entries of $\rvv$ are clamped to their initial values obtained from the input $\rvx$. This clamping and coordinate-wise gradient descent is discussed in \Cref{sec:enmin-ff} of \Cref{chap:am-blocks}, see also \Cref{fig:ebm-landscape-prediction} for a visualization of the clamped energy-descent.
\item If the learning rate $\eta$ is small, and $T$ is not too large, the input would not be modified significantly, and $f_\bXi(\rvx) \approx \rvx$. If $\eta$ is large (and not decayed appropriately), we might never arrive at a local minima of the energy.
\end{itemize}

Given the model parameters $\bXi$ (that can be interpreted as $K$ stored patterns), the various hyperparameters --- the functions $Q, F, S$ in the energy function, the inverse temperature $\beta$, the learning rate $\eta > 0$, the number of layers/steps $T$ --- define the inference process with this model $f_\bXi$. There is a tight relationship between the energy function and probability density through the \href{https://en.wikipedia.org/wiki/Boltzmann_distribution}{Boltzmann distribution} --- that is, the density $p(\rvv)$ of a state $\rvv$ is tied its energy $E(\rvv)$ as $p(\rvv) \propto \exp(-E(\rvv))$. Given this interpretation, the inference with the described Associative Memory network amounts to a form of likelihood maximization via gradient descent.

\begin{calloutInfo}[Classic energy for binary patterns]
As an example, if $\bxi^\mu \in \{-1, 1\}^D \, \forall \mu \in \iset{ K }$, $\sigma: \R \to \{-1, +1\}$, $S[\rvv, \rvv'] = \ip{v}{v'}$, $F(z) = z^2$, $\beta = 1$ and $Q$ is the identity function, then the corresponding energy function is that of the Classic Hopfield Network (CHN):
\begin{equation}\label{eq:en-chn}
E(\rvv; \bXi) = - \sum_{\mu = 1 }^K \left( \ip{ \sigma(\rvv) }{ \bxi^\mu } \right)^2.
\end{equation}
\end{calloutInfo}

\begin{calloutInfo}[Log-sum-exp energy with real valued patterns]
With $\bxi^\mu  \in \R^D \, \forall \mu \in \iset{ K }$, $\sigma$ as identity, $S[\rvv, \rvv'] = -\nicefrac{1}{2} \norm{\rvv - \rvv'}^2$, $F(z) = \exp(z)$ and $Q(z) = \log z$, we obtain the widely used log-sum-exp or LSE energy:
\begin{equation} \label{eq:en-lse}
E_\beta(\rvv; \bXi) = - \log \sum_{\mu = 1}^K \exp\left(- \nicefrac{\beta}{2} \norm{\rvv - \bxi^\mu}^2 \right).
\end{equation}
Note that common representations of the LSE energy contains a preceding $(\nicefrac{1}{\beta})$ term on the right-hand-side to cancel out the $\beta$-scaling in the gradient. However, we are removing this here since we only care about the direction of the gradient, and not the magnitude.
\end{calloutInfo}

\subsection{Memory Capacity and Expressivity} \label{sec:am-model:exp}

Given the above energy function, we can now consider the set $\gM \subset \gX$ of local minima of the energy function in \Cref{eq:en-gen} defined as:
\begin{equation} 
\label{eq:en-lmins}
\gM = \left \{ \rvv \in \gX: \nabla_\rvv E_\beta(\rvv; \bXi) = 0, \nabla_\rvv^2 E_\beta(\rvv; \bXi) \succ 0 \right\}.
\end{equation}

Note that this set of local minima will depend on all but two of the various hyperparameters previously discussed --- this set does not depend on the learning rate $\eta$ and the number of steps $T$. In the scenario where the learning rate $\eta \to 0$ and the number of layers/steps $T \to \infty$, for any input $\rvx \in \gX$, the output $f_\bXi(\rvx) \in \gM$ as the output is one of the local minima, thereby essentially making $f_\bXi: \gX \to \gM$ a surjective function (many-to-one mapping); note that $f_\bXi$ for the same $\bXi$ may not be a surjection with a finite $T$ especially when the value of $T$ is small.

For a stored pattern $\bxi^\mu$, if there exists a local minima $\rvv \in \gM$ such that $\rvv \approx \bxi^\mu$, then the model has {\em memorized} a stored pattern and is able to approximately retrieve it (given an appropriate input initializing the energy descent). The {\em memory capacity} of an Associative Memory network is informally defined as the largest number $K^{\text{max}}$ of randomly generated stored patterns $\bXi$ such that each stored pattern can be (approximately) retrieved. For example, with the classic energy in \Cref{eq:en-chn}, $K^{\text{max}} \sim O(D)$; with the log-sum-exp energy in \Cref{eq:en-lse}, there exists hyperparameters (specifically, values of $\beta$) such that $K^{\text{max}} \sim O(\exp(D))$. See \Cref{chap:DenseAM} for further discussion of memory capacity.

Given that, under appropriate configurations, the Associative Memory network operates as a surjection $f_\bXi: \gX \to \gM$, the expressivity or the approximation ability of the model $f_\bXi$ is related to the size (cardinality) of $\gM$. We can increase the cardinality of $\gM$ to up to $K^{\text{max}}$ by increasing the model size (in terms of the number of model parameters) to up to $D K^{\text{max}}$ corresponding to $\bXi$ containing $K^{\text{max}}$ stored patterns in $D$-dimensions. Beyond that point, increasing the number of (randomly generated) stored patterns in $K^{\text{max}}$ would not increase the cardinality of $\gM$.
There are ways to carefully design the stored patterns such that the cardinality of $\gM$ can go beyond $K^{\text{max}}$.

\subsection{Supervised Learning} 
\label{sec:am-model:supervised}

While we have discussed the Associative Memory network as a model $f_\bXi: \gX \to \gX$ that maps from $\gX$ to $\gX$, in the previously discussed supervised learning setup of \cref{sec:mlmodel}, we usually consider models of the form $f_\theta: \gX \to \gY$  mapping from a feature space $\gX$ to an output space $\gY$. One way to handle that with an Associative Memory network is to consider a model $f_\bXi: \gZ \to \gZ$, where $\gZ \triangleq \gX \times \gY$. If $\gX$ is a $d$-dimensional feature space, and $\gY$ is a $k$-dimensional output space, $\gZ$ would be a $D=(d+k)$-dimensional space with the features and output concatenated into the state vector. This concept is also described in \Cref{sec:enmin-ff} of \Cref{chap:am-blocks}. 

Consider the clamping vector $\vm = [\bm{0}_d^\top \bm{1}_k^\top ]^\top \in \{0,1\}^D$, and the matrix $\mM = \left[ \mathbf{0}_{k\times d} \  \mI_k \right] \in \{0, 1\}^{k \times D}$. Also consider an uninformative default (potentially learnable) prediction $\rvy_0 \in \gY$ --- as an example, $\rvy_0 = \bm{0}_k$ for regression or $\rvy_0 = (\nicefrac{1}{k}) \bm{1}_k$ for $k$-class classification. Then we can define a function $g_\bXi: \gX \to \gY$ mapping features $\rvx$ to predictions using parameters $\bXi$ as follows:
\begin{align}
\rvv^{(0)} & \gets [\rvx^\top, \rvy_0^\top]^\top, \\
\rvv^{(t)} 
  & \gets \rvv^{(t-1)} - \eta \, \vm \odot \left. \nabla_\rvv E_\beta(\rvv; \bXi) \right|_{\rvv = \rvv^{(t-1)}}, \quad t \in \iset{ T }, \\
g_\bXi(\rvx) & \triangleq \mM \rvv^{(T)}.
\end{align}
See \Cref{fig:ebm-landscape-prediction} for a visualization this energy minimization based inference process for a supervised learning problem. There are a few important things to note here:
\begin{itemize}
\item The energy $E_\beta(\rvv, \bXi)$ now depends on a similarity function $S: \gZ \times \gZ$, which can incorporate similarity between the features in $\gX$ and the outputs in $\gY$. For example, the similarity function $S$ can be defined as $S[\rvz, \rvz'] = \lambda S_\gX [\rvx, \rvx'] + (1 - \lambda) S_\gY[\rvy, \rvy']$ where $\rvz = [\rvx^\top \rvy^\top]^\top$ ($\rvz'$ defined accordingly with $\rvx',\rvy'$), and $S_\gX: \gX \times \gX \to \R$, $S_\gY: \gY \times \gY \to \R$ are feature and output specific similarity functions.
\item During the energy descent, the features values $\rvx \in \gX$ provide the initialization $\rvv^{(0)} = [\rvx^\top \rvy_0^\top]^\top$, and then are not modified at all because of the clamping mask $\vm$.
\item The final output of $g_\bXi$ is obtained by extracting the last $k$-dimensions of the final state $\rvv^{(T)}$ for the energy descent with the element selecting matrix $\mM$.
\item We are considering a gradient descent over the energy $E_\beta(\rvv; \bXi)$, which would define a density over $(\gX \times \gY)$. However, we are clamping the state variables corresponding $\gX$ to the input $\rvx$. This roughly corresponds to a conditional density over $\gY$, and the clamped energy descent corresponds to a conditional likelihood maximization.
\end{itemize}

\subsection{Nonparametric vs Parametric Models} \label{sec:am-model:params}

\subsubsection{Nonparametric Models}

An important question with an Associative Memory model $f_\bXi$ is the process of obtaining the model parameters $\bXi$ (also known as the ``stored patterns''). Given a set of patterns $\{ \bxi^\mu, \mu \in \iset{ K } \}$ (possibly from an unknown distribution $\pdata$, that is, $\bxi^\mu \sim \pdata$), we can consider a nonparametric form of the model where $\bXi$ is just all of the data $\{\bxi^\mu, \mu \in \iset{ K } \}$, with the size of the model (which would be $O( KD )$ when each stored pattern $\bxi^\mu$ is of size $D$) growing with the number of stored patterns (which is $K$). As discussed previously in \cref{sec:am-model:exp}, the corresponding energy function $E_\beta(\cdot; \bXi)$ can have up to $O( \min\{ K, K^{\text{max}} \})$ local minima, where $K^{\text{max}}$ is the capacity of the model. Note that if the stored patterns have a specific structure, then the number of local minima can be higher than $K^{\text{max}}$. For the supervised learning setup discussed in \cref{sec:am-model:supervised}, the stored patterns $\bxi^\mu$ would be feature-output pairs $(\rvx, \rvy), \rvx \in \gX, \rvy \in \gY$ in the training set.

The single-step retrieval dynamics (that is, number of layers $T = 1$) of the nonparametric Associative Memory networks has recently also been recently interpreted as the solution of a specific nonparametric support vector regression problem~\citep{hu2024nonparametric}. Different choices of kernel functions and training data preprocessing result in different energy functions $E_\beta(\cdot; \bXi)$.

\subsubsection{Parametric Models}

One can also consider a parametric form of the model $f_\bXi$, where the size of $\bXi$ is pre-specified, and the parameters are learned using the data. As an example, we can say that the size of $\bXi$ is such that it can store only $K$ patterns $\bxi^\mu, \mu \in \iset{K}$ of size $D$. However, here we are allowed to learn these patterns $\bxi^\mu$. Given a set of training examples $S = \{ \rvz^i \}_{i = 1}^m$, a loss function $\Ls$ and a regularizer $R$, at a high level, we can learn $\bXi$ by solving the following problem:
\begin{equation}
\min_{\bXi}  R(\bXi) + \frac{1}{m} \sum_{i = 1}^m \Ls(\rvz^i, f_\bXi(\rvz^i)).
\end{equation}
If the training example $\rvz_i$ is a feature-label pair $(\rvx_i,\rvy_i)$, and the loss is the negative cross-entropy loss between the labels, and the Associative Memory model is as defined in \cref{sec:am-model:supervised}, then $\Ls(\rvz^i, f_\bXi(\rvz^i))$ would simplify to $\texttt{NegativeCrossEntropy}(\rvy^i, f_\bXi(\rvx^i))$ as in a standard classification problem. The regularization $R(\bXi)$ can be utilized to avoid overfitting. For example, $R(\bXi) = \lambda \norm{\bXi}_{2,1} = \lambda \sum_{\mu \in \iset{ K }} \norm{\bxi^\mu}$ penalizes the norm of the learnable stored patterns, scaled with a $\lambda \in \R_+$. One can also consider a regularization that enforces the learnable stored patterns to be well separated so as to make the memory retrieval process more efficient and robust~\citep{wu2024uniform, hu2024provably}. The loss $\Ls$ and the regularization $R$ can be specified in a problem dependent manner. As the model $f_\bXi$ corresponds to a $T$-layer recursive network, we can learn $\bXi$ with (stochastic) gradient descent given that the loss $\Ls$ and the regularization $R$ are differentiable.

\section{Clustering} \label{sec:cluster}

\begin{figure}[htb]
\centering
\includegraphics[width=0.1265\textwidth]{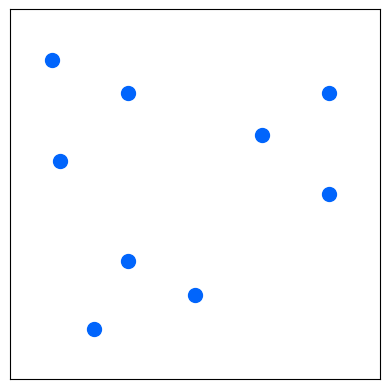}
~
\includegraphics[width=0.15\textwidth]{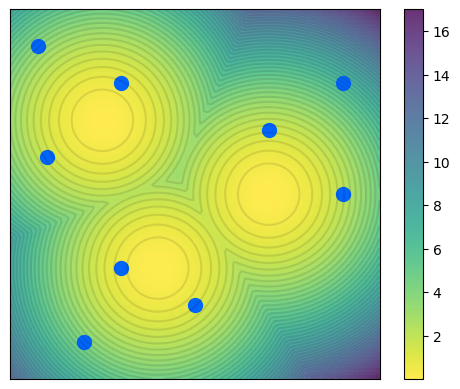}
~
\includegraphics[width=0.15\textwidth]{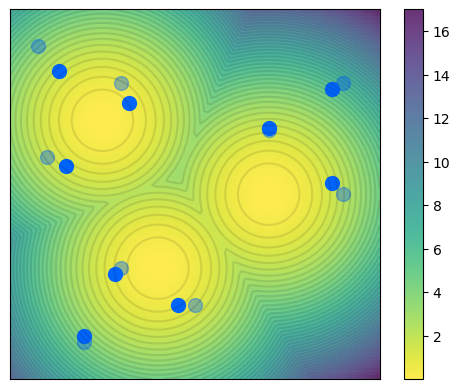}
~
\includegraphics[width=0.15\textwidth]{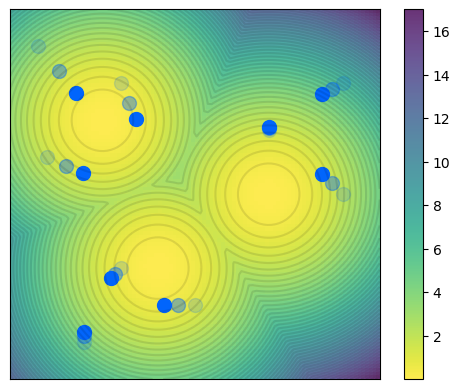}
~
\includegraphics[width=0.15\textwidth]{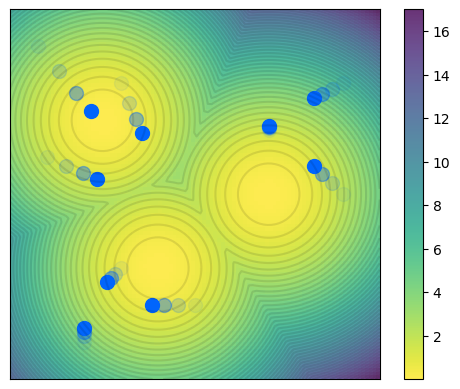}
~
\includegraphics[width=0.1265\textwidth]{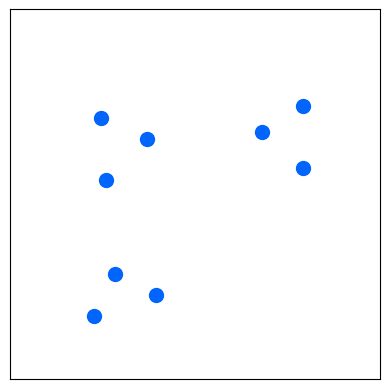}
\caption{{\bf Contraction of points over the energy landscape.} Here we demonstrate how the Associative Memory model can contract a set of points via gradient descent over the energy landscape.
{\bf Column~1}:~The initial set of 9 points.
{\bf Column~2}:~The initial set of points are overlaid on an energy landscape with 3 local minima  --- lighter colour denotes lower energy.
{\bf Columns~3-5}:~Each point separately undergoes the energy descent for 3 steps corresponding to a 3-layer model.
{\bf Column~6}:~The outputs of the model applied separately to each point in the set are now a contracted version of the initial set (column~1).}
\label{fig:point-contraction}
\end{figure}

Consider an Associative Memory network based model with parameters $\bXi = \{ \bxi^\mu \in \R^D, \mu \in \iset{ K } \}$ and an energy function $E_\beta(\cdot; \bXi): \R^D \to \R$ as defined in \Cref{sec:am-model}. With any input $\rvx \in \R^D$, the energy descent involved in the model output $f_\bXi(\rvx)$ intuitively would move $\rvx$ towards the local energy minimum closest to it. When the number of local minima is relatively small, for all input, the outputs will contract towards this small set of local minima. An example of this behaviour is shown in \Cref{fig:point-contraction}. One way of thinking about this contraction effect is the following --- the Associative Memory model moves relatively close-by points closer together, potentially moving far-away points even farther. What is interesting is that this is a collective contraction effect on the whole set of inputs even though the model operates on all the input independently.

If the set of input points are all close-by to begin with, and the energy local minima are relatively more spread out, then all points could potentially contract towards the same local minima. However, if the input points are as spread out as the energy local minima, then the contraction effect would lead to input points getting more {\em clustered} --- a subset of the points getting closer to each other while each subset getting farther away from each other. This capability of the Associative Memory network makes it quite useful for the classical problem of clustering.

\subsection{Euclidean Clustering} \label{sec:cluster:clam}

Given a set of points $S = \{\rvx^i \in \R^D, i \in \iset{m} \}$ in a $D$-dimensional Euclidean space, a commonly studied clustering problem is the $k$-means clustering problem, which seeks to solve the following discrete optimization problem:
\begin{equation} \label{eq:kmeans}
\min_{\rvc^1, \ldots, \rvc^k \in \R^D}  \sum_{i = 1}^m  \min_{j \in \iset{k} } \norm{ \rvx^i - \rvc^j }^2.
\end{equation}
This problem seeks to partition the set of points $S$ into $k$ disjoint subsets $C^j, j \in \iset{ k }$, with a prototype or center $\rvc^j \in \R^d$ for each subset $C^j$, ensuring that squared Euclidean distance between a point in the subset and the corresponding center is small. This is a NP-hard problem even for $k=2$~\citep{dasgupta2008hardness}, and Lloyd's algorithm~\citep{lloyd1982least} is the most commonly used approximate algorithm though many more efficient algorithms with improved approximation guarantees have been developed. The hardness of this problem is partially due to the discrete nature of the objective in \Cref{eq:kmeans}, and thus usually requires discrete algorithms. This objective in its current form is not conducive to gradient descent based solutions prevalent in modern machine learning. One can modify this discrete objective into a continuous one by replacing the $\min_{j \in \iset{k} }$ in \Cref{eq:kmeans} with a soft-min function, leading to soft or fuzzy $k$-means clustering~\citep{dunn1974fuzzy, bezdek2013pattern}:
\begin{equation} \label{eq:soft-kmeans}
\min_{\rvc^1, \ldots, \rvc^k \in \R^D}  \sum_{i = 1}^m  \sum_{j \in \iset{k}} \frac{
\exp(- \gamma \norm{  \rvx^i - \rvc^j }^2 ) \norm{ \rvx^i - \rvc^j }^2 
}{
\sum_{j' \in \iset{k}} \exp(- \gamma \norm{ \rvx^i - \rvc^{j'} }^2 )
},
\end{equation}
where $\gamma > 0$ is a hyperparameter. The above objective is an upper bound of the $k$-means objective in \Cref{eq:kmeans} and we would be minimizing the upper bound. Larger values of $\gamma$ make the upper bound tighter.

\begin{figure}[htb]
\centering
\includegraphics[width=0.2\textwidth]{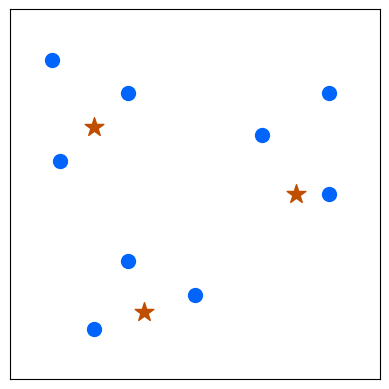}
~
\includegraphics[width=0.2\textwidth]{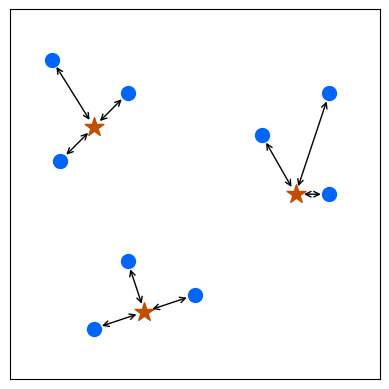}
~
\includegraphics[width=0.2\textwidth]{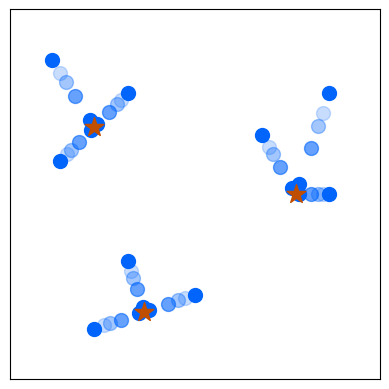}
~
\includegraphics[width=0.2\textwidth]{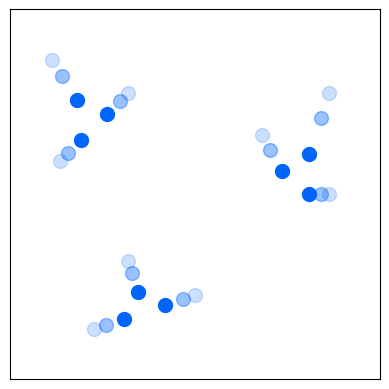}
\caption{{\bf Computing the $k$-means objective with points and cluster centers.} The computation of the $k$-means objective requires us to implicitly or explicitly assign points to clusters. 
{\bf Column~1}:~We are given a set of points~\textcolor{NavyBlue}{$\bullet$} and centers~\textcolor{BrickRed}{$\boldsymbol{\star}$}.
{\bf Column~2}:~The $k$-means objective in \Cref{eq:kmeans} assigns each point to its closest center ($\textcolor{NavyBlue}{\bullet}{\leftrightarrow}\textcolor{BrickRed}{\boldsymbol{\star}}$), and then sums this distance-to-closest-center over all points. 
{\bf Column~3}:~Instead of relaxing the discreteness in the $k$-means objective as in soft $k$-means in \Cref{eq:soft-kmeans}, ClAM~\citep{saha23end} uses an Associative Memory network with (learnable) parameters $\bXi$ to effectively relocate each point to its closest center, and then considers the sum of these per-point-relocation in \Cref{eq:kmeans-clam} as a surrogate for the $k$-means objective.
{\bf Column~4}:~Instead of complete contraction to the cluster centers, sometimes it might be beneficial to partially contract to the cluster centers~\citep{saha2024deep}.}
\label{fig:cluster-point-assign}
\end{figure}

Instead of relaxing the discrete assignments of points to clusters $\min_{j \in \iset{k}} \norm{  \rvx^i - \rvc^j}^2$, one can emulate the discrete assignments by ``moving'' a point $\rvx^i$ to its closest cluster $\rvc^{j^\star(\rvx^i)}$ with $j^\star(\rvx^i) \triangleq \argmin_{j \in \iset{k}} \norm{ \rvx^i - \rvc^j}^2$ using the contraction capability of Associative Memory networks and using the term $\norm{ \rvx^i - \rvc^{j^\star(\rvx_i)}}^2$, the amount by which the point was ``moved''~\citep{saha23end}. Thus, the $k$-means objective in \Cref{eq:kmeans} can be re-written as:
\begin{equation}\label{eq:kmeans-clam}
\begin{split}
\min_{\rvc^1, \ldots, \rvc^k \in \R^D} \sum_{i = 1}^m \min_{j \in \iset{ k } } \norm{ \rvx^i - \rvc^j }^2
& \equiv
\min_{\rvc^1, \ldots, \rvc^k \in \R^D} \sum_{i = 1}^m  \norm{\rvx^i - \rvc^{j^\star(\rvx^i)} }^2
\\
& \equiv
\min_\bXi \sum_{i = 1}^m \norm{ \rvx^i - f_\bXi(\rvx^i) }^2
\text{ if } f_\bXi(\rvx^i) \approx \rvc^{j^\star(\rvx^i)}.
\end{split}
\end{equation}
This distinction and equivalence is visualized in \cref{fig:cluster-point-assign}.

\begin{figure}[htb]
\centering
\includegraphics[width=0.8\textwidth]{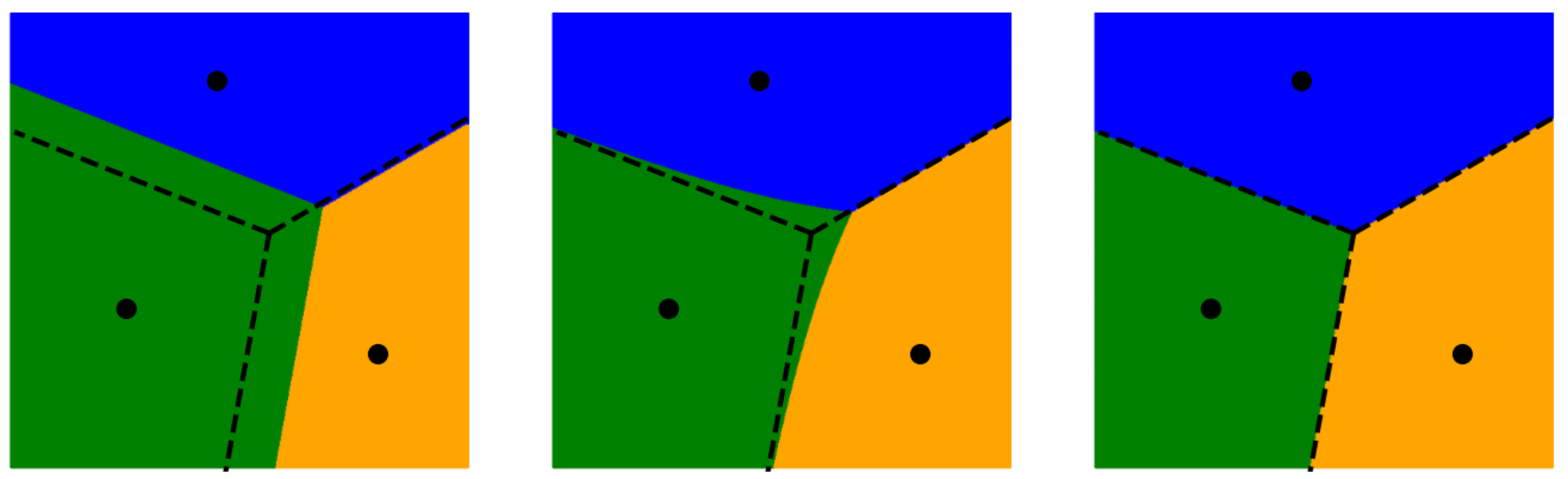}
\caption{{\bf Basins of attraction vs Voronoi partition.} The basins of attraction of the given   memories/centers (black dots $\bullet$) for different $\beta$ values with a 10-layer Associative Memory network $f_\bXi$ ($T=10$) are shown by the colored regions. Dashed lines show the desired Voronoi partition. As the value of $\beta$ increases, the basins of attraction start aligning with the desired Voronoi partitions. {\bf Column 1}: For a small inverse-temperature $\beta = 0.001$, the Voronoi partition does not align with the basins of attraction of the Associative Memory network. {\bf Column 2}: With a higher inverse-temperature $\beta = 10$, the basins of attraction partially align with the Voronoi partition. {\bf Column 3}: With a high inverse-temperature $\beta = 100$, the basins of attraction and the Voronoi partition are practically indistinguishable. This figure is replicated from Saha et al., 2023~\citep{saha23end}.}
\label{fig:basin-attract-voronoi}
\end{figure}

What we need then is an Associative Memory model $f_\bXi: \R^D \to \R^D$ such that $f_\bXi(\rvx^i) \approx \rvc^{j^\star(\rvx^i)}$. If use the $k$ cluster centers as the stored patterns, that is $\bXi \triangleq \{ \rvc^1, \ldots, \rvc^k \}$, then the condition $f_\bXi(\rvx^i) \approx \rvc^{j^\star(\rvx^i)}$ would be satisfied if the basins of attraction around each cluster center (which is also the stored pattern) matches the {\em Voronoi partition} of the input space given the cluster centers. Given $k$ centers, the Voronoi partition of the space is (i)~a $k$-partition of space with each partition corresponding to a specific center, and (ii)~any point in a specific partition has its corresponding center as its closest center. Given 3 centers, \Cref{fig:basin-attract-voronoi} shows the Voronoi partition of the input space.

\begin{figure}[htb]
\centering
\includegraphics[width=0.8\textwidth]{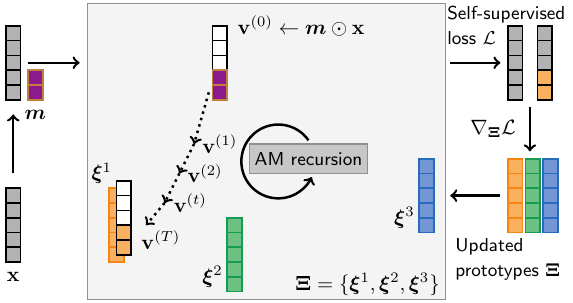}
\caption{{\bf Euclidean clustering with DenseAM.} For $\rvx \in S$, we first apply a mask (in \textcolor{Plum}{purple}) $\vm \in \{0, 1\}^D$ to $\rvx$ to get the initial iterate $\rvv^{(0)}$ for the AM recursion. With $T$ recursions, we have a completed version $\rvv^{(T)}$. The use of the mask $\vm$ is optional, and allows for a semi-supervised clustering loss by leveraging the clamped inference $f_\bXi^{\bar{\vm}}(\rvx)$ in Associative Memory networks; see \cref{eq:clam-ssl} and Saha et al., 2023~\citep[Section 3.4]{saha23end} for details. In the limiting case, there is no mask (that is, $\vm = \mathbf{1}_D$) and $\rvv^{(0)} \gets \rvx$ and we do the unclamped inference $f_\bXi(\rvx)$. The stored patterns $\bXi$ are updated with the gradient $\nabla_\bXi \Ls$ on the loss in \cref{eq:kmeans-clam}. This figure is replicated from Saha et al., 2023~\citep{saha23end}.}
\label{fig:clam-algorithm}
\end{figure}

With $\bXi \triangleq \{\rvc^1, \ldots, \rvc^k \}$, and the following energy function $E_\beta(\cdot; \bXi)$ for an appropriately large inverse temperature $\beta > 0$ and number of layers $T$, we can get the desired behaviour:
\begin{equation} \label{eq:clam-energy}
E_\beta(\rvv; \bXi) = - \frac{1}{\beta} \log \sum_{j = 1}^k \exp(-\beta \norm{ \rvv - \rvc^j}^2 ).
\end{equation}
The desired behavior of having $f_\bXi(\rvx^i) \approx \rvc^{j^\star(\rvx^i)}$ corresponds to the basins of attraction of each stored pattern $\bxi^\mu, \mu \in \iset{ k }$ matching Voronoi partition of the space given the memories/centers $\{ \bxi^\mu = \rvc^\mu, \mu \in \iset{ k } \}$. The dependence on $\beta$ for $T=10$ layers of the DenseAM is visualized in \Cref{fig:basin-attract-voronoi}.
This allows us to solve the discrete clustering problem in \Cref{eq:kmeans} with the re-written objective (\ref{eq:kmeans-clam}) completely with gradient descent, since we can differentiate through the $T$ layers of the Associative Memory network. Additionally, we can leverage the clamped inference procedure in Associative Memory networks to extend the standard clustering objective over the training set $S$ to effectively give us a {\em self-supervised clustering loss} by creating multiple versions of each point $\rvx \in S$ --- thereby enabling self-supervision in clustering. Concretely, we can mask each input $\rvx$ with a random $\vm \in \{0, 1\}^d$ to form the input $\vm \odot \rvx$ to the Associative Memory network. Then, we can perform clamped inference $f_\bXi^{\bar{\vm}}(\vm \odot \rvx)$  with the clamping mask $\bar{\vm}$ which is the complement of the mask $\vm$. Thus, our standard clustering loss in \Cref{eq:kmeans-clam} would be extended as following from the standard clustering loss on the left to the self-supervised clustering loss on the right:
\begin{equation} \label{eq:clam-ssl}
\min_\bXi \sum_{\rvx \in S} \| \rvx - f_\bXi(\rvx) \|^2
\quad \xrightarrow{\text{self-supervision}} \quad
\min_\bXi \sum_{\rvx \in S} \mathbb E_\vm \| \vm \odot ( \rvx - f_\bXi^{\bar{\vm}}( \vm \odot \rvx ) \|^2
\end{equation}
The overall clustering process, which involves the learning of the stored patterns $\bXi$, is visualized in \Cref{fig:clam-algorithm}.

\subsection{Deep Clustering} \label{sec:cluster:dcam}

For data modalities, such as image or text, it is often necessary to first learn information-preserving Euclidean representations before clustering these learned representations. This problem of jointly learning representations and clustering is often referred to as \textit{deep clustering}~\citep{min2018survey, ren2022deep, zhou2025comprehensive}. One common way to learn information-preserving representations is to use an auto-encoder and minimize the reconstruction error, where $e_\phi:\gX \to \R^D$ is a domain-specific encoder (parameterized with $\phi$) that maps the input (images, text) into a latent Euclidean space, which is then used to reconstruct the original data using a decoder $d_\vartheta: \R^D \to \gX$ (parameterized with $\vartheta$) which often mirrors the domain-specific encoder, and  the reconstruction loss defined as $\Ls_r( \rvx, d_\vartheta ( e_\phi ( \rvx ) ) )$ for a loss $\Ls_r: \gX \times \gX \to \R$, giving us the following learning problem given a dataset $S \subset \gX$:
\begin{equation} \label{eq:ae-recon}
\min_{\phi, \vartheta} \sum_{\rvx \in S}
\Ls_r(\rvx, d_\vartheta ( e_\phi (\rvx)))
\end{equation}
A simple but useful baseline is to just solve the above problem, and then perform $k$-means clustering on latent representations $e_\phi(S) = \{ e_\phi(\rvx), \rvx \in S \}$. However, as we are already learning representations, it is beneficial to steer the learned representations to already have a favourable clustered structure.

This is often obtained by augmenting the reconstruction loss $\Ls_r( \rvx, d_\vartheta (e_\phi ( \rvx )))$ with some form of a clustering loss $\Ls_c(e_\phi( \rvx ), \{ \rvc^1, \ldots, \rvc^k\})$, like a relaxed version of $\min_{j \in \iset{k}} \norm{e_\phi( \rvx) - \rvc^j }^2$, where $\rvc^j \in \R^D, j \in \iset{K}$ are learnable cluster centers in the latent space; see \cref{eq:soft-kmeans} as an example of a continuous clustering loss. Thus, the overall learning problem can be written as following for a regularization hyperparameter $\lambda \in [0, 1]$:
\begin{equation} \label{eq:recon-cluster-balance}
\min_{\phi, \vartheta, \{ \rvc^1, \ldots, \rvc^k \} \subset \R^D } \sum_{ \rvx \in S}
(1 - \lambda)
\Ls_r( \rvx, d_\vartheta ( e_\phi (\rvx)))
+ \lambda
\Ls_c(e_\phi(\rvx), \{\rvc^1, \ldots, \rvc^k\}).
\end{equation}
The hyperparameter $\lambda$ balances the reconstruction and clustering loss as there is an inherent tradeoff between (i)~preserving information in the latent space with $e_\phi(\rvx) \not\approx e_\phi(\rvx')$ for $\rvx \not= \rvx'$ --- thus low reconstruction loss, and (ii)~forming tight clusters in the latent space by mapping all points within a cluster to almost the same representation, that is $e_\phi(\rvx) \approx \rvc^{j^\star(\rvx)}$ where $j^\star(\rvx) \triangleq \argmin_{j \in \iset{k}} \norm{e_\phi(\rvx) - \rvc^j }^2$ --- giving us low clustering loss. The goal is to find the sweet spot, which allows us to have low reconstruction loss (preserving necessary information), while forming a clustered structure in the latent space by pushing both the representations and the cluster centers to have low clustering loss. There is also an implicit objective of forming well-balanced clusters and avoiding {\em representation collapse}, where all points end up in the same cluster in the latent space.

By viewing Associative Memory network as a contractive layer (see \cref{fig:point-contraction}), we can introduce a clustered structure in the learnable latent space in an alternate manner. Instead of maintaining an objective $\Ls_c$ for clustering that pushes the latent representations to be clustered when minimized, we can employ the contractive nature of the Associative Memory networks to directly have a clustered structure in the latent space, and just focus on optimizing the reconstruction loss of this structured latent space by contracting the latent space before reconstructing~\citep{saha2024deep}. 

Given a domain-specific encoder $e_\phi$, a corresponding decoder $d_\vartheta$, and an Associative Memory model $f_\bXi$ serving as a contraction layer, we can solve the following optimization problem:
\begin{equation}\label{eq:dcam-obj}
\min_{\phi, \vartheta, \bXi} \sum_{\rvx \in S}
\Ls_r ( \rvx, d_\vartheta ( f_\bXi ( e_\phi (  \rvx )))).
\end{equation}
Here, the input $\rvx$ is first encoded to the latent space as $e_\phi(\rvx) \in \R^D$, and passed through the contraction layer to get $f_\bXi( e_\phi (\rvx))$. Then the original input is reconstructed with the decoder to get $d_\vartheta ( f_\bXi ( e_\phi( \rvx ) ) )$. In contrast to the use of Associative Memory networks in vanilla Euclidean clustering~\citep{saha23end} where we want complete contraction --- the points are relocated to the closest cluster center --- in this setup, it is beneficial to only consider partial contraction --- the points are modified to have a more clustered structure, but the output of the model $f_\bXi$ are still distinct for distinct models. This is visualized in \cref{fig:cluster-point-assign} with \cref{fig:cluster-point-assign} (Column~3) showing complete contraction, while \cref{fig:cluster-point-assign} (Column~4) visualizing partial contraction.

\begin{figure}[htb]
\centering
\includegraphics[width=0.7\textwidth]{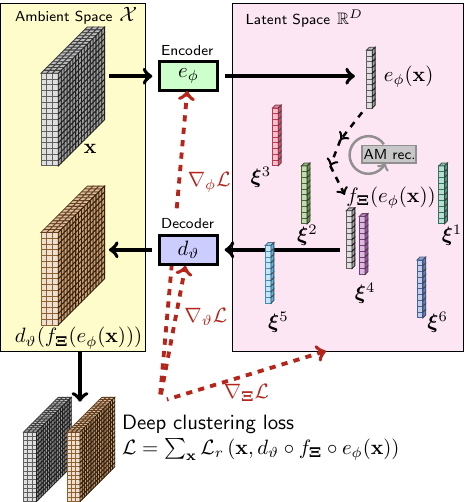}
\caption{{\bf Deep clustering with DenseAM.} Given an input $\rvx \in \gX$ in the ambient space $\gX$, the encoder $e_\phi$ maps $\rvx$ to the latent space to get $e_\phi(\rvx) \in \R^D$. Then we use the (partial) contraction capability of a Associative Memory model $f_\bXi$ to move the latent representation from $e_\phi(\rvx)$ to $f_\bXi \circ e_\phi(\rvx)$ towards one of the memories. This contracted representation is then mapped back to the ambient space $\gX$ with the decoder $d_\vartheta$ to get $d_\vartheta \circ f_\bXi \circ e_\phi( \rvx )$. For the purposes of learning the encoder, decoder and Associative Memory network parameters, we utilize the reconstruction loss $\Ls_r( \rvx, d_\vartheta \circ f_\bXi \circ e_\phi( \rvx ) )$ and backpropagate the gradients with respect to the parameters $\phi, \bXi, \vartheta$. The solid arrows denote the forward-pass $\rvx \to e_\phi(\rvx) \to f_\bXi(\rvx) \to d_\vartheta \circ f_\bXi \circ e_\phi( \rvx )$ to compute the single loss term in \cref{eq:dcam-obj}. The \textcolor{BrickRed}{dashed arrows} denote the backward pass showing the single loss driving all updates. This figure is replicated from Saha et al., 2024~\citep{saha2024deep}.}
\label{fig:dclam}
\end{figure}
The overall learning procedure is shown in \cref{fig:dclam}. 
This method provides a single loss function that simultaneously ensures that the loss of information is minimized, while pushing the latent representations to have a clustered structure through the Associative Memory network; no separate clustering loss is utilized here. This loss function, and thus the resulting deep clustering scheme, is agnostic to the data modality (images or text or something else) and the corresponding encoder and decoder architectures. Given that the Associative Memory network model $f_\bXi$ is differentiable, with respect to its learnable parameters $\bXi$ and its output, we can perform deep clustering by the solving \cref{eq:dcam-obj} with (stochastic) gradient descent provided the encoder $e_\phi$ and decoder $d_\vartheta$ are differentiable with respect to their respective parameters and output.

\section{Kernel Machines} \label{sec:kernels}

Revisiting the energy function of an Associative Memory network $f_\bXi$ with parameters $\bXi$ in \cref{eq:en-gen}, and denoting the $F( \beta S[ \rvv, \bxi^\mu])$ term with $\kappa( \rvv, \bxi^\mu)$, we can write the energy function as:
\begin{equation}\label{eq:en-gen-kernel}
E_\beta(\rvv; \bXi)  = - Q \left( \sum_{\mu \in \iset{ K }} \kappa(\rvv, \bxi^\mu) \right),
\end{equation}
where the $\sum_\mu \kappa(\rvv, \bxi^\mu)$ term can be interpreted as a {\em kernel sum} with the kernel $\kappa: \gX \times \gX \to \R$, the core computation in kernel machines~\citep{bottou2007large}. This simple observation allows us to leverage the rich literature on kernel machine for the development of novel Associative Memory networks with unique capabilities. Two main areas of research in kernel machines focus on the following:
\begin{itemize}
\item A lot of research focused on the development and use of expressive domain-specific kernels $\kappa$, and the understanding of their properties such as expressivity and generalization. In the context of Associative Memory networks, this corresponds to the development of novel domain-specific energy functions, since one can create an energy given a kernel function through \cref{eq:en-gen-kernel}, thereby expanding the applicability of these models to new domains.
\item Every inference in vanilla kernel machines requires the computation of the kernel sum $\sum_\mu \kappa(\rvv, \bxi^\mu)$ which (i)~implies that we need to keep the set $\{\bxi^\mu\}_{\mu \in \iset{ K }}$ even for inference, and (ii)~leads to extremely expensive training and inference as each inference is naively $O(K)$, the number of memories (or terms in the kernel sum). A lot of research focused on improving the computational time and space complexity of these kernel-sum computations. This corresponds to improving the computational time and space complexity of the computation of the energy and thus the energy gradient --- this would speed up each energy descent step and thus the overall inference with a model $f_\bXi$.
\end{itemize}

\subsection{Random Features} \label{sec:kernels:rf}

Roughly speaking, for a \href{https://en.wikipedia.org/wiki/Positive-definite_kernel}{symmetric positive definite kernel} $\kappa: \gX \times \gX \to \R$, there exists an implicit feature map $\phi: \gX \to \gH$, where $\gH$ is a Reproducing Kernel Hilbert space, such that for any $\rvx, \rvx' \in \gX$, $\kappa( \rvx, \rvx' ) = \ip{\phi( \rvx )}{\phi( \rvx' )}$. If an explicit feature map $\phi$ is available, a kernel sum can be simplified as follows:
\begin{equation} \label{eq:ksum-explicit-map}
\sum_{\mu = 1}^K \kappa( \rvv, \bxi^\mu) = \sum_{\mu = 1}^K \ip{\phi( \rvv )}{\phi(\bxi^\mu)}
= \ip{\phi( \rvv )}{\sum_{\mu=1}^K \phi(\bxi^\mu)},
\end{equation}
where we would just need to compute the $\sum_{\mu=1}^K \phi(\bxi^\mu)$ term once, and use it for any subsequent inference, thereby removing the $O(K)$ dependence both from the time and space complexity of the inference. However, note that the computational complexities now depends on the size of the feature map $\phi( \rvv)$ and $\phi(\bxi^\mu)$.

A commonly used and expressive kernel is the \href{https://en.wikipedia.org/wiki/Radial_basis_function_kernel}{RBF (radial basis function) kernel} $\kappa: \R^D \times \R^D \to \R_+$ with $\kappa( \rvx, \rvx' ) = \exp(-\gamma \norm{ \rvx - \rvx' }^2)$ for a scaling parameter $\gamma > 0$. This kernel possesses an infinite dimensional feature map $\phi: \R^D \to \R^\infty$, thus making the explicit feature map practically unusable. Various indexing schemes have been developed and analysed~\citep{ram2009linear, curtin2013tree, curtin2015plug} to speed up the computation of kernel sums..

As an alternate seminal approach~\citep{rahimi2007random}, random Fourier features were used to develop approximate feature maps $\Phi: \R^D \to \R^Y$ for the RBF kernel such that $\kappa( \rvx, \rvx' ) \approx \ip{\Phi(\rvx)}{\Phi(\rvx')}$~\citep{rahimi2007random}. More precisely, with high probability,
\begin{equation}\label{eq:rff-approx}
\forall \rvx, \rvx' \in \R^D, \left| \kappa( \rvx,  \rvx') - \ip{\Phi( \rvx )}{\Phi( \rvx' )} \right| \sim O \left( \sqrt{D / Y} \right),
\end{equation}
implying that an $\epsilon$ approximation guarantee requires $Y \sim O(D / \epsilon^2)$. The first set of random feature maps for the RBF kernel were defined as follows using trigonometric functions:
\begin{equation} \label{eq:rmap}
\Phi( \rvx ) = \frac{1}{\sqrt{Y}} \left[ \begin{array}{c}
     \cos(\ip{\bomega^1}{ \rvx } + b^1)  \\
     \cos(\ip{\bomega^2}{ \rvx } + b^2)  \\
     \cdots \\
     \cos(\ip{\bomega^Y}{ \rvx } + b^Y)  \\
\end{array} \right],
\text{ and }
\Phi( \rvx ) = \frac{1}{\sqrt{Y}} \left[ \begin{array}{c}
     \cos(\ip{\bomega^1}{ \rvx })  \\
     \sin(\ip{\bomega^1}{ \rvx })  \\
     \cos(\ip{\bomega^2}{ \rvx })  \\
     \sin(\ip{\bomega^2}{ \rvx })  \\
     \cdots \\
     \cos(\ip{\bomega^Y}{ \rvx })  \\
     \sin(\ip{\bomega^Y}{ \rvx })  \\
\end{array} \right],
\begin{array}{l}
\bomega^i \sim \gN(0, \mI_D),\\
b^i \sim \gU(0, 1), \\
\forall i \in \iset{Y}.
\end{array}
\end{equation}
Note that the first set of random features produces a $Y$-dimensional feature map with $Y$ random features while the second set produces a $2Y$-dimensional feature map using $Y$ random features, but provides better approximation guarantees. The $\gN(0, \mI_D)$ denotes the $D$-dimensional isotropic standard normal distribution, and the $\gU(0, 1)$ denotes the univariate uniform distribution over the range $[0, 1]$.

While this is not the main focus of these lecture notes, we will show how these random features are obtained because we think they are very neat. The original random features by Rahimi and Recht, 2007~\citep{rahimi2007random} made use of Bochner's theorem~\citep{rudin2017fourier} for shift-invariant kernels.

\begin{calloutDef}[Shift-invariant kernel]
A kernel $\kappa: \R^D \times \R^D \to \R$ is considered shift-invariant if $\kappa(\rvx, \rvx') = \kappa(\rvx - \rvx')$ is a function of $(\rvx - \rvx')$. Thus, for a fixed $\Delta \in \R^D$, $\kappa(\rvx, \rvx + \Delta) = \kappa(\rvz, \rvz+\Delta) = \kappa(\Delta)$ for any $\rvx, \rvz \in \R^D$.
\end{calloutDef}

In the following, we state Bochner's theorem, and show how the random features naturally emerge from them:
\begin{theorem}[Bochner's theorem]
A continuous kernel $\kappa(\rvx, \rvx') = \kappa(\rvx - \rvx')$ on $\R^D$ is positive definite if and only if $\kappa(\Delta)$ is Fourier transform of a non-negative measure $p$.
\end{theorem}
\begin{calloutInfo}
If we have access to this measure $p$, then we can see that
\begin{align}
\kappa(\rvx - \rvx') 
& = \int_{\R^D} p(\bomega) \exp(\mathrm{i} \ip{\bomega}{(\rvx - \rvx')} d\bomega \\
& = \mathbb{E}_{\bomega \sim p} \exp(\mathrm{i} \ip{\bomega}{\rvx}) \exp(\mathrm{i} \ip{\bomega}{-\rvx'})
\\
& =  \mathbb{E}_{\bomega \sim p}
\left( \cos \ip{\bomega}{\rvx} + \mathrm{i} \sin \ip{\bomega}{\rvx} \right)
\left( \cos \ip{\bomega}{\rvx'} - \mathrm{i} \sin \ip{\bomega}{\rvx} \right)
\\
& =  \mathbb{E}_{\bomega \sim p} \left(
\cos \ip{\bomega}{\rvx} \cos \ip{\bomega}{\rvx'} + \sin \ip{\bomega}{\rvx} \sin \ip{\bomega}{\rvx'}
\right)
\\
& =  \mathbb{E}_{\bomega \sim p} \ip{
\left[
\begin{array}{c} \cos \ip{\bomega}{\rvx} \\ \sin \ip{\bomega}{\rvx} \end{array}
\right]
}{
\left[
\begin{array}{c} \cos \ip{\bomega}{\rvx'} \\ \sin \ip{\bomega}{\rvx'} \end{array}
\right]
}
\\
& \approx \frac{1}{Y} \sum_{j = 1}^Y  \ip{
\left[
\begin{array}{c} \cos \ip{\bomega^j}{\rvx} \\ \sin \ip{\bomega^j}{\rvx} \end{array}
\right]
}{
\left[
\begin{array}{c} \cos \ip{\bomega^j}{\rvx'} \\ \sin \ip{\bomega^j}{\rvx'} \end{array}
\right]
}
 =  \ip{\Phi(\rvx)}{\Phi(\rvx')},
\end{align}
with $\Phi(\rvx)$ defined as in the right side of \cref{eq:rmap}, but with a generic $\bomega^j \sim p$.
\end{calloutInfo}
If $\kappa(\Delta)$ is properly scaled, Bochner's theorem also implies that the Fourier transform $p(\bomega)$ of the shift-invariant kernel is a proper probability measure, and can be used to generate the random features. Since the Fourier transform of a Gaussian is a Gaussian, we can use $\bomega \sim \gN(0, \mI_D)$ for the RBF kernel. But the above process also gives us a way to approximate any (properly scaled) shift-invariant kernel function, as long as we have access to the corresponding non-negative measure $p$.

Note that we approximate the expectation over $\bomega \sim p$ with a sample mean. This means that, even though the RBF kernel is always non-negative, the estimated kernel value can be negative as the trigonometric functions can take negative values. In applications where it is undesirable to have negative estimates for the non-negative RBF kernel (for example, with the log-sum-exp energy), these trigonometric random features are not appropriate. For this reason, positive random features were developed~\citep{choromanski2020rethinking}. Even their derivation is quite simple, and takes a different route than that of the original random features~\citep{rahimi2007random} but is only applicable to the RBF kernel; the original trigonometric random features presented above are applicable to any shift-invariant kernel as long as we have access to the corresponding non-negative measure. 

In the following we will develop the positive random features:
\begin{calloutInfo}
We begin with the following two equalities:
\begin{equation}\label{eq:rbf-to-sum}
\exp(- \nicefrac{1}{2} \norm{ \rvx - \rvx' }^2 ) = 
\exp(- \norm{\rvx}^2) \exp(\nicefrac{1}{2} \norm{ \rvx + \rvx' }^2 ) \exp(- \norm{\rvx'}^2),
\end{equation}
\begin{equation}\label{eq:gaussian-dist}
c_0 \int_{\R^D} \exp(- \nicefrac{1}{2} \norm{ \bomega - \rvc}^2 ) d \bomega = 1 \quad \forall \rvc \in \R^D,
\end{equation}
for an appropriate normalization $c_0 > 0$. The second equality is just from the fact that the normal density, centered around any point $\rvc \in \R^D$, integrates to 1.

Multiplying $\exp(\nicefrac{1}{2} \norm{\rvx + \rvx'}^2)$ to both sides of \cref{eq:gaussian-dist}, and  setting $\rvc =  (\rvx + \rvx')$, we have
\begin{align}
\exp(\nicefrac{1}{2} \norm{\rvx + \rvx'}^2) 
& = 
c_0 \int_{\R^D} \exp(\nicefrac{1}{2} \norm{\rvx + \rvx'}^2) \exp(- \nicefrac{1}{2} \norm{ \bomega - (\rvx + \rvx')}^2 ) d \bomega 
\\ &
= c_0 \int_{\R^D} \exp(- \nicefrac{1}{2} \norm{ \bomega }^2 ) \exp( \ip{\bomega}{(\rvx + \rvx')} ) d \bomega 
\\ &
= \mathbb{E}_{\bomega \sim \gN(0, \mI_D)} \exp( \ip{\bomega}{(\rvx + \rvx')} )
\\ &
= \mathbb{E}_{\bomega \sim \gN(0, \mI_D)} \exp( \ip{\bomega}{\rvx} ) \exp(\ip{\bomega}{\rvx'} )
\\ &
\approx \frac{1}{Y}\sum_{j = 1}^Y \exp( \ip{\bomega^j}{\rvx} ) \exp(\ip{\bomega^j}{\rvx'} ), \quad \bomega^j \sim \gN(0, \mI_D).
\end{align}
Thus, using above with \cref{eq:rbf-to-sum}, we can define the positive random features for the RBF kernel as follows (note that the following random features only make use of the exponential function, and thus ensure that all random features are positive, implying that the estimate of the RBF kernel value $\kappa(\rvx, \rvx')$ obtained via the dot-product of the random features $\ip{\Phi(\rvx)}{\Phi(\rvx')}$ is guaranteed to be positive):
\begin{equation}
\Phi(\rvx) = \frac{\exp(- \norm{\rvx}^2)}{\sqrt{Y}} \left[
\begin{array}{c}
\exp(\ip{\bomega^1}{\rvx}\\
\exp(\ip{\bomega^2}{\rvx}\\
\cdots \\
\exp(\ip{\bomega^Y}{\rvx}
\end{array}
\right].
\end{equation}
\end{calloutInfo}

Since then, various other random features have been developed for the RBF kernel~\citep{yu2016orthogonal, choromanski2017unreasonable, likhosherstov2023denseexponential}, and for various other kernels~\citep{kar2012random, hamid2014compact}. Liu et al., 2021~\citep{liu2021random} provide a comprehensive survey of random feature for kernel approximation.

In the context of Associative Memory, this allows us to approximate the energy function of the form in \cref{eq:en-gen-kernel} as follows:
\begin{equation}
E_\beta( \rvv; \bXi)  = - Q \left( \sum_{\mu \in \iset{ K } } \kappa( \rvv, \bxi^\mu) \right)
\approx
-Q \left( \ip{\Phi( \rvv )}{ \underbrace{\sum_{\mu=1}^K \Phi(\bxi^\mu)}_{\triangleq \mT} } \right) 
= \tilde{E}_\beta( \rvv; \mT),
\end{equation}
where the computation of the energy $E_\beta(\cdot; \bXi)$ (and its gradient) requires us to have access to all the stored patterns $\bXi = \{ \bxi^\mu, \mu \in \iset{ K } \}$ of size $KD$, while the computation of the approximate energy $\tilde{E}_\beta(\cdot; \mT)$ (and its gradient) only requires us to have access to the $\mT \triangleq \sum_{\mu=1}^K \Phi(\bxi^\mu)$ of size $Y$ and the random features $\{(\bomega^i, b^i), i \in \iset{Y} \}$ (which can be generated on the fly given the $Y$ random seeds and thus do not need to be stored explicitly)~\citep{hoover2024dense}. We can now perform inference via gradient descent on this approximate energy, providing an unique (Dense) Associative Memory model that does not require the stored patterns $\bXi$ for inference. It has been shown that the approximation in the energy translates to approximation in the inference --- the inference $f_\bXi(\rvx)$ by minimizing the exact energy $E_\beta(\cdot; \bXi)$ is approximated with the inference $f_\mT(\rvx)$ by minimizing the approximate energy $\tilde{E}_\beta(\cdot; \mT)$. This appproximation is affected by the following factors~\citep{hoover2024dense}:
\begin{itemize}
\item The approximation depends on the kernel approximation introduced by the random features with a factor of $O(\sqrt{N/Y})$, with larger number of random features $Y$ improving the approximation.
\item The approximation also depends on the initial energy $E_\beta(\rvx; \bXi)$ of the input $\rvx$ --- the initial state for the energy descent --- with larger initial energy leading to higher levels of approximation.
\item The hyperparameter $\eta$ which corresponds to the step-size (or learning rate) of the energy gradient descent, with smaller $\eta$ implying lower levels of approximation.
\end{itemize}

Of course, if we are already considering a kernel function $\kappa$ in \cref{eq:en-gen-kernel}, which has an explicit feature map $\phi$ (for example if $\kappa( \rvx, \rvx') = \ip{ \rvx }{\rvx'}$ or $\kappa(\rvx, \rvx') = \left(\ip{ \rvx }{ \rvx' }\right)^2$), then we can directly use the explicit exact feature map to simplify the exact energy, and incur no approximation in the kernel function evaluation, and thus in the Associative Memory model inference.

\begin{calloutNotebook}[Distributed Representation for Dense Associative Memory]
In this notebook, we demonstrate how we utilize random features to disentangle the size of the Dense Associative Memory network from the number of memories to be stored. Given the standard log-sum-exp energy $E_\beta(\cdot; \bXi)$, corresponding to a model $f_\bXi$ of size $O(DK)$, we demonstrate how we can use the trigonometric random features to develop an approximate energy $\tilde{E}_\beta(\cdot; \mT)$ using a distributed representation $\mT$ of the memories $\bXi = \{ \bxi^\mu, \mu \in \iset{ k } \}$, thus giving us a model $f_\mT$ of size $O(Y)$.

{\nblinks{https://tutorial.amemory.net/tutorial/distributed_memory.html}{https://github.com/bhoov/amtutorial/blob/main/tutorial_ipynbs/03_distributed_memory.ipynb}{03_distributed_memory}}

\begin{minipage}{0.55\textwidth}
{\small
\begin{equation*}
\begin{split}
\underbrace{E_\beta(\rvv; \bXi) }_{f_\bXi \sim O(DK) \text{ size}}
& = - \log \sum_{\mu = 1}^K \exp( - \nicefrac{\beta}{2} \norm{\rvv - \bxi^\mu}^2 ) 
\\
& \approx
- \log \sum_{\mu = 1}^K \ip{\Phi(\sqrt{\beta}\rvv)}{ \Phi( \sqrt{\beta} \bxi^\mu) } 
\\
& = - \log \ip{\Phi(\sqrt{\beta}\rvv)}{ \sum_{\mu=1}^K \Phi( \sqrt{\beta} \bxi^\mu) } 
\\
& = - \log \ip{\Phi(\sqrt{\beta} \rvv}{\mT} 
= \underbrace{\tilde{E}_\beta(\rvv; \mT)}_{f_\mT \sim O(Y) \text{ size}}
\end{split}
\end{equation*}
}
\end{minipage}
\hfill
\begin{minipage}{0.4\textwidth}
{\small
\begin{equation*}
\begin{split} 
\Phi( \rvx ) = \frac{1}{\sqrt{Y}} \left[ \begin{array}{c}
     \cos(\ip{\bomega^1}{ \rvx })  \\
     \sin(\ip{\bomega^1}{ \rvx })  \\
     \cos(\ip{\bomega^2}{ \rvx })  \\
     \sin(\ip{\bomega^2}{ \rvx })  \\
     \cdots \\
     \cos(\ip{\bomega^Y}{ \rvx })  \\
     \sin(\ip{\bomega^Y}{ \rvx })  \\
\end{array} \right]
&
\\
\bomega^i \sim \gN(0, \mI_D),
\forall i \in \iset{Y}
&
\end{split}
\end{equation*}
}
\end{minipage}

\end{calloutNotebook}

\subsection{Novel Energy Functions} \label{sec:kernels:new-en}

As discussed earlier in \cref{sec:am-model}, given an energy function, we can define a probability density through the Boltzmann distribution. Alternately, given a density function $p: \gX \to \R_{\geq 0}$, we can define a energy function $E(\rvv) \propto -\log p(\rvv)$ through the same relationship. 

Given a set of samples $\bXi = \{ \bxi^\mu \sim \pdata, \mu \in \iset{ K } \}$ from an unknown distribution $\pdata$ over $\gX \subset \R^D$, one way to define a density function $\hat{p}$ is through kernel density estimation, where the goal is to devise a $\hat{p}$ that closely approximates the unknown $\pdata$. A kernel density estimate or KDE at any point $ \rvv \in \gX$ is defined as:
\begin{equation} \label{eq:kde}
\hat{p}_h(\rvv; \bXi)  = \frac{1}{K h} \sum_{\mu = 1}^K \kappa\left( \frac{ \rvv - \bxi^\mu}{h} \right),
\end{equation}
where $\kappa: \R^d \to \R_{\geq 0}$ is the kernel function, and $h > 0$ is the kernel bandwidth. For this to be a valid density, the kernel function needs to satisfy the following conditions:
\begin{itemize}
\item Symmetry: $\kappa( \rvx ) = \kappa(- \rvx) \quad \forall \rvx \in \gX$
\item Nonnegativity: $\kappa( \rvx ) \geq 0 \,\,\, \quad \forall \rvx \in \gX$
\item Normalization: $\int_\rvx \kappa(\rvx) \, \mathrm{d} \rvx = 1$.
\end{itemize}

For multivariate data (that is $D > 1$), the kernel $\kappa$ has been defined both as $\kappa(\rvx) = c \kappa_1(\norm{\rvx})$ or $\kappa( \rvx ) = c' \prod_{i = 1}^D \kappa_1( | \ervx_i | )$, where $\kappa_1: \R \to \R_{\geq 0}$ is an univariate kernel function, $\ervx_i$ denotes the $i$-th coordinate of $\rvx$ for any $ \rvx \in \R^D$, and $c, c'$ are positive constants ensuring that normalization condition for $\kappa$ is satisfied. 
Note that, with the RBF kernel (which becomes the Gaussian kernel with proper scaling for normalization) with $\kappa_1(z) = \exp(-|z|^2), z \in \R$, $\kappa_1(\norm{\rvx}) = \prod_i \kappa_1( | \ervx_i | ) = \exp(-\norm{ \rvx }^2)$. We will consider the univariate case from hereon for the ease of exposition with $D = 1$, where $\kappa: \R \to \R_{\geq 0}$ satisfying the aforementioned symmetry, nonnegativity and normalization conditions.

For the purpose of KDE, the scale of the kernel function is not unique. That is, for a given $\kappa(\cdot)$, we can define $\tilde{\kappa}(\cdot)= b^{-1}\kappa(\cdot/b)$, for some $b>0$. Then, one obtains the same KDE by rescaling the choice of $h$. Hence, the shape of the kernel function plays a more important role in determining the choice of the kernel. We now introduce two parameters associated with the kernel -- the {\em scale} $\mu_\kappa$ and the {\em regularity} $\sigma_\kappa$ defined as:
\begin{equation} \label{eq:kprop}
\mu_\kappa \triangleq \int_x x^2 \kappa( x ) \, dx, \quad \sigma_\kappa \triangleq \int_x  \left( \kappa( x ) \right)^2 \, dx
\end{equation}
The quality or generalization of KDE depends on these two properties of the kernel. The generalization error of $ \hat{p}_h(\cdot; \bXi) $ is measured by the {\em Mean Integrated Squared Error} or \textsf{MISE}, and is given by 
\begin{equation} \label{eq:mise}
\textsf{MISE}(h) = \E{ \int_v ( \hat{p}_h(v; \bXi) - \pdata(v) )^2 dv },
\end{equation}
where the expectation is over the $K$ random samples $\bXi$ from $\pdata$.

\begin{figure}[htb]
\centering
\includegraphics[width=0.85\linewidth]{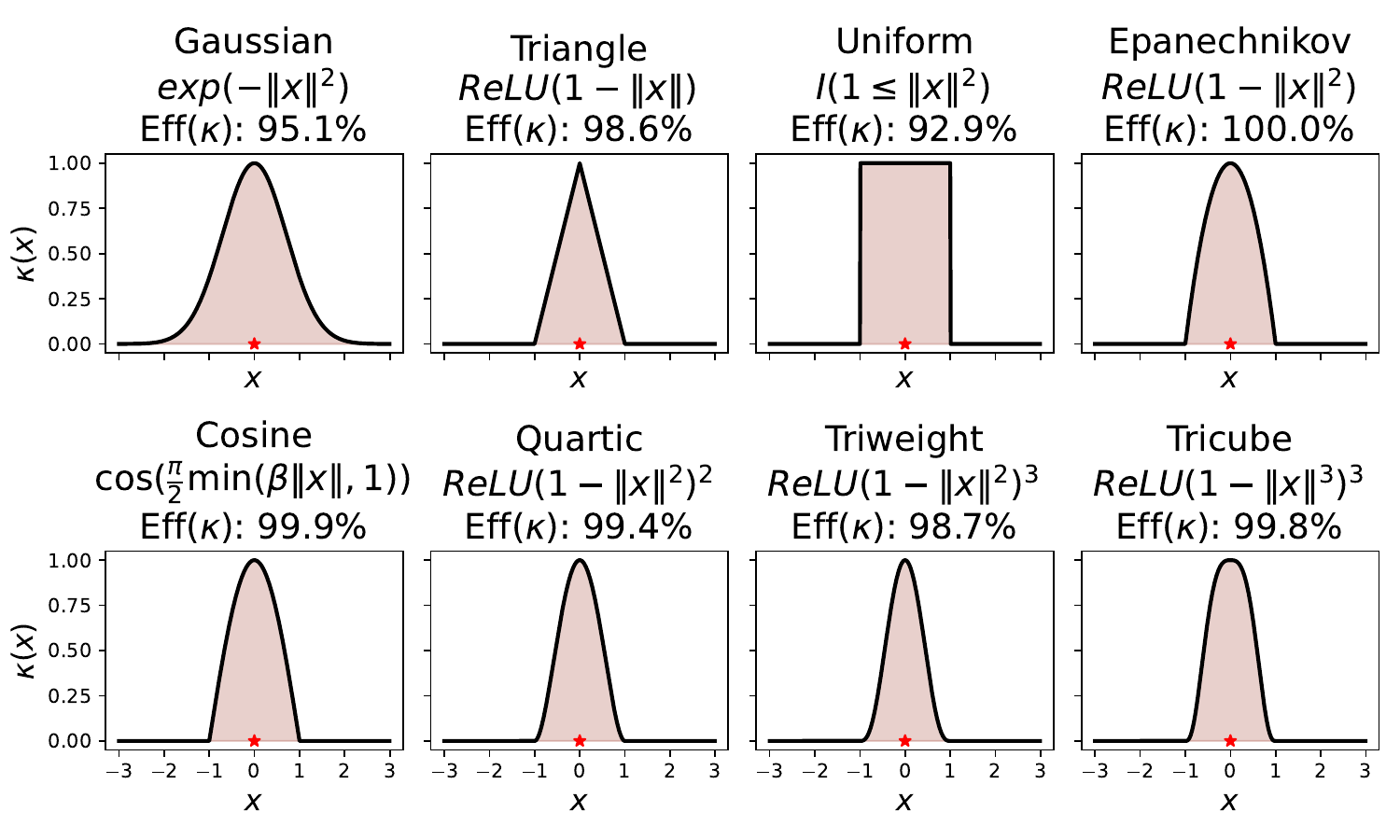}
\caption{{\bf Different kernels used for Kernel Density Estimation.} We show the shapes, expression and KDE efficiency relative to the Epanechnikov kernel ({\em higher is better}) for 8 kernels. The center of each kernel is marked with a \textcolor{Red}{red $\boldsymbol{\star}$}. To highlight the shape of the kernel, we have removed any scaling in the kernel expression. Note that all above kernels except Gaussian have finite support. The Epanechnikov kernel has the highest efficiency (100\%). The Gaussian kernel is extremely popular, and it is more efficient (95.1\%) than the Uniform kernel (92.9\%). However, there are various other kernels (such as the Triangle kernel) with better efficiency. This image is replicated from Hoover et al., 2025~\citep{hoover2025dense}.}
\label{fig:kernel-shapes}
\end{figure}

Assuming that the ground-truth density $\pdata$ is twice continuously differentiable, a second-order Taylor expansion gives the leading terms of the $\textsf{MISE}(h)$, which decomposes into squared bias and variance terms~\citep[Section 2.5]{wand1994kernel}:
\begin{equation} \label{eq:mise-bound}
\textsf{MISE}(h) \approx 
\underbrace{
\frac{\mu_\kappa^2}{4} h^4 \int_v \left| \pdata'' (v) \right|^2 \, dv 
}_{\text{bias-squared term}}
+ 
\underbrace{
\frac{\sigma_\kappa}{K h}
}_{\text{variance}}.
\end{equation}
Thus, reducing the bandwidth $h$ decreases bias but increase variance, and vice verse for increasing $h$, thereby highlighting the bias-variance tradeoff. Balancing the bias-squared and variance terms, we can have kernel-specific optimal choice $h_\kappa^\star$ for the bandwidth
\begin{equation}\label{eq:opt-bw}
 h_\kappa^\star \triangleq \left(
   \frac{\sigma_\kappa}{ K \mu_\kappa^2 }\frac{4}{\int_v \left| \pdata'' ( v ) \right| ^2 \, dv},
 \right)^{1/5}.
\end{equation}
Plugging this into \cref{eq:mise-bound} gives us the best possible \textsf{MISE}:
\begin{align}\label{eq:opt-mise}
\textsf{MISE}(h_\kappa^\star) \approx \frac{5}{4} \left(
   \frac{\sqrt{\mu_\kappa} \sigma_\kappa \int_v \left|\pdata''( v ) \right|^2 \, dv }{ K }
 \right)^{4/5},
\end{align}
where the choice of the kernel function $\kappa$ affects the \textsf{MISE} through its scale $\mu_\kappa$ and regularity $\sigma_\kappa$. Thus, it is intuitive to select the kernel function $\kappa$ based on the optimal $\textsf{MISE}(h_\kappa^\star)$. As discussed above, the scale of the kernel function is non-unique, and can be fixed to $\mu_\kappa = 1$ by appropriately scaling the kernel function. Hence, the kernel with the smallest regularity $\sigma_\kappa$, subject to $\mu_\kappa=1$ (without loss of generality), over the class of normalized, symmetric, and positive kernels is most desirable. This is a well-studied problem~\citep{epanechnikov1969non, muller1984smooth}~\citep[Section 2.7]{wand1994kernel}, and the Epanechnikov kernel $\kappa_\textsf{epan}(z) = \max\{1 - |z|^2, 0\} = \relu{1 - |z|^2}$ achieves the optimal $\textsf{MISE}(h_\kappa^\star)$. The quantity, $\textsf{Eff}(\kappa)\triangleq \sigma_\kappa /\sigma_{\kappa_\textsf{epan}}$ is known as the {\em efficiency} of any kernel relative to the Epanechnikov kernel. Various kernels with varying levels of efficiencies have been developed, and we present a representative subset of these kernel functions in \cref{fig:kernel-shapes}. Similar analysis and guarantees can be established for multivariate KDE.

The rich literature of KDE and its suite of well-studied kernel functions opens up the path to the development of various energy functions for Associative Memory networks --- one for each kernel function --- which have not been considered previously. Note that while leveraging the connection between the energy function and density estimation is one avenue to develop novel energy functions, there are other avenues such as ones building upon the connections between the LSE energy, regularized argmax transformations, and Fenchel-Young losses~\citep{blondel2020learning} to give us Hopfield-Fenchel-Young energy functions~\citep{santos2024hopfield}.

As a natural first choice, one can select the optimal Epanechnikov kernel, which leads to the following novel energy function~\citep{hoover2025dense}:
\begin{equation} \label{eq:en-epan}
E_\beta(\rvv; \bXi) = - \log \sum_{\mu = 1}^K \relu{1 - \nicefrac{\beta}{2} \norm{\rvv - \bxi^\mu}^2}.
\end{equation}
This makes use of a {\em shifted-ReLU} operation, and thus is termed the log-sum-ReLU or LSR energy. This can be contrasted with the popular LSE or log-sum-exp energy shown in \cref{eq:en-lse}, where we replace the exponential separation function $F(z) = \exp(z)$ with the shifted-ReLU separation function $F(z) = \relu{1+z}$ with the negative squared Euclidean distance based similarity function $S[\rvx, \rvx'] = -\nicefrac{1}{2} \norm{\rvx - \rvx'}$.

\begin{figure}[htb]
\centering
\includegraphics[width=\textwidth]{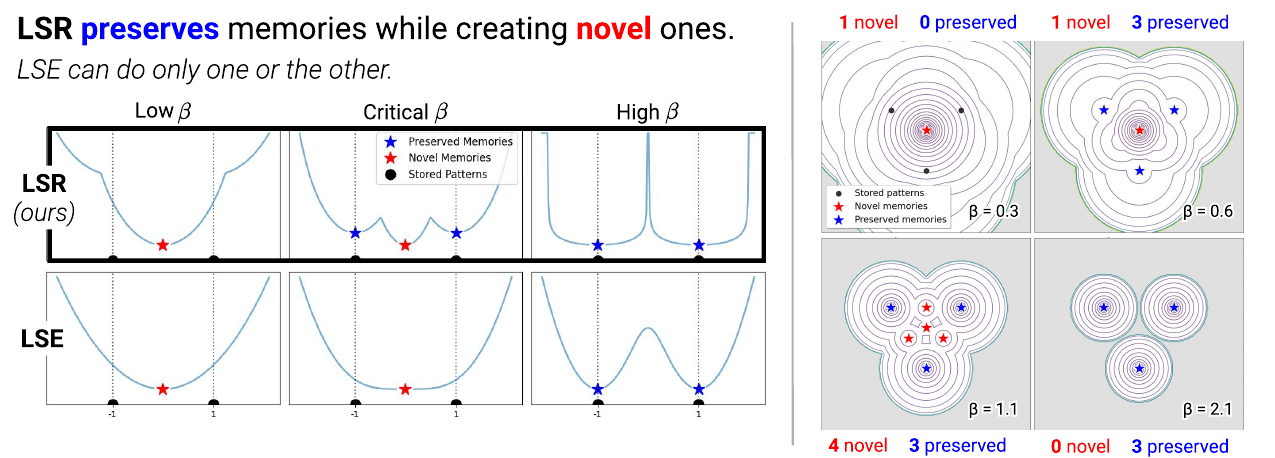}
\caption{{\bf Emergence of novel energy local minima.} LSR energy can create more memories than there are stored patterns under critical regimes of $\beta$. 
{\bf Left}:~1D LSR vs LSE energy landscape. Note that LSE is never capable of having more local minima than the number of stored patterns.
{\bf Right}:~2D LSR energy landscape, where increasing $\beta$ creates novel local minima where basins intersect. Unsupported regions are shaded gray.
This image is replicated from Hoover et al., 2025~\citep{hoover2025dense}.}
\label{fig:epan-en-emergence}
\end{figure}

This novel energy function has various desirable properties:

\begin{calloutInfo}[Exact single-step retrieval]
For an Associative Memory network with the LSR energy and appropriate hyperparameters, it is possible to have exact retrieval of stored patterns in a single energy gradient step. This is in contrast to LSE where only approximate retrieval is possible unless the inverse-temperature $\beta \to \infty$.
\end{calloutInfo}

\begin{calloutInfo}[Exponential memory capacity without exponential separation function]
This Associative Memory network equipped with the LSR energy has exponential memory capacity --- that is, the number of stored patterns that are retrievable is $O(\exp(D))$. This is similar to the LSE energy.
\end{calloutInfo}

\begin{calloutInfo}[Generation of a multitude of novel memories]
Finally, the LSE energy can introduce numerous novel energy local minima to the energy landscape, while also maintaining local minima around the stored patterns, enabling simultaneous retrieval of stored patterns and retrieval of novel memory, providing a path to data generation in Associative Memory networks with energy descent. This phenomena is visualized for data in one and two dimensions in \cref{fig:epan-en-emergence}, and has been utilized to create novel samples from an approximation of the underlying data distribution $\pdata$. Such a phenomena has not previously been seen in literature.
\end{calloutInfo}

However, this novel LSR energy can pose certain novel challenges:

\begin{calloutInfo}[Regions of infinite energy]
For a given configuration of an Associative Memory network, with LSR energy, there exists $\rvv \in \gX$ such that $E_\beta(\rvv; \bXi) = \infty$ given the finite support of the Epanechnikov kernel. This is visualized in \cref{fig:epan-en-emergence}~(Right) as the gray shaded region for two dimensional data.
\end{calloutInfo}

\clearpage{}%
\clearpage{}%
\chapter{Conclusion}
\label{chap:conclusion}
In this tutorial we have covered recent advances in energy-based Associative Memory, including information storage capacity calculations (Chapter \ref{chap:DenseAM}), relationship to transformers (Chapter \ref{chap:am-blocks}) and diffusion models (Chapter \ref{chap:am-ebm}), and connection to non-neural network machine learning (Chapter \ref{chap:am-stats}) and other ideas. In the past few years, Associative Memory became an active area of research with many lines of exploration coexisting and branching into several disciplines. Since this tutorial was prepared for ICML audience, we focused mostly on explaining the core ideas with only minimal derivations necessary to understand those ideas. We also prepared coding notebooks to help AI practitioners gain hands-on experience with AM basics. Inevitably, with this strategy in mind, many important and exciting aspects of AMs remained outside the scope of this tutorial. For instance, we did not discuss the biological implementations of DenseAMs \cite{krotov2021large,snow2022biological,tyulmankov2023memorization,kozachkov2025neuron,chandra2025episodic,kafrajbiologically,zhang2025maximizing}. Several valuable trends in AM-inspired statistical physics \cite{agliari2020neural,agliari2020tolerance,albanese2022replica,agliari2023dense,lucibello2024exponential,theriault2024dense, clark2025transient,mimura2025dynamical,nicoletti2025statistical} have also only been briefly mentioned. Memory augmentation of large language models (LLMs) \cite{burtsev2020memory,he2024camelot,rodkin2024associative,wang2025m+} is becoming an active area of research with clever ideas on how memory models can be used synergistically with feed-forward architectures. There are exciting ideas around novel neural architectures inspired by AMs \cite{ramsauer2021hopfield,krotov2021hierarchical,furst2022cloob,liang2022modern,chaudhry2023long,karakida2024hierarchical,niu2024beyond}, and domain specific applications \cite{widrich2020modern,zhang2025operator} that have not been covered with sufficient detail either. Quantum DenseAMs is an emerging topic \cite{kimura2024analysis}. Neuromorphic hardware based on DenseAMs \cite{musa2025dense} is becoming a promising area of research too. 

We expect these trends to grow and new trends to appear. We hope that this introductory tutorial may provide an entry point for new researchers in this exciting field. \clearpage{}%
\bibliographystyle{unsrt}
\bibliography{references}

\end{document}